\documentclass[10pt,twocolumn,letterpaper]{article}

\usepackage[pagenumbers]{cvpr}

\usepackage{graphicx}
\usepackage{times}
\usepackage{epsfig}
\usepackage{graphicx}
\usepackage{amsmath}
\usepackage{amssymb}
\usepackage{xcolor}
\usepackage{caption}
\usepackage{subcaption}
\usepackage{tabularx}
\usepackage{rotating}
\usepackage{verbatim}
\usepackage{xcolor,colortbl}
\usepackage{etoolbox}
\usepackage[normalem]{ulem}
\usepackage{booktabs}
\usepackage{multirow}
\usepackage{lipsum}
\usepackage{stmaryrd}
\usepackage{stackengine}
\usepackage{makecell}
\usepackage{pifont}
\usepackage{cancel}
\usepackage{adjustbox}
\usepackage{dblfloatfix}

\graphicspath{{imgs/}}

\usepackage[hyperindex,breaklinks]{hyperref}
\hypersetup{colorlinks}

\newcommand{\cmark}{\ding{51}}%
\newcommand{\xmark}{\ding{55}}%

\newcommand{\ma}[1]{\ensuremath{\mathsf{#1}}}

\definecolor{LightGrey}{rgb}{.9,.9,.9}
\definecolor{White}{rgb}{1.,0.,1.}
\definecolor{first}{rgb}{.8,.0,.0}
\definecolor{second}{rgb}{.0,.6,.0}
\definecolor{third}{rgb}{.0,.0,.8}
\newcolumntype{g}{>{\columncolor{White}}c}

\definecolor{ceiling}{RGB}{214,  38, 40}   %
\definecolor{floor}{RGB}{43, 160, 4}     %
\definecolor{wall}{RGB}{158, 216, 229}  %
\definecolor{window}{RGB}{114, 158, 206}  %
\definecolor{chair}{RGB}{204, 204, 91}   %
\definecolor{bed}{RGB}{255, 186, 119}  %
\definecolor{sofa}{RGB}{147, 102, 188}  %
\definecolor{table}{RGB}{30, 119, 181}   %
\definecolor{tvs}{RGB}{160, 188, 33}   %
\definecolor{furniture}{RGB}{255, 127, 12}  %
\definecolor{objects}{RGB}{196, 175, 214} %

\definecolor{car}{rgb}{0.39215686, 0.58823529, 0.96078431}
\definecolor{bicycle}{rgb}{0.39215686, 0.90196078, 0.96078431}
\definecolor{motorcycle}{rgb}{0.11764706, 0.23529412, 0.58823529}
\definecolor{truck}{rgb}{0.31372549, 0.11764706, 0.70588235}
\definecolor{other-vehicle}{rgb}{0.39215686, 0.31372549, 0.98039216}
\definecolor{person}{rgb}{1.        , 0.11764706, 0.11764706}
\definecolor{bicyclist}{rgb}{1.        , 0.15686275, 0.78431373}
\definecolor{motorcyclist}{rgb}{0.58823529, 0.11764706, 0.35294118}
\definecolor{road}{rgb}{1.        , 0.        , 1.        }
\definecolor{parking}{rgb}{1.        , 0.58823529, 1.        }
\definecolor{sidewalk}{rgb}{0.29411765, 0.        , 0.29411765}
\definecolor{other-ground}{rgb}{0.68627451, 0.        , 0.29411765}
\definecolor{building}{rgb}{1.        , 0.78431373, 0.        }
\definecolor{fence}{rgb}{1.        , 0.47058824, 0.19607843}
\definecolor{vegetation}{rgb}{0.        , 0.68627451, 0.        }
\definecolor{trunk}{rgb}{0.52941176, 0.23529412, 0.        }
\definecolor{terrain}{rgb}{0.58823529, 0.94117647, 0.31372549}
\definecolor{pole}{rgb}{1.        , 0.94117647, 0.58823529}
\definecolor{traffic-sign}{rgb}{1.        , 0.        , 0.    }   

\makeatletter
\newcommand{\car@semkitfreq}{3.92}
\newcommand{\bicycle@semkitfreq}{0.03}
\newcommand{\motorcycle@semkitfreq}{0.03}
\newcommand{\truck@semkitfreq}{0.16}
\newcommand{\othervehicle@semkitfreq}{0.20}
\newcommand{\person@semkitfreq}{0.07}
\newcommand{\bicyclist@semkitfreq}{0.07}
\newcommand{\motorcyclist@semkitfreq}{0.05}
\newcommand{\road@semkitfreq}{15.30}  %
\newcommand{\parking@semkitfreq}{1.12}
\newcommand{\sidewalk@semkitfreq}{11.13}  %
\newcommand{\otherground@semkitfreq}{0.56}
\newcommand{\building@semkitfreq}{14.1}  %
\newcommand{\fence@semkitfreq}{3.90}
\newcommand{\vegetation@semkitfreq}{39.3}  %
\newcommand{\trunk@semkitfreq}{0.51}
\newcommand{\terrain@semkitfreq}{9.17} %
\newcommand{\pole@semkitfreq}{0.29}
\newcommand{\trafficsign@semkitfreq}{0.08}
\newcommand{\semkitfreq}[1]{{\csname #1@semkitfreq\endcsname}}

\newcommand{\ceiling@nyufreq}{1.37}
\newcommand{\floor@nyufreq}{17.58}
\newcommand{\wall@nyufreq}{15.26}
\newcommand{\window@nyufreq}{1.99}
\newcommand{\chair@nyufreq}{3.01}
\newcommand{\bed@nyufreq}{7.08}
\newcommand{\sofa@nyufreq}{4.70}
\newcommand{\table@nyufreq}{4.31}
\newcommand{\tvs@nyufreq}{0.47}
\newcommand{\furniture@nyufreq}{30.04}
\newcommand{\objects@nyufreq}{14.19}
\newcommand{\nyufreq}[1]{{\csname #1@nyufreq\endcsname}}

\newcommand{\OurMethod}{MonoScene}
\newcommand{\ContextPrior}{3D Context Relation Prior}

\usepackage[capitalize]{cleveref}
\crefname{section}{Sec.}{Secs.}
\Crefname{section}{Section}{Sections}

\Crefname{table}{Table}{Tables}
\crefname{table}{Tab.}{Tabs.}

\crefname{equation}{Eq.}{Eqs.}
\Crefname{equation}{Equation}{Equations}

\newcommand*{\x}{\mathsf{x}\mskip1mu}
\newcommand{\condenseparagraph}[1]{\noindent\textbf{#1}\quad}

\begin{document}

\title{\OurMethod{}: Monocular 3D Semantic Scene Completion}

\author{Anh-Quan Cao\\
Inria\\
{\tt\small anh-quan.cao@inria.fr}
\and
Raoul de Charette\\
Inria\\
{\tt\small raoul.de-charette@inria.fr}
}
\maketitle

\begin{abstract}
MonoScene proposes a 3D Semantic Scene Completion (SSC) framework, where the dense geometry and semantics of a scene are inferred from a single monocular RGB image.
Different from the SSC literature, relying on 2.5 or 3D input, we solve the complex problem of 2D to 3D scene reconstruction while jointly inferring its semantics.
Our framework relies on successive 2D and 3D UNets, bridged by a novel 2D-3D features projection inspired by optics, and introduces a 3D context relation prior to enforce spatio-semantic consistency. Along with architectural contributions, we introduce novel global scene and local frustums losses.
Experiments show we outperform the literature on all metrics and datasets while hallucinating plausible scenery even beyond the camera field of view. Our code and trained models are available at 
\href{https://github.com/cv-rits/MonoScene}{https://github.com/cv-rits/MonoScene}.
\end{abstract}
\section{Introduction}
Estimating 3D from an image is a problem that goes back to the roots of computer vision~\cite{roberts1963machine}.
While we, humans, naturally understand a scene from a single image, reasoning all at once about geometry and semantics, this was shown remarkably complex by decades of research~\cite{zhang1999shape,saxena2008make3d,zhang2021holistic}.
Subsequently, many algorithms use dedicated depth sensors such as Lidar~\cite{3dreclidar1, vizzo2021icra, park2018high} or depth cameras~\cite{Choe2021VolumeFusionDD, Avetisyan_2019_CVPR, dai2020sgnn}, easing the 3D estimation problem.
These sensors are often more expensive, less compact and more intrusive than cameras which are widely spread and shipped in smartphones, drones, cars, etc.
Thus, being able to estimate a 3D scene from an image would pave the way for new applications.

3D Semantic Scene Completion (SSC) addresses scene understanding as it seeks to \textit{jointly} infer its geometry and semantics. While the task gained popularity recently~\cite{sscsurvey}, the existing methods still rely on depth data (\ie occupancy grids, point cloud, depth maps, etc.) and are custom designed for \textit{either} indoor or outdoor scenes.
\begin{figure}
	\centering
	\footnotesize
	\newcolumntype{P}[1]{>{\centering\arraybackslash}p{#1}}
	\renewcommand{\arraystretch}{0.0}
	\setlength{\tabcolsep}{0.035\linewidth}
	\begin{tabular}{P{0.45\columnwidth} P{0.45\columnwidth}}
		\includegraphics[width=1.0\linewidth]{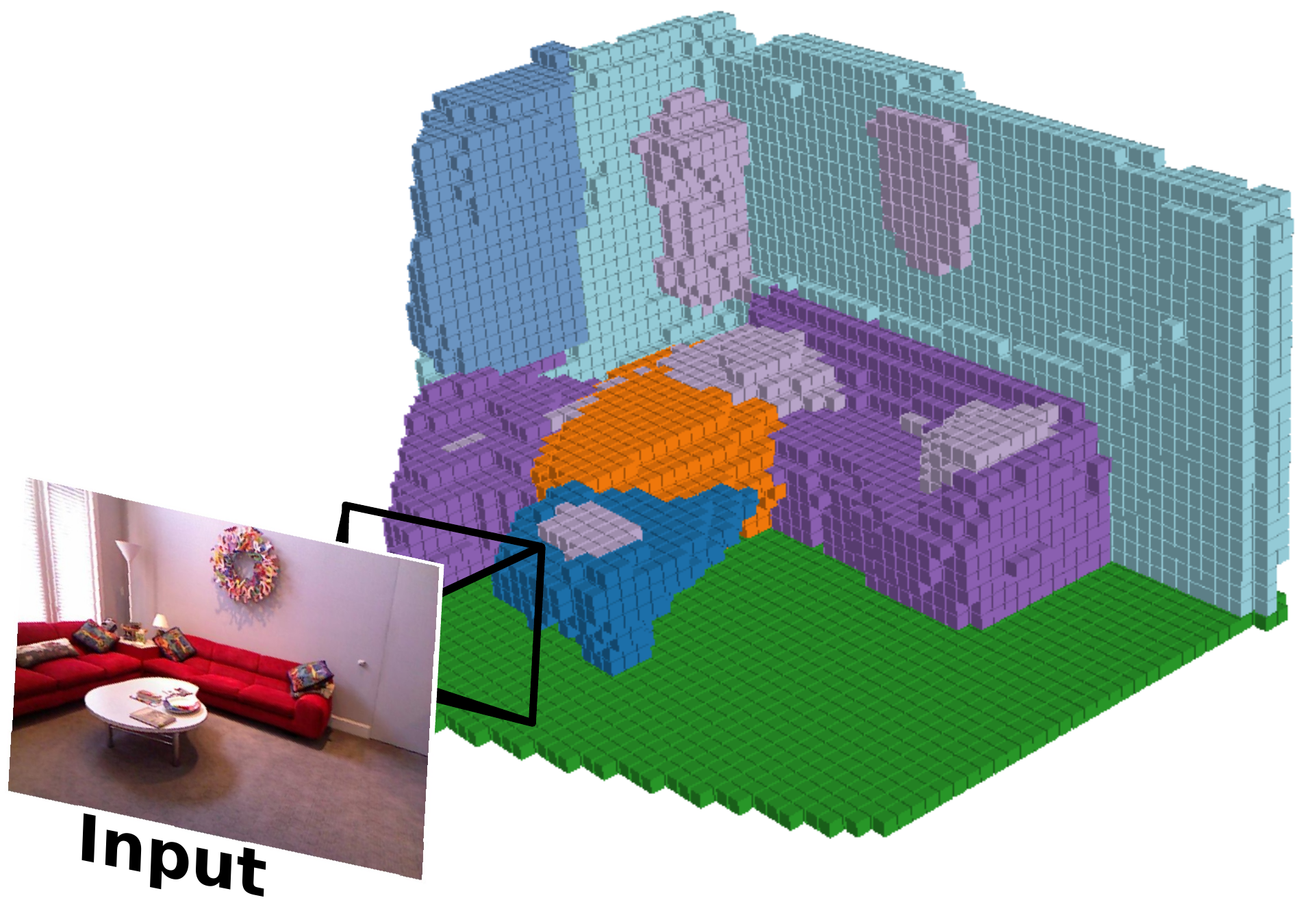} & \includegraphics[width=1.0\linewidth]{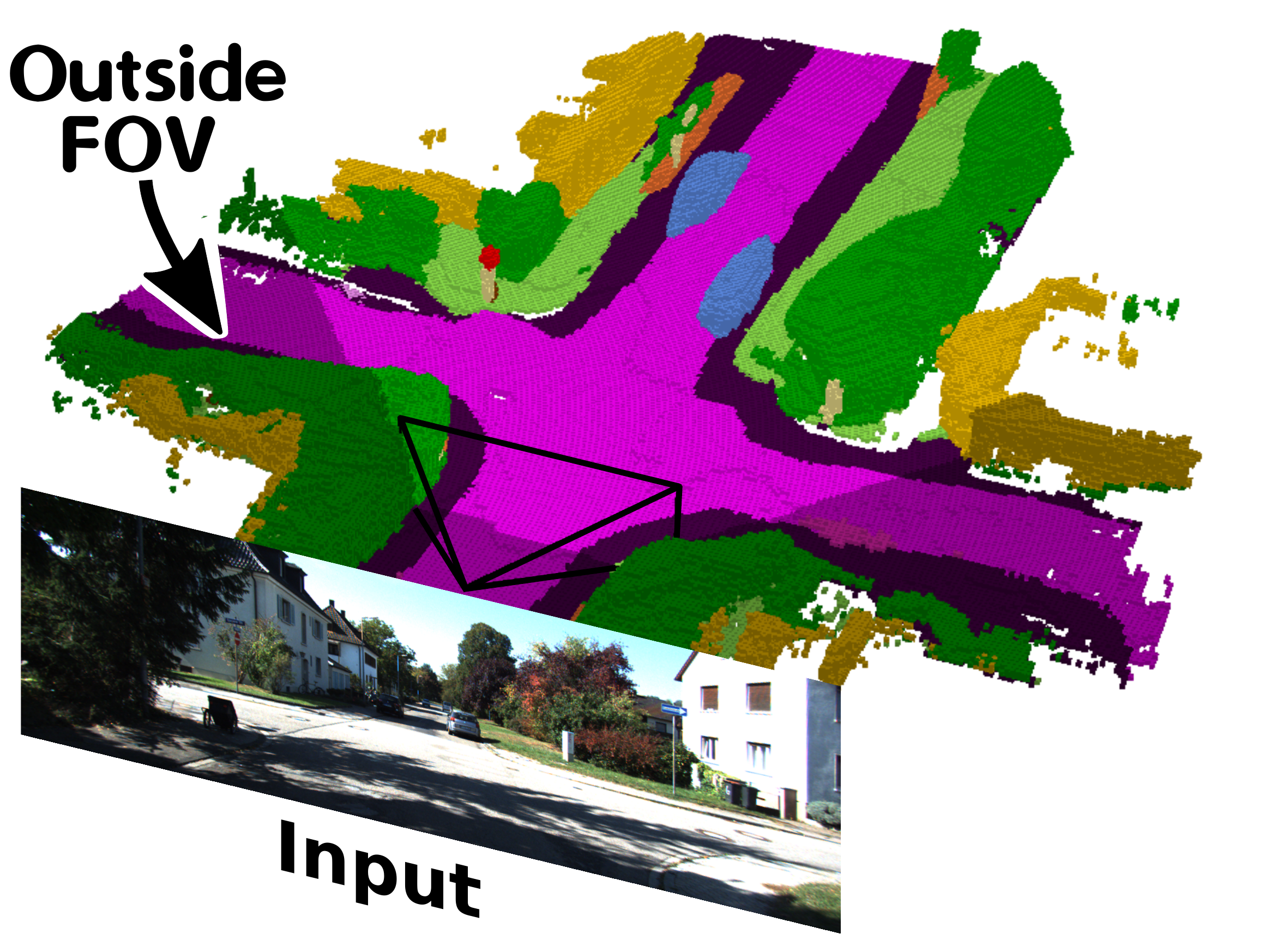}
		\\
		Indoor (NYUv2\cite{NYUv2}) & Outdoor (SemanticKITTI~\cite{behley2019iccv})
	\end{tabular}
	\caption{\textbf{RGB Semantic Scene Completion with MonoScene.} Our framework infers dense semantic scenes, hallucinating scenery outside the field of view of the image (dark voxels, right).}
	\label{fig:teaser}
\end{figure}

Here, we present MonoScene which -- unlike the literature -- relies on a \textit{single} RGB image to infer the dense 3D voxelized semantic scene working indifferently for indoor and outdoor scenes. To solve this challenging problem, we project 2D features along their line of sight, inspired by optics, bridging 2D and 3D networks while letting the 3D network self-discover relevant 2D features. 
The SSC literature mainly relies on cross-entropy loss which considers each voxel independently, lacking context awareness. 
We instead propose novel SSC losses that optimize the semantic distribution of group of voxels, both globally and in local frustums.
Finally, to further boost context understanding, we design a 3D context layer to provide the network with a global receptive field and insights about the voxels semantic relations.
We extensively tested MonoScene on indoor and outdoor, see~\cref{fig:teaser}, where it outperformed all comparable baselines and even some 3D input baselines. Our main contributions are summarized as follows.
\begin{itemize}
	\setlength{\itemsep}{1pt}
	\setlength{\parskip}{0pt}
	\setlength{\parsep}{0pt}
	\item MonoScene: the first SSC method tackling both outdoor and indoor scenes from a single RGB image.
	\item A mechanism for 2D Features Line of Sight Projection bridging 2D and 3D networks (FLoSP, \cref{sec:unprojection}).
	\item A 3D Context Relation Prior (3D CRP, \cref{sec:context_prior}) layer that boosts context awareness in the network.
	\item New SSC losses to optimize scene-class affinity (\cref{sec:global_loss}) and local frustum proportions (\cref{sec:frustum_loss}).
\end{itemize}
\section{Related works}
\begin{figure*}
	\centering
	\includegraphics[width=\textwidth]{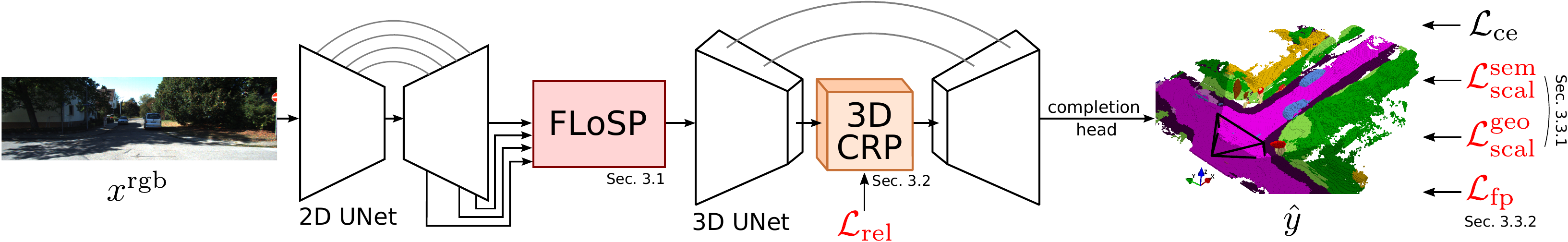}
	\caption{\textbf{\OurMethod{} framework.} We infer 3D SSC from a single RGB image, leveraging 2D and 3D UNets, bridged by our Features Line of Sight Projection (FLoSP \cref{sec:unprojection}), and a 3D Context Relation Prior (3D CRP, \cref{sec:context_prior}) to enhance spatio-semantic awareness. On top of standard cross-entropy ($\mathcal{L}_{\text{ce}}$), our Scene-Class Affinity loss ($\mathcal{L}_{\text{scal}}$, \cref{sec:global_loss}) improves the global semantics ($\mathcal{L}^{\text{sem}}_{\text{scal}}$) and geometry ($\mathcal{L}^{\text{geo}}_{\text{scal}}$), and our Frustums Proportion loss ($\mathcal{L}_{\text{fp}}$, \cref{sec:frustum_loss}) enforces class distribution in local frustums, providing supervision beyond occlusions.}
	\label{fig:network-architecture}
\end{figure*}
\noindent \textbf{3D from a single image.}\quad
Despite early researches~\cite{zhang1999shape,hoiem2005geometric,saxena2008make3d}, in the deep learning era the first works focused on single 3D object reconstruction with explicit~\cite{pix2vox, pix2voxplus, choy20163dr2n2, wu2017learning, girdhar2016learning, achlioptas2018poincloud, Fan_2017_CVPR, yang2019pointflow, wang2018pixel2mesh, chen2020bspnet, deepmarchingcube} or implicit representations~\cite{Park2019DeepSDFLC, OccupancyNetworks, PengNMP020convoccnet, xu_disn_2019,Popov2020CoReNetC3}. A comprehensive survey on the matter is~\cite{han2019image}.
For multiple objects, a common practice is to couple reconstruction with object detection~\cite{izadinia2017im2cad, kundu20183DRCNN, gkioxari2020meshRCNN,zhang2020perceiving,grabner2019location}.
Closer to our work, holistic 3D understanding seeks to predict the scene and objects layout~\cite{nie2020total3dunderstanding,huang2018holistic,zhang2021holistic,lee2017roomnet,sun2019horizonnet,zou2018layoutnet}, reaching a sparse scene representation. Only~\cite{dahnert2021panoptic} recently addressed indoor dense \textit{visible} panoptic reconstruction by backprojecting individual 2D task features to 3D. 
We instead densely estimate semantics and geometry for both indoor and outdoor scenarios.

\noindent \textbf{3D semantic scene completion (SSC).}\quad
SSCNet~\cite{song2016ssc} first defined the `SSC' task where geometry and semantics are \textit{jointly} inferred. The task gained attention lately, and is thoroughly reviewed in a survey~\cite{sscsurvey}.
Existing works all use geometrical inputs like depth \cite{SATNet,Li2019ddr,TS3D,Li2020AttentionbasedMF,PALNet,Li2020aicnet,Cheng2020S3CNetAS}, occupancy grids~\cite{TS3D,roldao2020lmscnet,Wu2020SCFusionRI,yan2021JS3CNet} or point cloud~\cite{Zhong2020SemanticPC,localDif}. Truncated Signed Distance Function (TSDF) were also proved informative~\cite{song2016ssc,zhang2018efficient,EdgeNet,dourado360degreeSSC,Chen20193DSS,ForkNet,CCPNet,PALNet,3DSketch,Cheng2020S3CNetAS,SISNet}. Among others originalities, some SSC works use adversarial training to guide realism~\cite{ForkNet, Chen20193DSS}, exploit multi-task~\cite{IMENet, SISNet}, or use lightweight networks~\cite{Li2019ddr,roldao2020lmscnet}.
Of interest for us, while others have used RGB as input~\cite{Guedes2018SemanticSC,Cherabier2018LearningPF,SATNet,Li2019ddr,TS3D,EdgeNet,Chen2019AM2FNetAM,EdgeNet,Li2020AttentionbasedMF,3DSketch,Li2020aicnet,Zhong2020SemanticPC,SISNet} it is always along other geometrical input (\eg~depth, TSDF, etc.). A remarkable point in~\cite{sscsurvey} is that existing methods are designed for either indoor or outdoor, performing suboptimally in the other setting. The same survey highlights the poor diversity of losses for SSC.
Instead, we address SSC only using a single RGB image, with novel SSC losses, and are robust to various types of scenes.

\noindent \textbf{Contextual awareness.}\quad
Contextual features are crucial for semantics \cite{Yu-CVPR-ContextPrior-2020} and SSC~\cite{sscsurvey} tasks.
A simple strategy is to concatenate multiscale features with skip connections~\cite{song2016ssc, zhang2018efficient, dai2020sgnn,3DSketch, SATNet} or use dilated convolutions for large receptive fields~\cite{Yu2016MultiScaleCA}, for example with 
the popular Atrous Spatial Pyramid Pooling (ASPP)~\cite{Chen2018DeepLabSI} also used in SSC~\cite{SATNet, roldao2020lmscnet, Li2019ddr, Li2020aicnet}. 
Long-range information is gathered by self-attention in~\cite{CCNet, DANet} and global pooling in~\cite{dfn, encnet}. Explicit contextual learning is shown to be beneficial in~\cite{Yu-CVPR-ContextPrior-2020}. 
We propose a 3D contextual component that leverages multiple relation priors and provides a global receptive field.

\section{Method}
\label{sec:method}
3D Semantic Scene Completion (SSC) aims to jointly infer geometry and semantics of a 3D scene $\hat{y}$ by predicting labels $C = \{c_0, c_1, \dots, c_M\}$, being free class $c_0$ and $M$ semantic classes.
This has been almost exclusively addressed with 2.5D or 3D inputs~\cite{sscsurvey}, such as point cloud, depth or else, which act as strong geometrical cues.

Instead, MonoScene solves voxel-wise SSC from a single RGB image $x^{\text{rgb}}$, learning $\hat{y} = f(x^{\text{rgb}})$. 
This is significantly harder due to the complexity of recovering 3D from 2D.
Our pipeline in \cref{fig:network-architecture} uses 2D and 3D UNets bridged by our Features Line of Sight Projection module (FLoSP, \cref{sec:unprojection}), lifting 2D features to plausible 3D locations, that boosts information flow and enables 2D-3D disentanglement. 
Inspired by~\cite{Yu-CVPR-ContextPrior-2020}, we capture long-range semantic context with our \ContextPrior{} component (3D CRP, \cref{sec:context_prior}) inserted between the 3D encoder and decoder.
To guide the SSC training, we introduce new complementary losses.
First, a Scene-Class Affinity Loss (\cref{sec:global_loss}) optimizes the intra-class and inter-class scene-wise metrics.
Second, a Frustum Proportion Loss{} (\cref{sec:frustum_loss}) aligns the classes distribution in local frustums, which provides supervision beyond scene occlusions.\\

\condenseparagraph{2D-3D backbones.}We rely on consecutive 2D and 3D~UNets with standard skip connections. The 2D UNet bases on a pre-trained EfficientNetB7~\cite{effcientnet} taking as input the image $x^{\text{rgb}}$. 
The 3D UNet is a custom shallow encoder-decoder with 2 layers.
The SSC output $\hat{y}$ is obtained by processing the 3D UNet output features with our completion head holding a 3D ASPP~\cite{Chen2018DeepLabSI} block and a softmax layer.
\begin{figure}
	\centering
	\includegraphics[width=.9\linewidth]{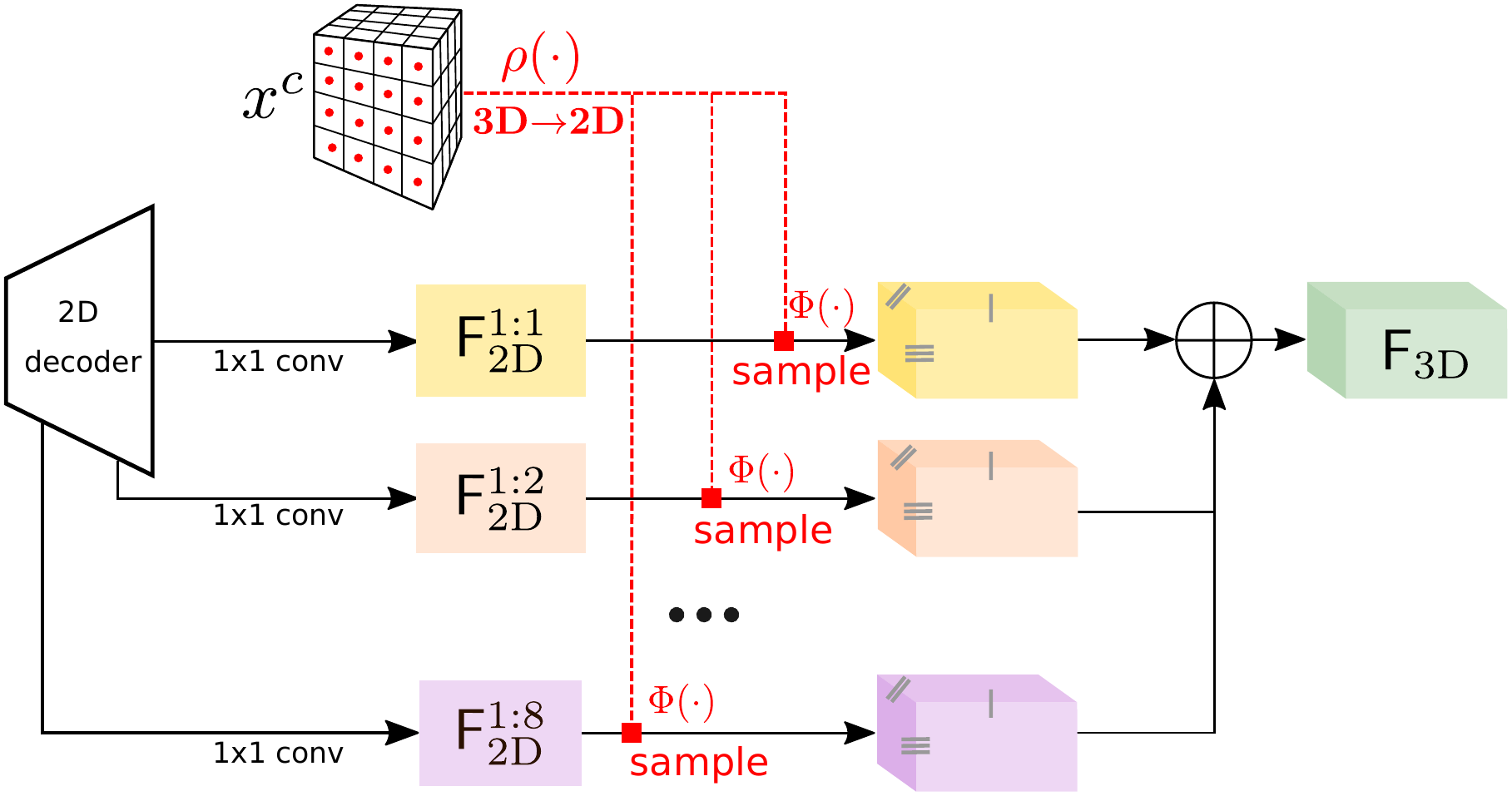}
	\caption{\textbf{Features Line of Sight Projection (FLoSP).} 
		We project multi-scale 2D features $\ma{F}^\text{1:s}_\text{2D}$ (here, $s \in \{1,2,4,8\}$) along their line of sight by sampling ($\Phi(\cdot)$) them where the 3D voxels centroids ($x^c$) project ($\rho(\cdot)$).
		This boosts the 2D-3D information flow, and lets the 3D network discover which 2D features are relevant.}
	\label{fig:projection}
\end{figure}
\subsection{Features Line of Sight Projection (FLoSP)}
\label{sec:unprojection}
Lifting 2D to 3D is notoriously ill-posed due to the scale ambiguity of single view point~\cite{single_view_survey}. 
We rather reason from optics and backproject multiscale 2D features to \textit{all possible} 3D correspondences, that is along their optical ray, aggregated in a unique 3D representation.
Our intuition here is that processing the latter with a 3D network will provide guidance from the ensemble of 2D features.
Our projection mechanism is akin to~\cite{Popov2020CoReNetC3} but the latter projects each 2D map to a given 3D map -- acting as 2D-3D skip connections. 
In contrast, our component bridges the 2D and 3D networks by lifting multiscale 2D features to a \textit{single} 3D feature map. 
We argue this enables 2D-3D disentangled representations, providing the 3D network with the freedom to use high-level 2D features for fine-grained 3D disambiguation. 
Compared to \cite{Popov2020CoReNetC3}, ablation in \cref{sec:exp_ablations} shows our strategy is significantly better.

Our process is illustrated in \cref{fig:projection}. 
In practice, assuming known camera intrinsics, we project 3D voxels centroids~($x^{\text{c}}$) to 2D and sample corresponding features from the 2D decoder feature map $\ma{F}^{1:s}_\text{2D}$ of scale $1{:}s$.
Repeating the process at all scales $S$, the final 3D feature map $\ma{F}_\text{3D}$ writes
\begin{equation}
	\ma{F}_{\text{3D}} = \sum_{s \in S} \Phi_{\rho(x^{\text{c}})} (\ma{F}^{1\text{:}s}_{\text{2D}})\,,
	\label{eq:project2d3d}
\end{equation} 
where $\Phi_a(b)$ is the sampling of $b$ at coordinates $a$, and $\rho(\cdot)$ is the perspective projection. 
In practice, we backproject from scales $S{=}\{1,2,4,8\}$, and apply a 1x1 conv on 2D maps before sampling to allow summation. 
Voxels projected outside the image have their feature vector set to 0.
The output map $\ma{F}_{\text{3D}}$ is used as 3D UNet input. 

\begin{figure}
	\centering
	\begin{subfigure}[b]{0.35\linewidth}
		\centering
		\captionsetup{font=scriptsize}
		\includegraphics[width=0.8\textwidth]{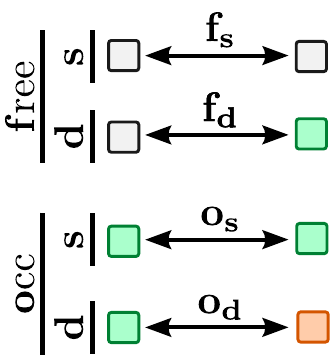} 
		\caption{Voxel$\leftrightarrow$Voxel relations}
		\label{fig:relationship_categories}
	\end{subfigure}\hfill%
	\begin{subtable}[b]{0.6\linewidth}
		\captionsetup{font=scriptsize}
		\newcommand{\cellvmiddle}[1]{\multicolumn{1}{c|}{\begin{minipage}{12mm}%
					\vspace{0.5mm}%
					\centering%
					{#1}%
					\vspace{0.5mm}%
		\end{minipage}}}
		
		\footnotesize
		\centering
		\setlength\tabcolsep{4pt}
		\renewcommand{\arraystretch}{1.2}
		\begin{tabular}{c|cc|cc|}
			& \multicolumn{2}{c|}{free} & \multicolumn{2}{c|}{occ} \\
			& \multicolumn{1}{c}{\textbf{f}$_{\textbf{s}}$} & \multicolumn{1}{c|}{\textbf{f}$_{\textbf{d}}$} & \multicolumn{1}{c}{\textbf{o}$_{\textbf{s}}$} & \multicolumn{1}{c|}{\textbf{o}$_\textbf{d}$} \\ 
			\hline
			\cellvmiddle{\includegraphics[height=5mm]{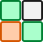}$\leftrightarrow$\includegraphics[width=2.5mm]{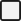}} & \cmark & \cmark & \xmark & \xmark\\
			\cellvmiddle{\includegraphics[height=5mm]{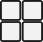}$\leftrightarrow$\includegraphics[width=2.5mm]{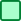}} & \xmark & \cmark & \xmark & \xmark\\
			\cellvmiddle{\includegraphics[height=5mm]{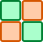}$\leftrightarrow$\includegraphics[width=2.5mm]{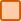}} & \xmark & \xmark & \cmark & \cmark\\ 
			\hline
		\end{tabular}
		
		\caption{Samples Supervoxel$\leftrightarrow$Voxel relations.}
		\label{table:supervoxel_relationship}
	\end{subtable}
	\caption{\textbf{2D illustration of 4-way relations.} \subref{fig:relationship_categories} We consider voxel$\leftrightarrow$voxel relations whether one is \textbf{f}ree or both are \textbf{o}ccupied, and if their semantics is \textbf{s}imilar or \textbf{d}ifferent.
		\subref{table:supervoxel_relationship} For memory reason, we encode Supervoxel$\leftrightarrow$Voxel relations framed as multi-label classification. 
		(\includegraphics[width=2.5mm]{imgs/rv1.png} free, \includegraphics[width=2.5mm]{imgs/rv2.png}\includegraphics[width=2.5mm]{imgs/rv3.png} occupied - colors denote semantics)}
	\label{fig:4_way_context_relation}
\end{figure}

\subsection{3D Context Relation Prior (3D CRP)}
\label{sec:context_prior}
Because SSC is highly dependent on the context~\cite{sscsurvey}, we inspire from CPNet~\cite{Yu-CVPR-ContextPrior-2020} that demonstrates the benefit of binary context prior for 2D segmentation.
Here, we propose a \ContextPrior{} (3D CRP) layer, inserted at the 3D UNet bottleneck, which learns \textit{n}-way voxel$\leftrightarrow$voxel semantic scene-wise relation maps.
This provides the network with a global receptive field, and increases spatio-semantic awareness due to the relations discovery mechanism.

Because SSC is a highly imbalanced task, learning binary (i.e. $n{=}2$) relations as in~\cite{Yu-CVPR-ContextPrior-2020} is suboptimal. 
We instead consider $n{=}4$ bilateral voxel$\leftrightarrow$voxel relations, grouped into \textbf{f}ree and \textbf{o}ccupied corresponding to `at least one voxel is free' and `both voxels are occupied', respectively. 
For each group, we encode whether the voxels semantic classes are \textbf{s}imilar or \textbf{d}ifferent, leading to the 4 non-overlapping relations: $\mathcal{M}{=}\{\textbf{f}_{\textbf{s}}, \textbf{f}_{\textbf{d}}, \textbf{o}_{\textbf{s}}, \textbf{o}_{\textbf{d}}\}$. ~\cref{fig:relationship_categories} illustrates the relations in 2D (see caption for colors meaning).

As voxels relations are greedy with $N^2$ relations for $N$ voxels, we present the lighter supervoxel$\leftrightarrow$voxel relations.\\

\noindent\textbf{Supervoxel$\leftrightarrow$Voxel relation.} \quad
We define supervoxels as non-overlapping groups of $s^3$ neighboring voxels each, and learn the smaller supervoxel$\leftrightarrow$voxel relation matrices of size $\frac{N^2}{s^3}$. 
Considering a supervoxel $\mathcal{V}$ having voxels  $\{\nu_1, \dots, \nu_{s^3}\}$ and a voxel $\nu$, there are $s^3$ pairwise relations $\{\nu_1{\leftrightarrow}\nu, \dots, \nu_{s^3}{\leftrightarrow}\nu\}$.
Instead of regressing the complex count of $\mathcal{M}$ relations in $\mathcal{V}{\leftrightarrow}\nu$, we predict which of the $\mathcal{M}$ relations exist, as depicted in \cref{table:supervoxel_relationship}. 
This writes,
\begin{equation}
	\mathcal{V}{\leftrightarrow}\nu = \{\nu_1{\leftrightarrow}\nu, \dots, \nu_{s^3}{\leftrightarrow}\nu\}_{\neq},
\end{equation}
where $\{\cdot\}_{\neq}$ returns distinct elements of a set.\\

\begin{figure}
	\centering
	\includegraphics[width=0.88\linewidth]{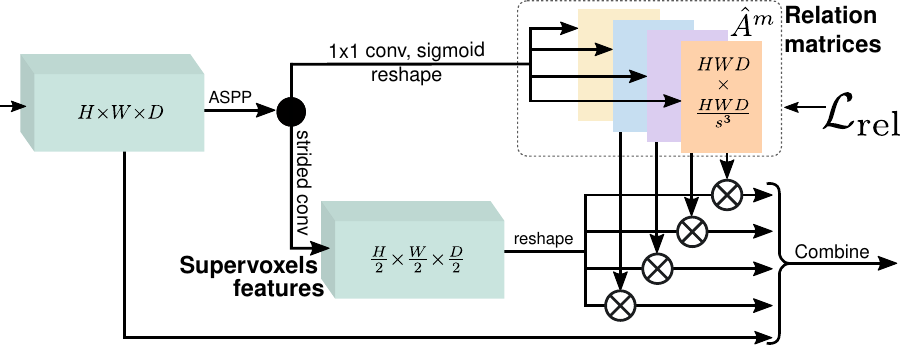}
	\caption{\textbf{\ContextPrior{} (3D CRP)}. We infer relation matrices $\hat{A}^m$ (here, 4), where each encodes a unique relation ${m \in \mathcal{M}}$ -- optionally supervised with a relation loss ($\mathcal{L}_\text{rel}$).
		The matrices are multiplied with the supervoxels features to gather context, and later combined (concate, conv, DDR~\cite{Li2019ddr}) with input features. The feature dimension is omitted for clarity.}
	\label{fig:context_prior_3D}
\end{figure}

\noindent\textbf{\ContextPrior{} Layer.}\quad \cref{fig:context_prior_3D} illustrates the architecture of our layer. 
It takes as input a 3D map of spatial dimension $H\x{}W\x{}D$, on which is applied a serie of ASPP convolutions~\cite{Chen2018DeepLabSI} to gather a large receptive field, then split into $n{=}|\mathcal{M}|$ matrices of size $HWD{\times}\frac{HWD}{s^3}$.

Each matrix $\hat{A}^m$ encodes a relation $m {\in} {M}$, supervised by its ground truth $A^m$.
We then optimize a weighted multi-label binary cross entropy loss:
\begin{equation}
	\mathcal{L}_{rel}{=}{-}\sum_{m\in \mathcal{M},i}[(1\text{-}A^m_i)\log (1\text{-}\hat{A}^m_i)\text{+}w_m A^m_i \log \hat{A}^m_i],
	\label{eq:loss_rel}
\end{equation}
where $i$ loops through all elements of the relation matrix and $w_m = \frac{\sum_i (1 - A^m_i)}{\sum_i A^m_i}$. 
The relation matrices are multiplied with reshaped supervoxels features to gather global context.

Alternatively, relations in $A^m$ can be self-discovered (w/o $\mathcal{M}$) by removing $\mathcal{L}_{\text{rel}}$, \ie behaving as attention matrices.
\subsection{Losses}
We now introduce new losses pursuing distinct global (\cref{sec:global_loss}) or local (\cref{sec:frustum_loss}) optimization objectives. 
\subsubsection{Scene-Class Affinity Loss}
\label{sec:global_loss}
We seek to explicitly let the network be aware of the global SSC performance. To do so, we build upon the 2D binary affinity loss in~\cite{Yu-CVPR-ContextPrior-2020} and introduce a multi-class version directly optimizing the scene- \textit{and} class- wise metrics. 

Specifically, we optimize the class-wise derivable (P)recision, (R)ecall and (S)pecificity where $P_c$ and $R_c$ measure the performance of similar class $c$ voxels, and $S_c$ measures the performance of dissimilar voxels (\ie not of class $c$).	
Considering $p_i$ the ground truth class of voxel $i$, and $\hat{p}_{i,c}$ its \textit{predicted} probability to be of class~$c$, we define:
\begingroup
\allowdisplaybreaks
\begin{align}
	P_{c}(\hat{p}, p) &= \log\frac{\sum_{i} \hat{p}_{i,c} \llbracket p_i = c \rrbracket}{\sum_{i}\hat{p}_{i,c}},\\
	R_{c}(\hat{p}, p) &= \log\frac{\sum_{i} \hat{p}_{i,c} \llbracket p_i = c \rrbracket}{\sum_{i}\llbracket p_i = c \rrbracket},\\
	S_{c}(\hat{p}, p) &= \log\frac{\sum_{i}({1 - \hat{p}_{i,c}}) (1 - \llbracket p_i = c \rrbracket)}{\sum_{i} (1 - \llbracket p_i = c \rrbracket)},
\end{align}
\endgroup
with $\llbracket.\rrbracket$ the Iverson brackets. For more generality, our loss $\mathcal{L}_{\text{scal}}$ maximizes the above class-wise metrics with:
\begin{equation}
	\mathcal{L}_{\text{scal}}(\hat{p}, p) = -\frac{1}{C}\sum_{c = 1}^{C} ({P_c(\hat{p}, p)}  + {R_c(\hat{p}, p)} + {S_c(\hat{p}, p)}).
\end{equation}
In practice, we optimize semantics $\mathcal{L}^{\text{sem}}_{\text{scal}}{=} \mathcal{L}_{\text{scal}}(\hat{y}, y)$ and geometry $\mathcal{L}^{\text{geo}}_{\text{scal}}{=} \mathcal{L}_{\text{scal}}(\hat{y}^{\text{geo}}, y^{\text{geo}})$, where $\{y, y^{\text{geo}}\}$ are semantic and geometric labels with respective predictions $\{\hat{y}, \hat{y}^{\text{geo}}\}$.
\subsubsection{Frustum Proportion Loss}
\label{sec:frustum_loss}
\begin{figure}
	\centering
	\includegraphics[width=.82\linewidth]{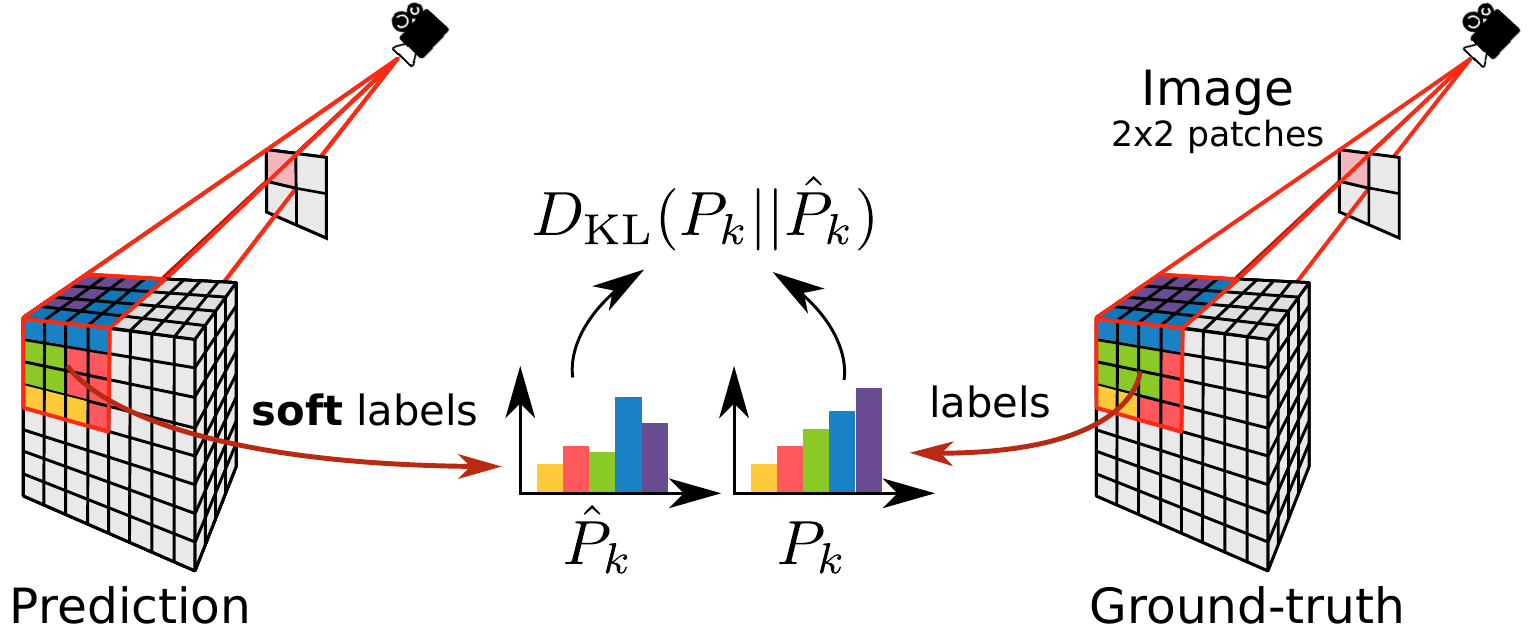}
	\caption{\textbf{Frustum Proportion Loss}. Considering an image divided into same-size 2D patches (here, $2{\times}2$), each corresponds to a 3D frustum in the scene, we align the predicted frustum class probabilities (e.g. $\hat{P}^k$) with the corresponding ground truth ($P^k$). This provides cues to the network for occlusions disambiguation.
	}
	\label{fig:frustum_loss}
\end{figure}
Disambiguation of occlusions is impossible from a single viewpoint and we observe that occluded voxels tend to be predicted as part of the object that shadows them. To mitigate this effect, we propose a Frustum Proportion Loss that explicitly optimizes the class distribution in a frustum. 

As illustrated in~\cref{fig:frustum_loss}, rather than optimizing the camera frustum distribution, we divide the
input image into $\ell{\times}\ell$ local patches of equal size and apply our loss on each local frustum (defined as the union of the individual pixels frustum in the patch). 
Intuitively, aligning the frustums distributions provide additional cues to the network on the scene visible and \textit{occluded} structure, giving a sense of what is likely to be occluded (eg. cars are likely to occlude road).

Given a frustum $k$, we compute $P_k$ the ground truth class distribution of voxels in $k$, and $P_{k,c}$ the proportion of class $c$ in $k$. Let $\hat{P}_k$ and $\hat{P}_{k,c}$ be their soft predicted counterparts, obtained from summing per-class predicted probabilities. To enforce consistency, we compute $\mathcal{L}_{\text{fp}}$ as the sum of local frustums Kullback-Leibler (KL) divergence:
\begin{equation}
	\mathcal{L}_{\text{fp}} {=} \sum_{k=1}^{\ell^2} D_{\text{KL}}(P_k || \hat{P}_k)\ {=} \sum_{k=1}^{\ell^2} \sum_{c \in C_k} P_k(c) \log \frac{P_k(c)}{\hat{P}_k(c)}\,.
	\label{eq:frustum_loss}
\end{equation}
Note the use of $C_k$ instead of $C$. Indeed, frustums include small scene portions where some classes may be missing, making KL locally \textit{undefined}. Instead, we compute the KL on $C_k$, the ground truth classes that exist in the frustum $k$.
\subsection{Training strategy}
MonoScene is trained end-to-end from scratch by optimizing our 4 losses and the standard cross-entropy ($\mathcal{L}_{\text{ce}}$):
\begin{equation}
	\mathcal{L}_\text{total} = \mathcal{L}_{\text{ce}}
	+ \mathcal{L}_{\text{rel}}
	+ \mathcal{L}^{\text{sem}}_{\text{scal}}
	+ \mathcal{L}^{\text{geo}}_{\text{scal}}
	+ \mathcal{L}_{\text{fp}}\,.
\end{equation}
Because real-world data comes with sparse ground truth $y$ due to occlusions, the losses are computed only where $y$ is defined~\cite{song2016ssc, SATNet, sscsurvey}. 
Ground truths $y^\text{geo}$ and $A^m$, for $\mathcal{L}^{\text{geo}}_{\text{scal}}$ and $\mathcal{L}_{\text{rel}}$, respectively, are simply obtained from $y$.
We employ class weighting for $\mathcal{L}_{\text{ce}}$ following~\cite{roldao2020lmscnet, 3DSketch}. 

\section{Experiments}
We evaluate MonoScene on popular real-world SSC datasets being, indoor NYUv2~\cite{NYUv2} and outdoor SemanticKitti~\cite{behley2019iccv}. 
Because we first address 3D SSC from a 2D image, we detail our non-trivial adaptation of recent SSC baselines~\cite{3DSketch,Li2020aicnet,roldao2020lmscnet,yan2021JS3CNet} (\cref{sec:exp_baselines}) and then detail our performance (\cref{sec:exp_perf}) and ablations (\cref{sec:exp_ablations}).\\

\condenseparagraph{Datasets.}NYUv2~\cite{NYUv2} has 1449 Kinect captured indoor scenes, encoded as 240x144x240 voxel grids labeled with 13 classes (11 semantics, 1 free, 1 unknown). The input RGBD is 640x480.
Similar to~\cite{3DSketch, Li2020aicnet, Li2019ddr, SATNet} we use 795/654 train/test splits and evaluate on the test set at the scale 1:4.

SemanticKITTI~\cite{behley2019iccv} holds outdoor Lidar scans voxelized as 256x256x32 grid of 0.2m voxels, labeled with 21 classes (19 semantics, 1 free, 1 unknown). 
We use RGB image of cam2 of size 1226x370, left cropped to 1220×370.
We use the official 3834/815 train/val splits and always evaluate at full scale (\ie 1:1). Main results are from the hidden test set (online server), and ablations are from the validation set.\\

\condenseparagraph{Training setup.}Unless otherwise mentioned, we use FLoSP at scales (1,2,4,8), 4 supervised relations for 3D~CRP (\ie $n{=}4$, with $\mathcal{L}_{rel}$), and $\ell{\times}\ell{=}8{\times}8$ frustums for $\mathcal{L}_{\text{fp}}$. The 3D UNet input is 60x36x60 (1:4) for NYUv2 and 128x128x16 (1:2) for Sem.KITTI due to memory reason. The output of Sem.KITTI is upscaled to 1:1 with a deconv layer in the completion head. Details in~\cref{sec:supp-architecture}.
We train 30 epochs with an AdamW~\cite{adamW} optimizer, a batch size of 4 and a weight decay of $1\mathrm{e}{\text{-}4}$. The learning rate is $1\mathrm{e}{\text{-}4}$, divided by 10 at epoch 20/25 for NYUv2/SemanticKITTI. \\

\condenseparagraph{Metrics.}Following common practices, we report the intersection over union~(IoU) of occupied voxels, regardless of their semantic class, for the scene completion~(SC) task and the mean IoU~(mIoU) of all semantic classes for the SSC task. Note the strong interaction between IoU and mIoU since better geometry estimation (\ie high IoU) can be achieved by invalidating semantic labels (\ie low mIoU).

As mentioned in~\cite{sscsurvey}, the training and evaluation practices differ for indoor 
(with metrics evaluated only on \textit{observed surfaces} and \textit{occluded} voxels)
or outdoor settings (evaluated on \textit{all} voxels) due to the different depth/Lidar sparsity. To cope with both settings, we use the harder metrics on \textit{all} voxels. We subsequently retrained \textit{all} baselines.
\begin{table*}
	\begin{subtable}{1.0\textwidth}
		\scriptsize
		\setlength{\tabcolsep}{0.0045\linewidth}
		\captionsetup{font=scriptsize}
		\newcommand{\nyuclassfreq}[1]{{~\tiny(\nyufreq{#1}\%)}}  %
		\centering
		\begin{tabular}{l|c|c|c c c c c c c c c c c|c}
			\toprule
			& & SC & \multicolumn{12}{c}{SSC} \\[-0.9em]
			Method
			& Input
			& {IoU}
			& \rotatebox{90}{\parbox{2cm}{\textcolor{ceiling}{$\blacksquare$} ceiling\nyuclassfreq{ceiling}}} 
			& \rotatebox{90}{\textcolor{floor}{$\blacksquare$} floor\nyuclassfreq{floor}}
			& \rotatebox{90}{\textcolor{wall}{$\blacksquare$} wall\nyuclassfreq{wall}} 
			& \rotatebox{90}{\textcolor{window}{$\blacksquare$} window\nyuclassfreq{window}} 
			& \rotatebox{90}{\textcolor{chair}{$\blacksquare$} chair\nyuclassfreq{chair}} 
			& \rotatebox{90}{\textcolor{bed}{$\blacksquare$} bed\nyuclassfreq{bed}} 
			& \rotatebox{90}{\textcolor{sofa}{$\blacksquare$} sofa\nyuclassfreq{sofa}} 
			& \rotatebox{90}{\textcolor{table}{$\blacksquare$} table\nyuclassfreq{table}} 
			& \rotatebox{90}{\textcolor{tvs}{$\blacksquare$} tvs\nyuclassfreq{tvs}} 
			& \rotatebox{90}{\textcolor{furniture}{$\blacksquare$} furniture\nyuclassfreq{furniture}} 
			& \rotatebox{90}{\textcolor{objects}{$\blacksquare$} objects\nyuclassfreq{objects}} 
			& mIoU\\
			\midrule
			LMSCNet$^\text{rgb}$~\cite{roldao2020lmscnet} & $\hat{x}^{\text{occ}}$ & 33.93 & 4.49 & 88.41 & 4.63 & 0.25 & 3.94 & 32.03 & 15.44 & 6.57 & 0.02 & 14.51 & 4.39 & 15.88 
			\\
			AICNet$^\text{rgb}$~\cite{Li2020aicnet} & $x^{\text{rgb}}$, $\hat{x}^{\text{depth}}$ & 30.03 & 7.58 & 82.97 & 9.15 & 0.05 & 6.93 & 35.87 & 22.92 & 11.11 & 0.71 & 15.90 & 6.45 & 18.15 
			\\
			3DSketch$^\text{rgb}$~\cite{3DSketch}& $x^{\text{rgb}}$, $\hat{x}^{\text{TSDF}}$  & \underline{38.64} & \underline{8.53} & \underline{90.45} &	\underline{9.94} & \underline{5.67} & \underline{10.64} & \underline{42.29} & \underline{29.21} & \underline{13.88} & \underline{9.38} & \underline{23.83} & \underline{8.19} & \underline{22.91}
			\\
			MonoScene (ours) & $x^{\text{rgb}}$ & \textbf{42.51} & \textbf{8.89} & \textbf{93.50} & \textbf{12.06} & \textbf{12.57} & \textbf{13.72} & \textbf{48.19} & \textbf{36.11} & \textbf{15.13} & \textbf{15.22} & \textbf{27.96} & \textbf{12.94} & \textbf{26.94}\\
			\bottomrule
		\end{tabular}
		\caption{NYUv2 (test set)}
		\label{tab:perf_nyu}
	\end{subtable}

	\begin{subtable}{1.0\textwidth}
		\scriptsize
		\captionsetup{font=scriptsize}
		\setlength{\tabcolsep}{0.004\linewidth}
		\newcommand{\classfreq}[1]{{~\tiny(\semkitfreq{#1}\%)}}  %
		\centering
		\begin{tabular}{l|c|c|c c c c c c c c c c c c c c c c c c c|c}
			\toprule
			& & SC & \multicolumn{20}{c}{SSC} \\
			Method
			& SSC Input
			& {IoU}
			& \rotatebox{90}{\textcolor{road}{$\blacksquare$} road\classfreq{road}} 
			& \rotatebox{90}{\textcolor{sidewalk}{$\blacksquare$} sidewalk\classfreq{sidewalk}}
			& \rotatebox{90}{\textcolor{parking}{$\blacksquare$} parking\classfreq{parking}} 
			& \rotatebox{90}{\textcolor{other-ground}{$\blacksquare$} other-grnd\classfreq{otherground}} 
			& \rotatebox{90}{\textcolor{building}{$\blacksquare$} building\classfreq{building}} 
			& \rotatebox{90}{\textcolor{car}{$\blacksquare$} car\classfreq{car}} 
			& \rotatebox{90}{\textcolor{truck}{$\blacksquare$} truck\classfreq{truck}} 
			& \rotatebox{90}{\textcolor{bicycle}{$\blacksquare$} bicycle\classfreq{bicycle}} 
			& \rotatebox{90}{\textcolor{motorcycle}{$\blacksquare$} motorcycle\classfreq{motorcycle}} 
			& \rotatebox{90}{\textcolor{other-vehicle}{$\blacksquare$} other-veh.\classfreq{othervehicle}} 
			& \rotatebox{90}{\textcolor{vegetation}{$\blacksquare$} vegetation\classfreq{vegetation}} 
			& \rotatebox{90}{\textcolor{trunk}{$\blacksquare$} trunk\classfreq{trunk}} 
			& \rotatebox{90}{\textcolor{terrain}{$\blacksquare$} terrain\classfreq{terrain}} 
			& \rotatebox{90}{\textcolor{person}{$\blacksquare$} person\classfreq{person}} 
			& \rotatebox{90}{\textcolor{bicyclist}{$\blacksquare$} bicyclist\classfreq{bicyclist}} 
			& \rotatebox{90}{\textcolor{motorcyclist}{$\blacksquare$} motorcyclist.\classfreq{motorcyclist}} 
			& \rotatebox{90}{\textcolor{fence}{$\blacksquare$} fence\classfreq{fence}} 
			& \rotatebox{90}{\textcolor{pole}{$\blacksquare$} pole\classfreq{pole}} 
			& \rotatebox{90}{\textcolor{traffic-sign}{$\blacksquare$} traf.-sign\classfreq{trafficsign}} 
			& mIoU\\
			\midrule
			LMSCNet$^\text{rgb}$~\cite{roldao2020lmscnet} & $\hat{x}^{\text{occ}}$ &  31.38 & 46.70 & 19.50 & 13.50 & \underline{3.10} & 10.30 & 14.30 & 0.30 & 0.00 & 0.00 & 0.00 & 10.80 & 0.00 & 10.40 & 0.00 & 0.00 & 0.00 & 5.40 & 0.00 & 0.00 & 7.07  \\
			3DSketch$^\text{rgb}$~\cite{3DSketch} & $x^{\text{rgb}}$,$\hat{x}^{\text{TSDF}}$ & 26.85 & 37.70 & 19.80 & 0.00 & 0.00 & 12.10 & 17.10 & 0.00 & 0.00 & 0.00 & 0.00 & 12.10 & 0.00 & \underline{16.10} & 0.00 & 0.00 & 0.00 & 3.40 & 0.00 & 0.00 & 6.23  \\
			AICNet$^\text{rgb}$~\cite{Li2020aicnet} & $x^{\text{rgb}}$,$\hat{x}^{\text{depth}}$ & 23.93	& 39.30	& 18.30 & 19.80 & 1.60 & 9.60	& 15.30	& 0.70	& 0.00	& 0.00	& 0.00	& 9.60	& 1.90	& 13.50	& 0.00	& 0.00	& 0.00	& 5.00	& 0.10	& 0.00	& 7.09 \\
			*JS3C-Net$^\text{rgb}$~\cite{yan2021JS3CNet} & $\hat{x}^{\text{pts}}$& \underline{34.00} & \underline{47.30} & \underline{21.70} & \underline{19.90} & 2.80 & \underline{12.70} & \textbf{20.10} & \underline{0.80} & 0.00 & 0.00 & \underline{4.10} & \underline{14.20} & \textbf{3.10} & 12.40 & 0.00 & \underline{0.20} & \underline{0.20} & \underline{8.70} & \underline{1.90} & \underline{0.30} & \underline{8.97}  \\
			MonoScene (ours) & $x^{\text{rgb}}$ & \textbf{34.16} & \textbf{54.70} & \textbf{27.10} & \textbf{24.80} & \textbf{5.70} & \textbf{14.40} & \underline{18.80} & \textbf{3.30} & \textbf{0.50} & \textbf{0.70} & \textbf{4.40} & \textbf{14.90} & \underline{2.40} & \textbf{19.50} & \textbf{1.00} & \textbf{1.40} & \textbf{0.40} & \textbf{11.10} & \textbf{3.30} & \textbf{2.10} & \textbf{11.08} \\
			\bottomrule
		\end{tabular}\\
		{\scriptsize* Uses pretrained semantic segmentation network.}
		\caption{Semantic KITTI (hidden test set)}
		\label{tab:perf_semkitti}
	\end{subtable}
	\caption{\textbf{Performance on \subref{tab:perf_nyu} NYUv2~\cite{NYUv2} and \subref{tab:perf_semkitti} SemanticKITTI~\cite{behley2019iccv}.} We report the performance on semantic scene completion (SSC~-~mIoU) and scene completion (SC~-~IoU) for RGB-inferred baselines and our method. Despite the various indoor and outdoor setups, we significantly outperform other RGB-inferred baselines, in both mIoU and IoU.}
	\label{tab:perf}
\end{table*}
\subsection{Baselines}\label{sec:exp_baselines}
We consider 4 main SSC baselines among the best opensource ones available -- selecting two indoor-designed methods, 3DSketch~\cite{3DSketch} and AICNet~\cite{Li2020aicnet}, and two outdoor-designed, LMSCNet~\cite{roldao2020lmscnet} and JS3CNet~\cite{yan2021JS3CNet}. We also locally compare against S3CNet~\cite{Cheng2020S3CNetAS}, Local-DIFs~\cite{localDif}, CoReNet~\cite{Popov2020CoReNetC3}.
We evaluate baselines in their \emph{3D-input} version and \emph{main} baselines also in an \emph{RGB-inferred} version.

\condenseparagraph{RGB-inferred baselines.}
Unlike us, all baselines need a 3D input \eg occupancy grid, point cloud or depth map, giving them an unfair geometric advantage. 
For fair comparison, we adapt main baselines to infer their 3D inputs directly from the 2D image ($x^\text{rgb}$) -- relying on the best found methods --, coined as `RGB-inferred', denoted with a superscript, \eg AICNet$^\text{rgb}$. Note that baselines are unchanged. Inferred 3D inputs are denoted with a hat, eg. $\hat{x}^{\text{depth}}$.

We use the pretrained AdaBin~\cite{bhat_adabins_2021} to infer a depth map ($\hat{x}^{\text{depth}}$) serving as input for AICNet$^\text{rgb}$.
Using intrinsic calibration, we further converted depth to TSDF ($\hat{x}^{\text{TSDF}}$) with~\cite{zeng20163dmatch} for 3DSketch$^\text{rgb}$ input, and unproject depth to get a point cloud ($\hat{x}^{\text{pts}}$) directly used as input for JS3CNet$^\text{rgb}$ or discretized as occupancy grid ($\hat{x}^{\text{occ}}$) input for LMSCNet$^\text{rgb}$.
For training \textit{only}, JS3CNet$^\text{rgb}$ also requires a semantic point cloud ($\hat{x}^{\text{sem pts}}$), obtained by augmenting $\hat{x}^{\text{pts}}$ with 2D semantic labels from a pretrained network~\cite{zhu2019labelrelaxation}.
\subsection{Performance}\label{sec:exp_perf}
\subsubsection{Evaluation}
\cref{tab:perf} reports performance of MonoScene and RGB-inferred baselines for NYUv2 (test set) and SemanticKITTI official benchmark (hidden test set).
The low numbers for all methods advocate for task complexity.

On both datasets we outperform all methods by a significant mIoU margin of +4.03 on NYUv2 (\cref{tab:perf_nyu}) and +2.11 on SemanticKITTI (\cref{tab:perf_semkitti}). 
Importantly, the IoU is improved or on par (+3.87 and +0.16) which demonstrates our network captures the scene geometry while avoiding naively increasing the mIoU by lowering the IoU.
On individual classes, MonoScene performs either best or second, excelling on large structural classes for both datasets (\eg floor, wall ; road, building). 
On SemanticKITTI we get outperformed mostly on small moving objects classes (car, motorcycle, person, etc.) which we ascribe to the aggregation of moving objects in the ground truth, highlighted in~\cite{sscsurvey,localDif}. This forces to predict the individual object's motion which we argue is eased when using a 3D input.\\
\begin{figure*}
	\begin{subfigure}{\linewidth}		
		\centering
		\footnotesize
		\renewcommand{\arraystretch}{0.0}
		\setlength{\tabcolsep}{0.003\textwidth}
		\newcolumntype{P}[1]{>{\centering\arraybackslash}m{#1}}
		\captionsetup{font=scriptsize}
		\begin{tabular}{P{0.11\textwidth} P{0.13\textwidth} P{0.13\textwidth} P{0.13\textwidth} P{0.13\textwidth} P{0.13\textwidth}}
			Input & LMSCNet$^\text{rgb}$~\cite{roldao2020lmscnet} & AICNet$^\text{rgb}$~\cite{Li2020aicnet} & 3DSketch$^\text{rgb}$~\cite{3DSketch} & MonoScene \scriptsize{(ours)} & Ground Truth
			\\[-0.1em]
			\includegraphics[width=\linewidth]{imgs/nyu_qualitative/NYU1250_0000_2_color.jpg} & 
			\includegraphics[width=\linewidth]{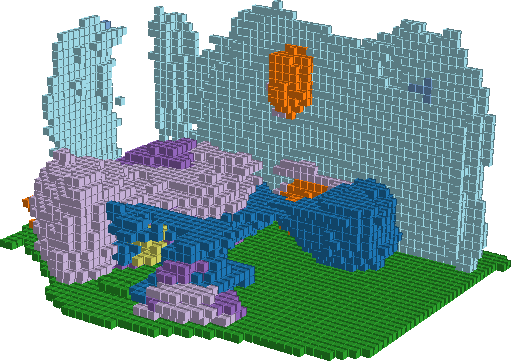} &
			\includegraphics[width=\linewidth]{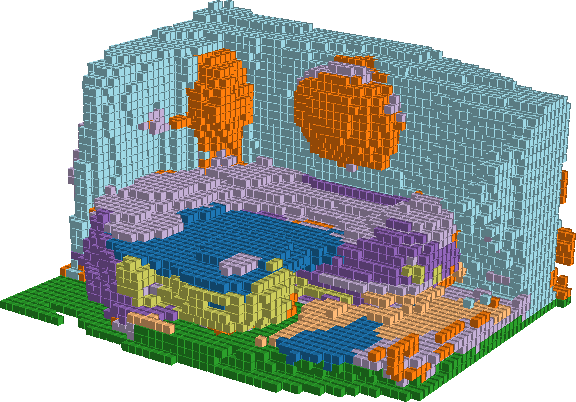} &
			\includegraphics[width=\linewidth]{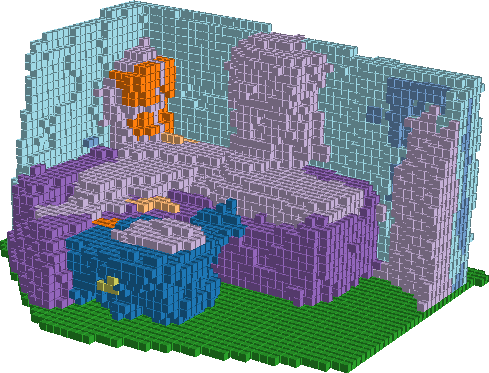} &
			\includegraphics[width=\linewidth]{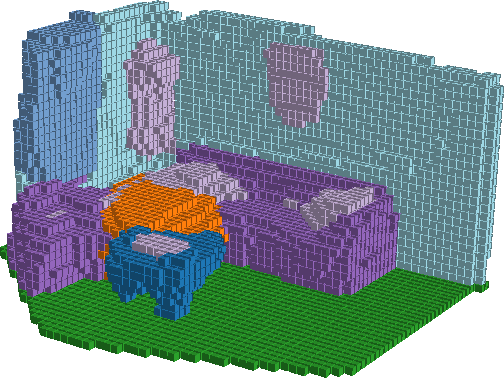} &
			\includegraphics[width=\linewidth]{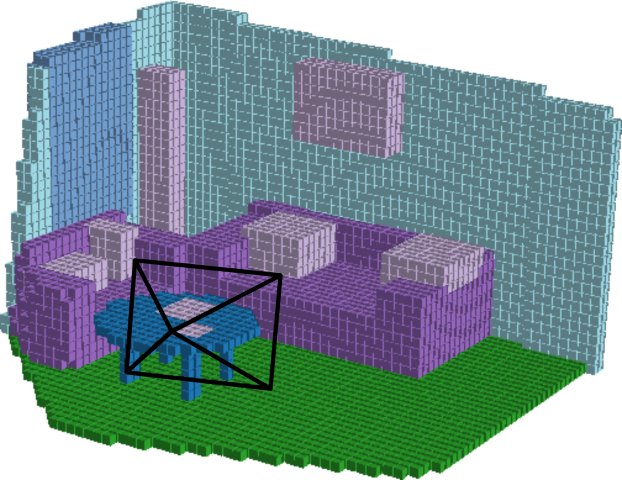}
			\\[-0.3em]
			\includegraphics[width=\linewidth]{imgs/nyu_qualitative/NYU0994_0000_2_color.jpg} & 
			\includegraphics[width=\linewidth]{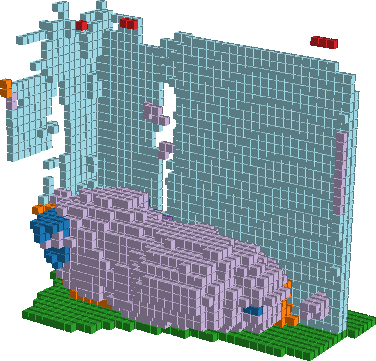} &
			\includegraphics[width=\linewidth]{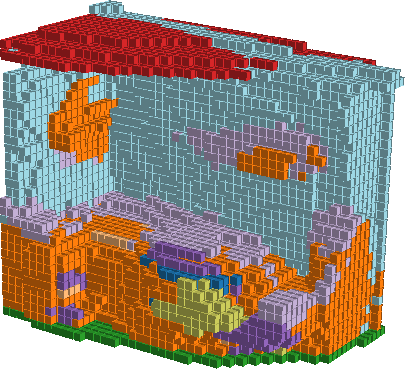} &
			\includegraphics[width=\linewidth]{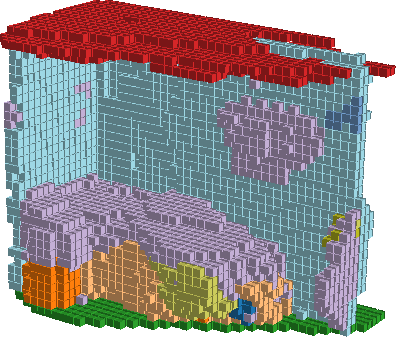} &
			\includegraphics[width=\linewidth]{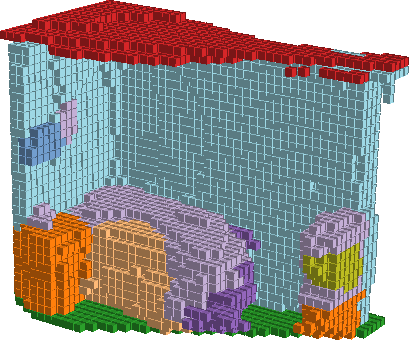} &
			\includegraphics[width=\linewidth]{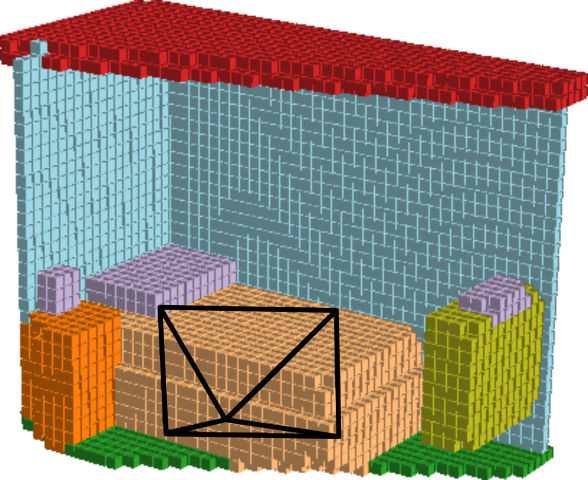}
			\\[-0.3em] 
			\includegraphics[width=\linewidth]{imgs/nyu_qualitative/NYU0970_0000_0_color.jpg} & 
			\includegraphics[width=\linewidth]{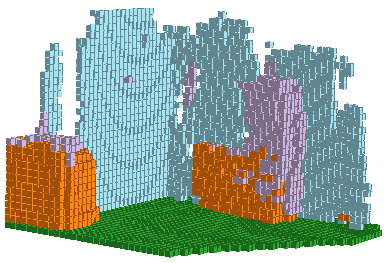} &
			\includegraphics[width=\linewidth]{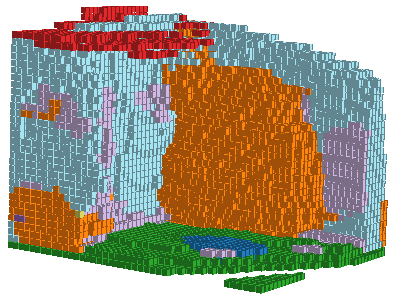} &
			\includegraphics[width=\linewidth]{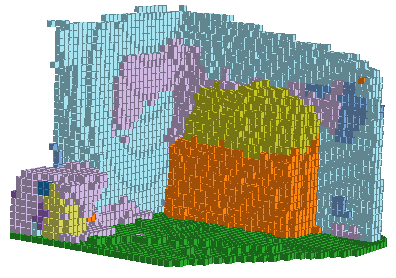} &
			\includegraphics[width=\linewidth]{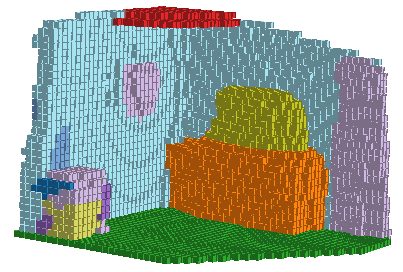} &
			\includegraphics[width=\linewidth]{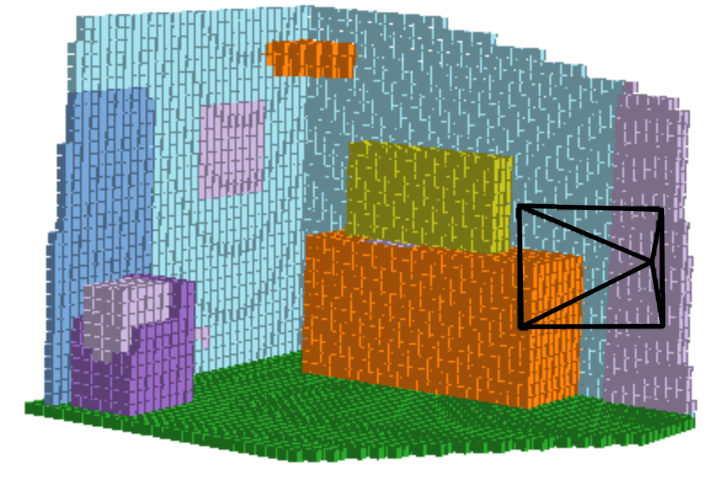}
			\\
			\multicolumn{6}{c}{
				\scriptsize
				\textcolor{ceiling}{$\blacksquare$}ceiling~
				\textcolor{floor}{$\blacksquare$}floor~
				\textcolor{wall}{$\blacksquare$}wall~
				\textcolor{window}{$\blacksquare$}window~
				\textcolor{chair}{$\blacksquare$}chair~
				\textcolor{sofa}{$\blacksquare$}sofa~
				\textcolor{table}{$\blacksquare$}table~
				\textcolor{tvs}{$\blacksquare$}tvs~
				\textcolor{furniture}{$\blacksquare$}furniture~
				\textcolor{objects}{$\blacksquare$}objects
			}
		\end{tabular}
		\caption{\textbf{NYUv2\cite{NYUv2} (test set).}\\[0.5em]}
		\label{fig:qualitative_NYU}
	\end{subfigure}
	\begin{subfigure}{\linewidth}	
		\centering
		\newcolumntype{P}[1]{>{\centering\arraybackslash}m{#1}}
		\setlength{\tabcolsep}{0.001\textwidth}
		\renewcommand{\arraystretch}{0.8}
		\footnotesize
		\captionsetup{font=scriptsize}
		\begin{tabular}{P{0.157\textwidth} P{0.157\textwidth} P{0.157\textwidth} P{0.157\textwidth} P{0.157\textwidth} P{0.157\textwidth}}		
			Input & AICNet$^\text{rgb}$~\cite{Li2020aicnet} & LMSCNet$^\text{rgb}$~\cite{roldao2020lmscnet}  & JS3CNet$^\text{rgb}$~\cite{yan2021JS3CNet}  & \OurMethod\ \scriptsize{(ours)} & Ground Truth
			\\[-0.1em]
			\includegraphics[width=.9\linewidth]{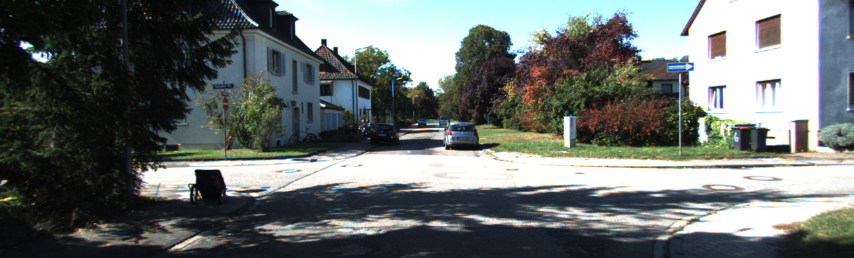} &  
			\includegraphics[width=.9\linewidth]{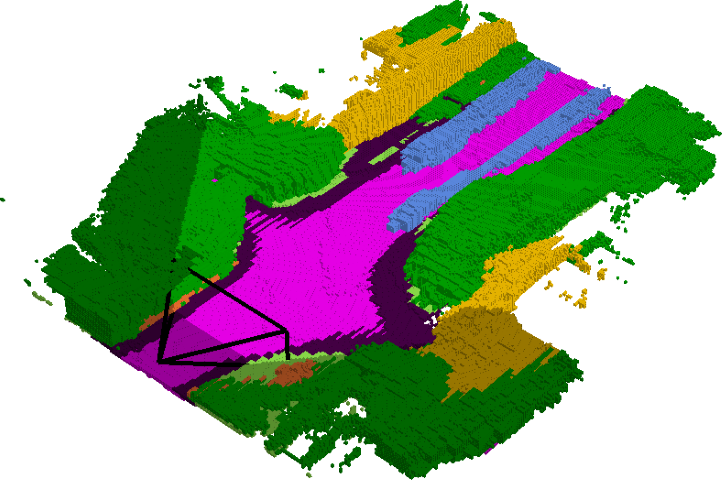} & 
			\includegraphics[width=.9\linewidth]{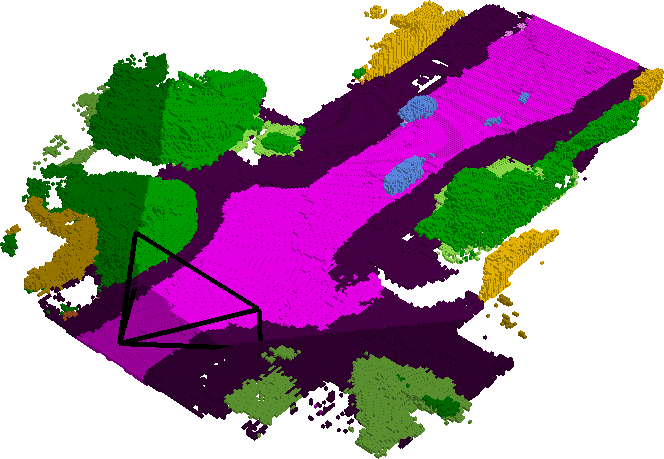} & 
			\includegraphics[width=.9\linewidth]{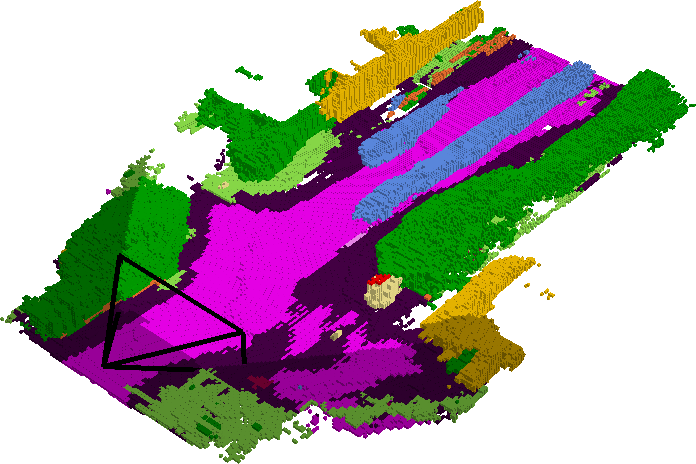} & 
			\includegraphics[width=.9\linewidth]{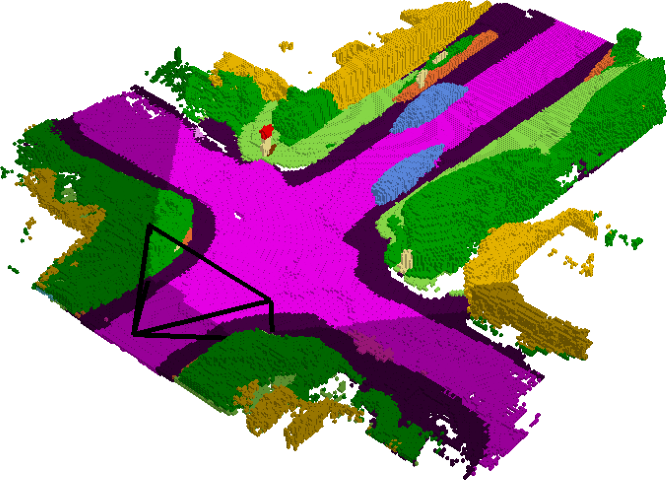} & 
			\includegraphics[width=\linewidth]{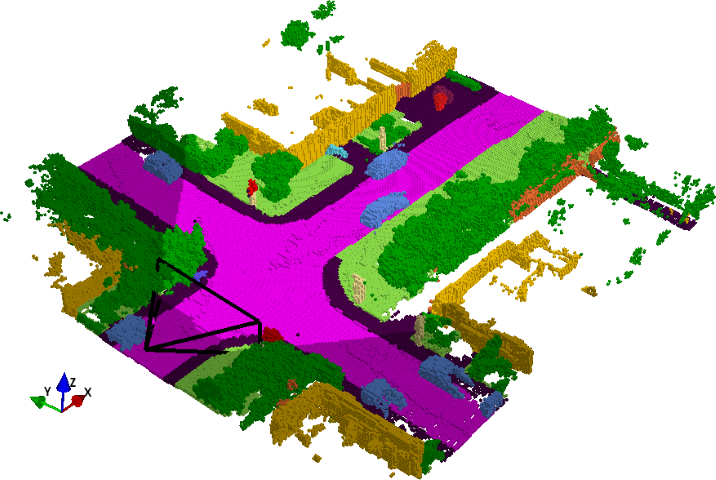} 
			\\[-0.6em]
			\includegraphics[width=\linewidth]{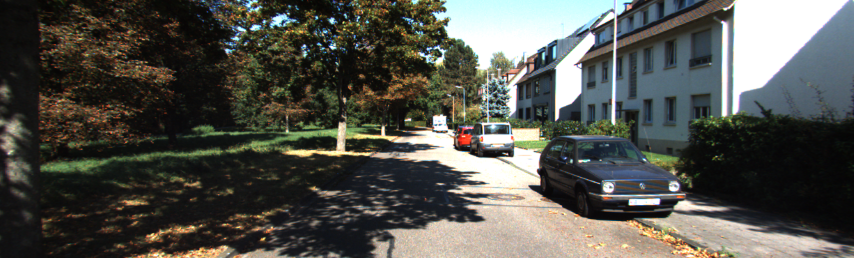} & 
			\includegraphics[width=\linewidth]{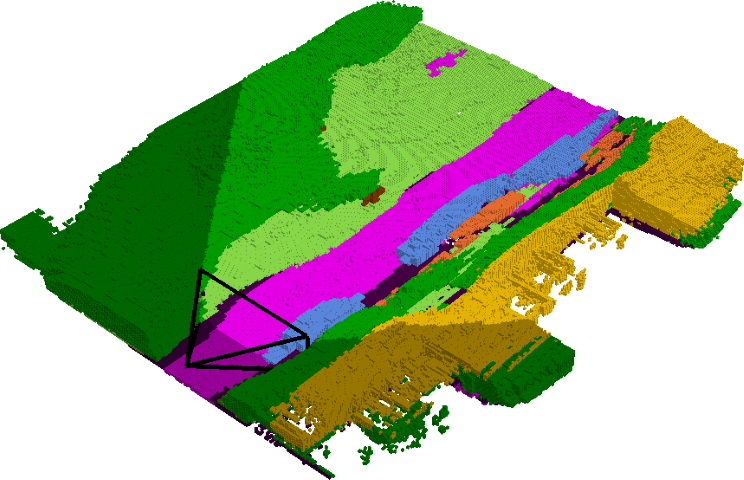} & 
			\includegraphics[width=\linewidth]{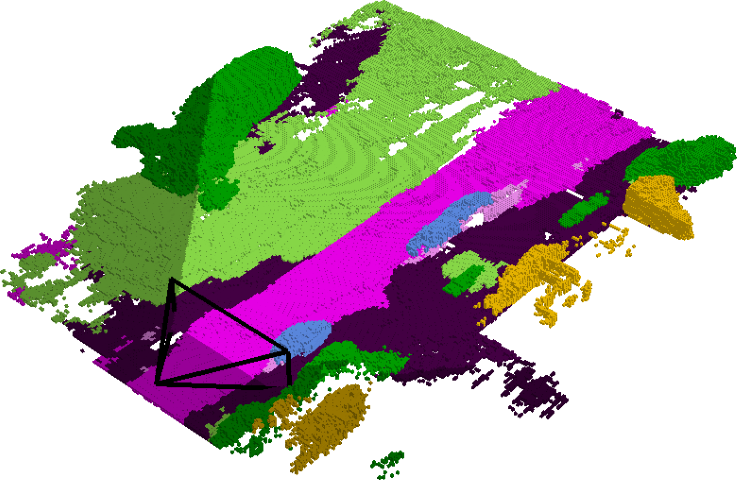} & 
			\includegraphics[width=\linewidth]{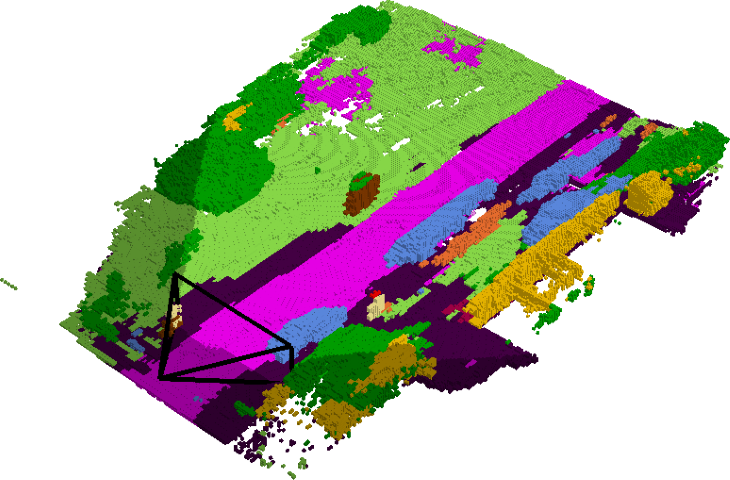} & 
			\includegraphics[width=\linewidth]{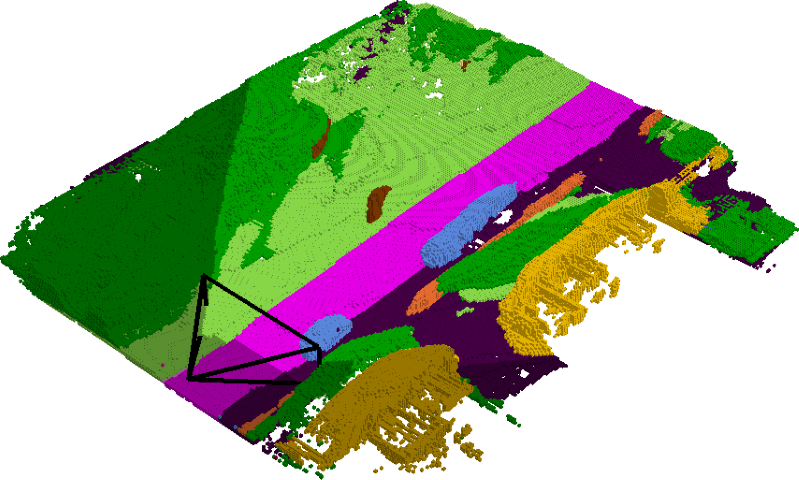} & 
			\includegraphics[width=\linewidth]{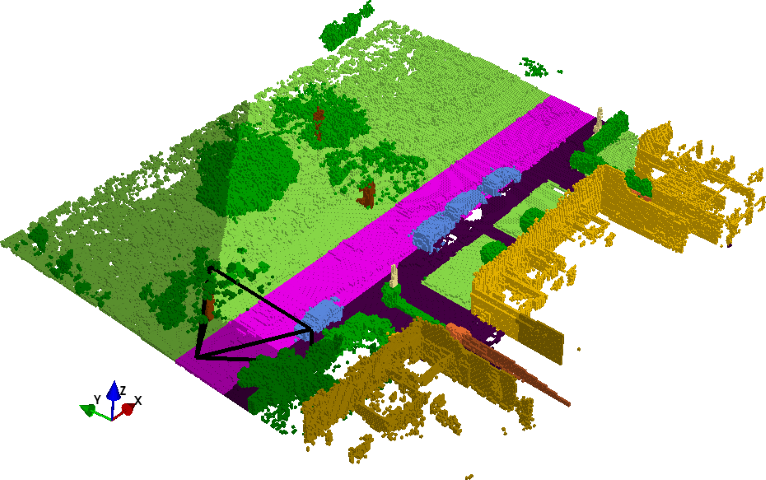} 
			\\[-0.6em]
			\includegraphics[width=\linewidth]{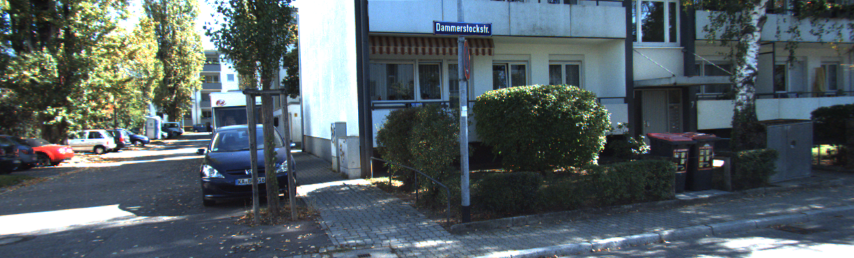} & 
			\includegraphics[width=0.8\linewidth]{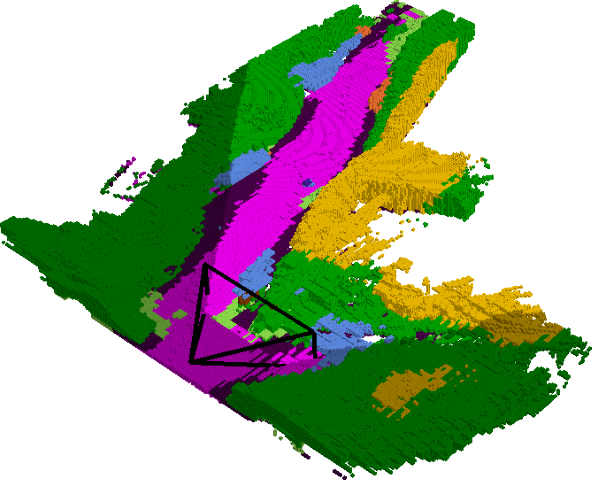} & 
			\includegraphics[width=0.7\linewidth]{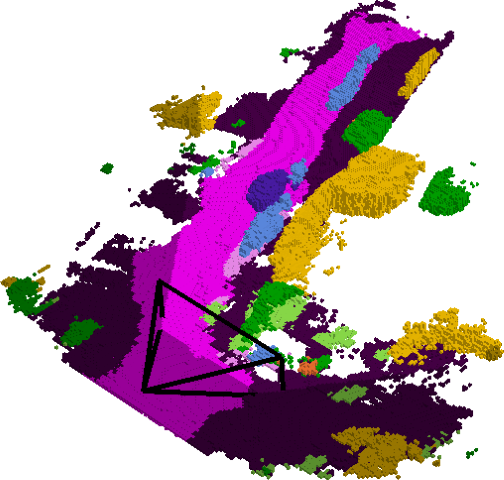} & 
			\includegraphics[width=0.7\linewidth]{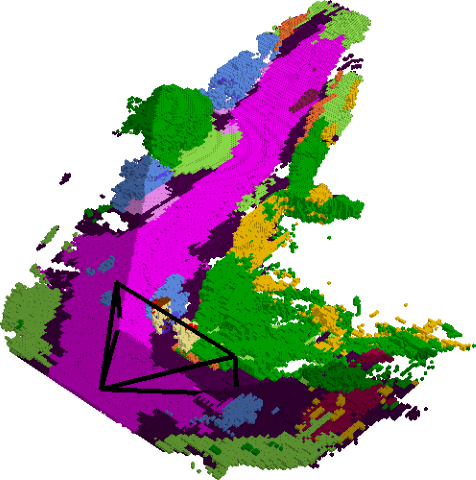} & 
			\includegraphics[width=0.8\linewidth]{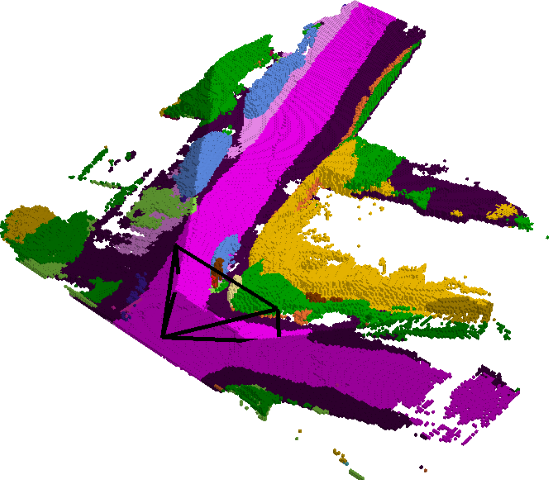} & 
			\includegraphics[width=\linewidth]{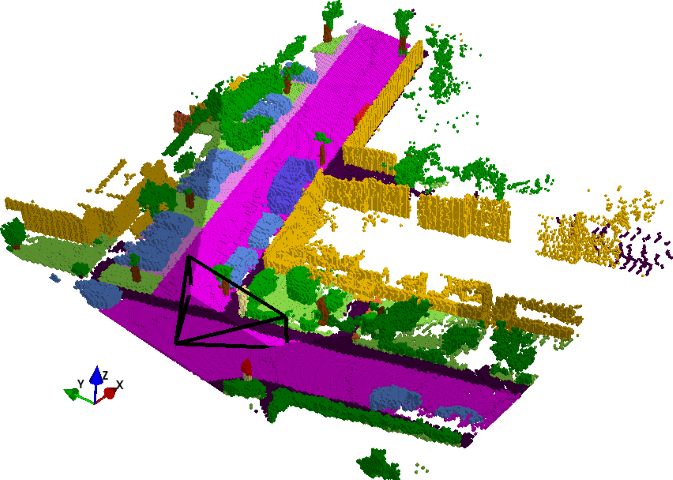} 
			\\		
			\multicolumn{6}{c}{
				\scriptsize
				\textcolor{bicycle}{$\blacksquare$}bicycle~
				\textcolor{car}{$\blacksquare$}car~
				\textcolor{motorcycle}{$\blacksquare$}motorcycle~
				\textcolor{truck}{$\blacksquare$}truck~
				\textcolor{other-vehicle}{$\blacksquare$}other vehicle~
				\textcolor{person}{$\blacksquare$}person~
				\textcolor{bicyclist}{$\blacksquare$}bicyclist~
				\textcolor{motorcyclist}{$\blacksquare$}motorcyclist~
				\textcolor{road}{$\blacksquare$}road~
				\textcolor{parking}{$\blacksquare$}parking~}\\
			\multicolumn{6}{c}{
				\scriptsize
				\textcolor{sidewalk}{$\blacksquare$}sidewalk~
				\textcolor{other-ground}{$\blacksquare$}other ground~
				\textcolor{building}{$\blacksquare$}building~
				\textcolor{fence}{$\blacksquare$}fence~
				\textcolor{vegetation}{$\blacksquare$}vegetation~
				\textcolor{trunk}{$\blacksquare$}trunk~
				\textcolor{terrain}{$\blacksquare$}terrain~
				\textcolor{pole}{$\blacksquare$}pole~
				\textcolor{traffic-sign}{$\blacksquare$}traffic sign			
			}
		\end{tabular}		
		\caption{\textbf{SemanticKITTI~\cite{behley2019iccv} (val set).}}
		\label{fig:qualitative_kitti}
	\end{subfigure}
	\caption{\textbf{Outputs on~\subref{fig:qualitative_NYU} NYUv2~\cite{NYUv2} and~\subref{fig:qualitative_kitti} SemanticKITTI~\cite{behley2019iccv}.} 
		In both, the input is shown left and the camera viewing frustum is shown in the ground truth (rightmost) with darker colors being parts of scenes \emph{unseen} by the image in~\subref{fig:qualitative_kitti}.
		MonoScene better captures the scene layout on both datasets. On indoor scene~\subref{fig:qualitative_NYU}, it reconstructs thin objects like table legs (row 1), painting and tv (row 2), while in outdoor~\subref{fig:qualitative_kitti}, it better estimates occluded geometry \eg car (row 1--3) and better hallucinates the scenery beyond the field of view (row 1--4).}
	\label{fig:qualitative}
\end{figure*} 

\condenseparagraph{Qualitative.}We compare our SSC outputs in Fig.~\ref{fig:qualitative} showing the input image (leftmost column) and its corresponding camera frustum in ground truth (rightmost).
Notice the noisy labels in NYUv2 having missing objects (\eg windows, rows~2; ceiling, row~3), and in SemanticKITTI having sparse geometry (\eg holes in buildings, rows~1--3).

On indoor scenes (NYUv2, \cref{fig:qualitative_NYU}), all methods correctly capture global scene layouts though only MonoScene recovers thin elements as table legs and cushions (row~1), or the painting frame and properly sized TV (row~2). 

On complex cluttered outdoor scenes (SemanticKITTI, \cref{fig:qualitative_kitti}), compared to baselines, MonoScene evidently captures better the scene layout, \eg cross-roads (rows~1,3).
It also infers finer occluded geometry which is apparent with cars in rows 1--3 having better shapes.
Interestingly, despite a narrow camera field of view (FOV) with respect to the scene, MonoScene properly hallucinates scenery not imaged, \ie \textit{outside} of the camera FOV (darker voxels). This is striking in rows 3,4 where the bottom part of the scene is reasonably guessed, though not in the viewing frustum. \cref{sec:supp-addresults-kitti} provides in-/out-FOV performance details.
\begin{table}
	\centering
	\begin{subtable}{0.48\linewidth}
		\scriptsize
		\setlength{\tabcolsep}{0.009\linewidth}
		\captionsetup{font=scriptsize}
		\newcommand{\nyuclassfreq}[1]{{~\tiny(\nyufreq{#1}\%)}}  %
		\centering
		\begin{tabular}{l|c|c c}
			\toprule
			Method
			& Input
			& {IoU}	
			& mIoU\\
			\midrule
			\textbf{2.5/3D} &&&\\[-0.1em]
			LMSCNet~\cite{roldao2020lmscnet} & $x^{\text{occ}}$ & \underline{44.1} & 20.4
			\\
			AICNet~\cite{Li2020aicnet} & $x^{\text{rgb}}$, $x^{\text{depth}}$ & 43.8 & 23.8
			\\
			3DSketch~\cite{3DSketch}& $x^{\text{rgb}}$, $x^{\text{TSDF}}$  & \textbf{49.5} & \textbf{29.2}
			\\
			\midrule
			\textbf{2D} &&&\\[-0.1em]
			MonoScene & $x^{\text{rgb}}$ & {42.5} & \underline{26.9}\\
			\bottomrule
		\end{tabular}
		\caption{NYUv2 (test set)}
		\label{tab:perf3d_nyu}
	\end{subtable}\hfill%
	\begin{subtable}{0.48\linewidth}
		\scriptsize
		\captionsetup{font=scriptsize}
		\setlength{\tabcolsep}{0.009\linewidth}
		\newcommand{\classfreq}[1]{{~\tiny(\semkitfreq{#1}\%)}}  %
		\centering
		\begin{tabular}{l|c|c c}
			\toprule
			Method & Input & {IoU} & mIoU\\
			\midrule
			\textbf{3D} & & &\\[-0.1em]
			LMSCNet~\cite{roldao2020lmscnet} & $x^{\text{occ}}$ & \underline{56.7} & 17.6	
			\\
			Local-DIFs~\cite{localDif} & $x^{\text{occ}}$ & \textbf{57.7} & 22.7
			\\
			JS3C-Net~\cite{yan2021JS3CNet} & $x^{\text{pts}}$ & 56.6 & \underline{23.8}
			\\
			S3CNet~\cite{Cheng2020S3CNetAS} & $x^{\text{occ}}$ & 45.6 & \textbf{29.5}
			\\
			\midrule
			\textbf{2D} &&&\\[-0.1em]
			MonoScene & $x^{\text{rgb}}$ & {34.2} & {11.1} \\
			\bottomrule
		\end{tabular}\\
		\caption{Sem.KITTI (hidden test set)}
		\label{tab:perf3d_semkitti}
	\end{subtable}
	\caption{\textbf{2.5/3D input baselines}. Despite a single RGB, MonoScene still outperforms the mIoU of \textit{some} indoor baselines.} 
	\label{tab:perf3d}
\end{table}
\subsubsection{Comparison against 2.5/3D-input baselines}
For completeness, we also compare with some original baselines (\ie using real 3D input) in~\cref{tab:perf3d}.
Despite the unfair setup since we use only RGB, in NYUv2 (\cref{tab:perf3d_nyu}) we still beat the recent LMSCNet and AICNet in mIoU by a comfortable margin (+6.48 and +3.17), but with a lower IoU (-1.6 and -1.26). Of note AICNet also uses RGB in addition to depth, showing our method excels at recovering geometry from image only. 
3DSketch, using RGB + TSDF, outperforms us on both mIoU and IoU showing the benefit of TSDF for SSC as mentioned in~\cite{sscsurvey}.
In SemanticKITTI (\cref{tab:perf3d_semkitti}), the baselines clearly surpass us in all metrics which relates both to the lidar-originated 3D input having a much wider horizontal FOV than the camera ($180^{\circ}$ vs $82^{\circ}$), and to the far more complex and cluttered outdoor scenes -- thus harder to reconstruct from a single image viewpoint.
\begin{table}
	\centering
	\setlength{\tabcolsep}{0.015\linewidth}
	\centering
	\footnotesize
	\begin{tabular}{l|cc|cc}
		\toprule
		& \multicolumn{2}{c}{NYUv2} & \multicolumn{2}{c}{SemanticKITTI} \\
		& IoU $\uparrow$ & mIoU $\uparrow$ & IoU $\uparrow$ & mIoU $\uparrow$ \\
		\midrule
		Ours &  \underline{42.51} \tiny$\pm$0.15 & \textbf{26.94} \tiny$\pm$0.10 & \textbf{37.12} \tiny$\pm$0.15 & \textbf{11.50} \tiny$\pm$0.14				
		\\
		Ours w/o FLoSP & 28.39 \tiny$\pm$0.53 & 14.11 \tiny$\pm$0.21 & 27.55 \tiny$\pm$0.87 & 4.78\tiny$\pm$0.10
		\\
		Ours w/o 3D CRP & 41.39	\tiny$\pm$0.08  & 26.27 \tiny$\pm$0.15 & 36.20  \tiny$\pm$0.19 & 10.96 \tiny$\pm$0.21
		\\
		Ours w/o $\mathcal{L}^{\text{sem}}_{\text{scal}}$ & \textbf{42.82} \tiny$\pm$0.22 & 25.33 \tiny$\pm$0.26 & \underline{36.78} \tiny$\pm$0.34 & 9.89 \tiny$\pm$0.11				
		\\
		Ours w/o $\mathcal{L}^{\text{geo}}_{\text{scal}}$  & 40.96 \tiny$\pm$0.28 & 26.34 \tiny$\pm$0.23 & 34.92 \tiny$\pm$0.34 & \underline{11.35} \tiny$\pm$0.22				
		\\
		Ours w/o $\mathcal{L}_{\text{fp}}$  & 41.90 \tiny$\pm$0.26 & \underline{26.37} \tiny$\pm$0.16 & 36.74 \tiny$\pm$0.33 & 11.11 \tiny$\pm$0.24  \\
		\bottomrule
	\end{tabular}
	\caption{\textbf{Architecture ablation.} Our components boost performance on NYUv2~\cite{NYUv2} (test set) and SemanticKitti~\cite{behley2019iccv} (val. set).}
	\label{tab:general_ablation}
\end{table}

The large 2D -- 2.5/3D gaps in \cref{tab:perf3d} partly result of the low depth accuracy. For example, on NYUv2/Sem.KITTI AdaBins~\cite{bhat_adabins_2021} gets 0.36/2.36m RMSE while voxels size is 0.08/0.2m. 
Furthermore, as we account for occluded voxels, our geometry is expectedly better.
This is assessed using MonoScene geometry as input of LMSCNet, which improves IoU/mIoU on Sem.KITTI val. set from 28.61/6.70 to 35.94/9.44.
Still, \cref{tab:general_ablation} shows that MonoScene predicts better geometry \textit{and} semantics reaching {37.12}/{11.50}.

\subsection{Ablation studies}
\label{sec:exp_ablations}
We ablate our MonoScene framework on both NYUv2 (test set) and SemanticKITTI (validation set), and report the average of 3 runs to account for training variability.

\condenseparagraph{Architectural components.}\cref{tab:general_ablation} shows that all components contribute to the best results.
For `w/o FLoSP', we instead interpolate and convolve the 2D decoder features to the required 3D UNet input size. Specifically, FLoSP~(\cref{sec:unprojection}) is shown to be the most crucial as it improves remarkably both semantics ($[+12.83$,$+6.72]$ mIoU) and geometry ($[+14.11$,$+9.56]$ IoU). 3D CRP~(\cref{sec:context_prior}) contributes equally to IoU (in $[+0.77$,$+1.12]$) and mIoU (in $[+0.54, +1.33]$). Both SCAL losses (\cref{sec:global_loss}) contribute differently as expected, since $\mathcal{L}^{\text{sem}}_{\text{scal}}$ helps semantics ($+1.61$ mIoU in both), while $\mathcal{L}^{\text{geo}}_{\text{scal}}$ boosts geometry ($[+1.55, +2.20]$ IoU). In NYUv2 only, $\mathcal{L}^{\text{sem}}_{\text{scal}}$ harms IoU (-0.31) but improves the same metric on SemanticKITTI (+0.34). 
Finally, the frustums proportion loss~(\cref{sec:frustum_loss}) boosts both metrics on both datasets by at least $+0.38$ and up to $+0.61$.
\begin{table}
	\footnotesize
	\begin{subtable}[b]{0.46\linewidth}
		\centering
		\captionsetup{font=scriptsize}
		\setlength{\tabcolsep}{0.01\linewidth}
		\begin{tabular}[b]{l|cc}
			\toprule
			\makecell{2D scales\\[-0.4em]\scriptsize{~($S$)}} & IoU $\uparrow$ & mIoU $\uparrow$ \\
			\midrule
			$1, 2, 4, 8$ & \textbf{42.51} \tiny$\pm$0.15 & \textbf{26.94} \tiny$\pm$0.10 \\
			$1, 2, 4$ & \underline{42.08} \tiny$\pm$0.69 & \underline{26.28} \tiny$\pm$0.24\\
			$1, 2$ & 41.56 \tiny$\pm$0.18 &	25.66 \tiny$\pm$0.21 \\
			$1$ & 41.57 \tiny$\pm$0.11 & 25.61 \tiny$\pm$0.43 \\
			\midrule
			w/o FLoSP & 28.39 \tiny$\pm$0.53 & 14.11 \tiny$\pm$0.21 \\
			\bottomrule
		\end{tabular}
		\caption{Scales in FLoSP}
		\label{tab:ablate_unprojection}
	\end{subtable}\hfill%
	\begin{subtable}[b]{0.49\linewidth}
		\centering
		\captionsetup{font=scriptsize}
		\setlength{\tabcolsep}{0.01\linewidth}
		\newcolumntype{P}[1]{>{\centering\arraybackslash}p{#1}}
		\begin{tabular}[b]{P{0.1\linewidth}|c|cc}
			\toprule
			$n$ & $\mathcal{L}_{rel}$ & IoU $\uparrow$ & mIoU $\uparrow$ \\
			\midrule
			\multirow{2}{*}{4} & \cmark & \textbf{42.51} \tiny$\pm$0.15 & \textbf{26.94} \tiny$\pm$0.10 \\
			& \xmark & \underline{42.24} \tiny$\pm$0.15 &	26.55 \tiny$\pm$0.29 \\
			\midrule
			\multirow{2}{*}{2}  & \cmark & 42.09 \tiny$\pm$0.15 & \underline{26.63} \tiny$\pm$0.05 \\
			& \xmark & 42.15 \tiny$\pm$0.26 & 26.47 \tiny$\pm$0.16 \\
			\midrule
			\multicolumn{2}{c|}{w/o 3D CRP} & 41.39 \tiny$\pm$0.08 & 26.27 \tiny$\pm$0.15 \\
			\bottomrule
		\end{tabular}
		\caption{Relations and supervision in 3D CRP}
		\label{tab:ablation_CP3D}
	\end{subtable}\vspace{-0.5em}
	\caption{\textbf{Study of FLoSP and 3D CRP.} \subref{tab:ablate_unprojection} Projecting from different 2D scales ($S$) in FLoSP (\cref{sec:unprojection}) show more scales is better. \subref{tab:ablation_CP3D} In our 3D CRP (\cref{sec:context_prior}) using more relations ($n$) and supervision ($\mathcal{L}_{rel}$) lead to higher metrics. Results are on NYUv2.}
	\label{tab:temps}
\end{table}
\begin{figure}
	\centering
	\begin{subfigure}[m]{0.37\linewidth}
		\centering
		\scriptsize
		\captionsetup{font=scriptsize}
		\newcolumntype{P}[1]{>{\centering\arraybackslash}p{#1}}
		\setlength{\tabcolsep}{0.02\linewidth}
		\begin{tabular}{P{0.05\linewidth}c}
			\rotatebox{90}{CoReNet}&\includegraphics[width=0.63\linewidth]{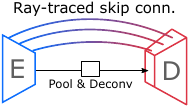}\\
			\rotatebox{90}{Ours-light}&\includegraphics[width=0.93\linewidth]{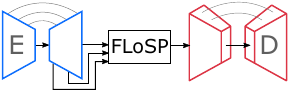}\\
		\end{tabular}\vspace{-0.2em}
		\caption{Architectures ({\textcolor{blue}{2D}} or {\textcolor{red}{3D}})}
		\label{fig:corenet_comparison_arch}
	\end{subfigure}\hfill%
	\begin{subfigure}[m]{0.55\linewidth}
		\centering
		\centering
		\footnotesize
		\captionsetup{font=scriptsize}
		\setlength{\tabcolsep}{0.01\linewidth}
		\begin{tabular}{l|cc}
			\toprule
			& IoU $\uparrow$ & mIoU $\uparrow$ 
			\\
			\midrule
			CoReNet \scriptsize{(1,2,4)} & 30.60 \tiny$\pm$0.46 & 17.34 \tiny$\pm$0.37
			\\
			Ours-light \scriptsize{(1,2,4)} & \textbf{40.80} \tiny$\pm$0.43 & \textbf{25.90} \tiny$\pm$0.63
			\\
			\midrule
			Ours-light \scriptsize{(1)} & 40.13 \tiny$\pm$0.31 & 25.33 \tiny$\pm$0.58
			\\
			\bottomrule
		\end{tabular}
		\caption{Performance}
		\label{fig:corenet_comparison_perf}
	\end{subfigure}
	\caption{\textbf{Type of 2D-3D features projections.} \subref{fig:corenet_comparison_arch} Comparing our FLoSP and `Ray-traced skip connections' from CoReNet~\cite{Popov2020CoReNetC3} (cf. text) shows in \subref{fig:corenet_comparison_perf} we get significantly better results at comparable (1,2,4) scales or even with only one (1).} \vspace{-0.7em}
	\label{fig:corenet_comparison}
\end{figure}
\\
\condenseparagraph{Effect of features projection.}We now study in-depth the effect of FLoSP (\cref{sec:unprojection}) at the core our RGB-based task.
In \cref{tab:ablate_unprojection}, we use our FLoSP projecting only 2D features from given 2D scales by changing $S$ in \cref{eq:project2d3d}. More 2D scales boosts IoU and mIoU consistently and leans to lower variance -- showing (1,2,4,8) projections are indeed best.

We further compare FLoSP to the `Ray-traced skip connections' of CoReNet~\cite{Popov2020CoReNetC3} being close in nature, putting our best effort to push CoReNet performance.
To properly evaluate only the effect of features projection, we remove our other components, producing a light version (`Ours-light') with the same 2D encoder (E), 3D decoder (D), and projection scales (1,2,4), corresponding to all possible scales in the 3D decoder, as in CoReNet, shown in \cref{fig:corenet_comparison_arch}. 
On NYUv2, \cref{fig:corenet_comparison_perf} shows FLoSP is very significantly better (+10.2~IoU, +8.56~mIoU). We conjecture this relates to the fact that CoReNet applies same-scale 2D-3D connections, while FLoSP disentangles 2D-3D scales, letting the network relies on self-learned relevant features, confirmed by the good performance of the (1) scale in \cref{fig:corenet_comparison_perf}.

\condenseparagraph{Effect of relations in 3D CRP.}While 3D CRP (\cref{sec:context_prior}) is shown beneficial in \cref{tab:general_ablation}, we evaluate the effect of different numbers of relations (\ie~$n$).
\cref{tab:ablation_CP3D} shows the benefit of our 4 relations 
$\mathcal{M}{=}\{\textbf{f}_{\textbf{s}},\textbf{f}_{\textbf{d}},\textbf{o}_{\textbf{s}},\textbf{o}_{\textbf{d}}\}$
instead of only 2 (\ie $\mathcal{M}{=}\{\textbf{s},\textbf{d}\}$),
which matches our expectation due to the overwhelming imbalance of free/occupied voxels ($\approx{}9{:}1$ in NYUv2). 
Our supervision of relation matrices $\hat{A}$ with the relation loss $\mathcal{L}_{rel}$ from \cref{eq:loss_rel} also shows an increase of all metrics. Without supervision, our 3D CRP acts as a self-attention layer that learns the context information implicitly.
\begin{table}
	\centering
	\setlength{\tabcolsep}{0.01\linewidth}
	\centering
	\footnotesize
	\begin{tabular}{l|cc|cc}
		\toprule
		& \multicolumn{2}{c}{NYUv2} & \multicolumn{2}{c}{SemanticKITTI} \\
		$\ell{\times}\ell$ & IoU $\uparrow$ & mIoU $\uparrow$ &  IoU $\uparrow$ & mIoU $\uparrow$ \\
		\midrule	
		$8 \times 8$ & \underline{42.51} \tiny$\pm$0.15 & \textbf{26.94} \tiny$\pm$0.10 &
		\textbf{37.12} \tiny$\pm$0.15 & \textbf{11.50} \tiny$\pm$0.14
		\\					
		$4 \times 4$ & \textbf{42.52} \tiny$\pm$0.12 & \underline{26.85} \tiny$\pm$0.19 &
		\underline{37.09} \tiny$\pm$0.09 & \underline{11.45} \tiny$\pm$0.15 
		\\
		$2 \times 2$ & 42.41 \tiny$\pm$0.13 & \underline{26.85} \tiny$\pm$0.22 
		& 36.88 \tiny$\pm$0.11 & 11.27 \tiny$\pm$0.25
		\\
		$1 \times 1$ & 42.39 \tiny$\pm$0.18 & 26.52 \tiny$\pm$0.31
		& 36.83 \tiny$\pm$0.42 & 11.33 \tiny$\pm$0.11
		\\									
		\midrule
		w/o $\mathcal{L}_{\text{fp}}$ & 41.90 \tiny$\pm$0.26 & 26.37 \tiny$\pm$0.16 
		& 36.74 \tiny$\pm$0.33 & 11.11 \tiny$\pm$0.24
		\\
		\bottomrule
	\end{tabular}
	\caption{\textbf{Frustums Proportion loss ablation.} Varying the number of local frustums ($\ell{\times}\ell$) in our loss shows more frustums (\ie smaller) lead to finer guidance and better results on both datasets.} 
	\label{tab:frustum_loss_ablation}
\end{table}
\condenseparagraph{Effect of local frustums loss.}\cref{tab:frustum_loss_ablation} shows the effect of varying number of $\ell{\times}\ell$ frustums (\cref{eq:frustum_loss}, \cref{sec:frustum_loss}) on both datasets. 
Higher numbers result in smaller frustums, \ie finer local supervision. As $\ell{\times}\ell$ increases, all metrics increase accordingly, showing the loss benefit, especially when compared to applied image-wise (\ie~$1{\times}1$).
\begin{figure}
	\centering
	\scriptsize
	\setlength{\fboxsep}{0.005pt}
	\newcommand{\tmpframe}[1]{\fbox{#1}}
	\newcommand{\im}[3]{\stackinset{r}{0cm}{b}{0pt}{\tmpframe{\includegraphics[height=#3]{#1}}}{\includegraphics[width=\linewidth]{#2}}} %
	\newcommand{\imcs}[1]{\im{imgs/generalization/cityscapes/#1.png}{imgs/generalization/cityscapes/#1_our.png}{0.5cm}}
	\newcommand{\imns}[1]{\im{imgs/generalization/nuscenes/#1_nclass.jpg}{imgs/generalization/nuscenes/#1_nclass_0_our.png}{0.5cm}}
	\newcommand{\imkt}[1]{\im{imgs/kitti_qualitative/#1.png}{imgs/kitti_qualitative/#1_0_our.png}{0.36cm}}
	\newcommand{\imktnew}[1]{\im{imgs/generalization/kitti360/#1.png}{imgs/generalization/kitti360/#1_0_our.png}{0.35cm}}
	\newcolumntype{P}[1]{>{\centering\arraybackslash}m{#1}}
	\setlength{\tabcolsep}{0.01\linewidth}
	\renewcommand{\arraystretch}{0.0}
	\begin{tabular}{P{0.235\linewidth} P{0.235\linewidth} P{0.235\linewidth} P{0.235\linewidth}}		
		Cityscapes~\cite{cityscapes} & 
		Nuscenes~\cite{nuscenes2019} & 
		\textbf{SemanticKITTI~\cite{behley2019iccv}}
		& KITTI-360~\cite{KITTI360}\\
		\imcs{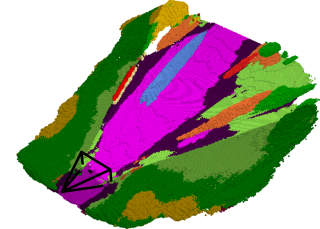} & \imns{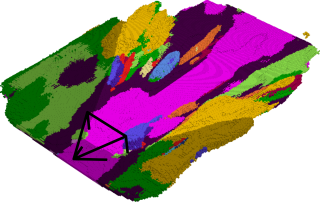} & \imkt{000465} & \imktnew{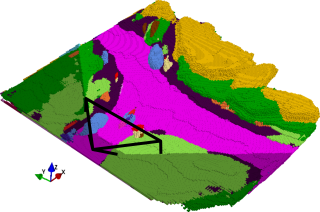}
		\\
		\imcs{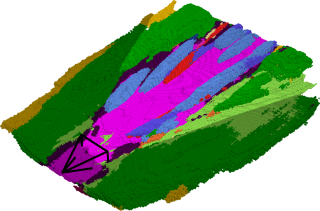} & \imns{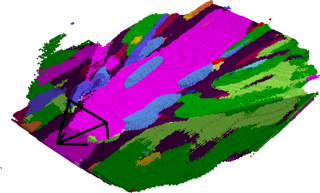} & \imkt{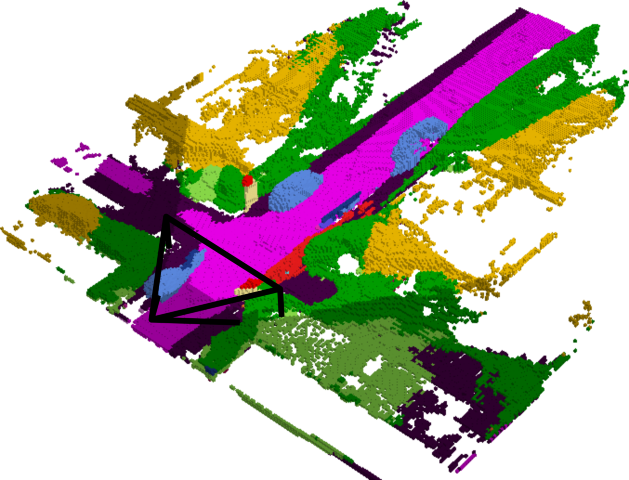} & \imktnew{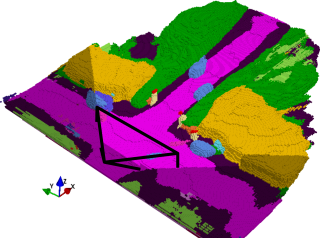}\\
		(49$^{\circ}$, 25$^{\circ}$) & (65$^{\circ}$, 39$^{\circ}$) & (H=82$^{\circ}$, V=29$^{\circ}$) & (104$^{\circ}$, 38$^{\circ}$)
		\\
	\end{tabular}
	\caption{\textbf{Camera effects.} Outputs of MonoScene when trained on Sem.KITTI having horizontal FOV of 82$^\circ$, and tested on datasets with decreasing (left) or increasing (right) FOV.}
	\label{fig:fov_generalization}
\end{figure}
\section{Discussion}
MonoScene tackles monocular SSC originally using successive 2D-3D UNets, bridged by a new features projection, with increased contextual awareness and new losses. 

\condenseparagraph{Limitations.}Despite good results, our framework still struggles to infer fine-grained geometry, or to separate semantically-similar classes, \eg car/truck or chair/sofa. 
It also performs poorly on small objects partly due to their scarcity~(${<}0.3\%$ in Sem.KITTI~\cite{behley2019iccv}).
Due to the single viewpoint, occlusion artefacts such as distortions are visible along the line of sight in outdoor scenes.
Additionally, as we exploit 2D-3D projection with the FLoSP module (\cref{sec:unprojection}), we evaluate the effect of \textit{inferring} from datasets having various camera setups, showing in~\cref{fig:fov_generalization} that results -- though consistent -- have increasingly greater distortion when departing from the camera settings of the training set. 

\condenseparagraph{Broader impact, Ethics.}Jointly understanding the 3D geometry and semantics from image paves ways for better mixed reality, photo editing or mobile robotics applications. 
But the inevitable errors in the scene understanding could have fatal issues (\eg autonomous driving) and such algorithms should always be seconded by other means.
{\footnotesize \condenseparagraph{Acknowledgment}This work used HPC resources from GENCI–IDRIS (Grant 2021-AD011012808). It was done in the SAMBA collaborative project, co-funded by BpiFrance in the Investissement d’Avenir Program.}

\appendix
\newpage
We provide details about the main baselines and MonoScene in \cref{sec:supp-architecture}, and include additional qualitative and quantitative results in \cref{sec:supp-addresults}.

Results on image sequences are in the supplementary video:
\url{https://youtu.be/qh7La1tRJmE}.

\section{Architectures details}
\label{sec:supp-architecture}
\subsection{Baselines}
\paragraph{AICNet~\cite{Li2020aicnet}.}
\label{para:aic}
We use the official implementation of AICNet\footnote{\url{https://github.com/waterljwant/SSC}}. For the RGB-inferred version, \ie AICNet$^\text{rgb}$, we infer depth with the pre-trained AdaBins~\cite{bhat_adabins_2021} on NYUv2~\cite{NYUv2} and SemanticKITTI~\cite{behley2019iccv} from the official repository\footnote{\url{https://github.com/shariqfarooq123/AdaBins}}.
\paragraph{3DSketch~\cite{3DSketch}.}
We use 3DSketch official code\footnote{\url{https://github.com/charlesCXK/TorchSSC}}. For 3DSketch$^\text{rgb}$, we again use AdaBins (\cf above) and convert depth to TSDF with `tsdf-fusion'\footnote{\url{https://github.com/andyzeng/tsdf-fusion}} from the 3DMatch Toolbox~\cite{zeng20163dmatch}.
\paragraph{JS3C-Net~\cite{yan2021JS3CNet}.}
We use the official code of JS3C-Net\footnote{\url{https://github.com/yanx27/JS3C-Net}}. For JS3C-Net$^\text{rgb}$, we generate the input point cloud by unprojecting the predicted depth (using AdaBins) to 3D using the camera intrinsics. 
The semantic point clouds, required to train JS3C-Net, are obtained by augmenting the unprojected point clouds with the 2D semantics obtained using the code\footnote{\url{https://github.com/YeLyuUT/SSeg}} of~\cite{zhu2019labelrelaxation}.
\paragraph{LMSCNet~\cite{roldao2020lmscnet}.}
We use the official implementation of LMSCNet\footnote{\url{https://github.com/cv-rits/LMSCNet}}. For LMSCNet$^\text{rgb}$, the input occupancy grid is obtained by discretizing the unprojected point cloud.

\subsection{MonoScene}
\label{sec:3DUnet_detail}
\begin{figure}
	\centering
	\includegraphics[width=1.0\linewidth]{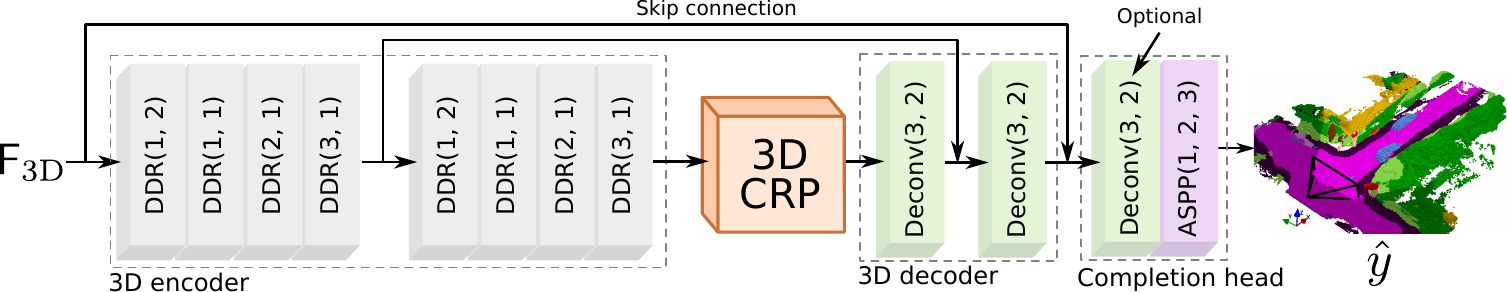}
	\caption{\textbf{MonoScene 3D network.} The 3D UNet uses 2 downscale layers with DDR blocks~\cite{Li2019ddr} and 2 upscale layers with deconv. The completion head uses ASPP and an \textit{optional} deconv layer. Notations: DDR(dilation, downsample rate), Deconv(kernel size, dilation), ASPP(dilations).}
	\label{fig:3DUnet}
\end{figure}

\cref{fig:3DUnet} details our 3D UNet. 
Similar to 3DSketch~\cite{3DSketch}, we adopt DDR~\cite{Li2019ddr} as the basic building block for large receptive field and low memory cost. 
The 3D encoder has 2 layers, each downscales by half and has 4 DDR blocks. The 3D decoder has two deconv layers, each doubles the scale. 
Similar to others~\cite{roldao2020lmscnet} the completion head has an ASPP with dilations (1, 2, 3) to gather multi-scale features and an \textit{optional} deconv to reach output size -- used in SemanticKITTI only.

For training, MonoScene took 7 hours using 2 V100 32g GPUs (2 items per GPU) on NYUv2~\cite{NYUv2} and 28 hours to train using 4 V100 32g GPUs (1 item per GPU) on SemanticKITTI~\cite{behley2019iccv}.

\begin{table*}
	\scriptsize
	\setlength{\tabcolsep}{0.0035\linewidth}
	\newcommand{\classfreq}[1]{{~\tiny(\semkitfreq{#1}\%)}}  %
	\centering
	\begin{tabular}{l|c|c|c c c c c c c c c c c c c c c c c c c|c}
		\toprule
		& & SC & \multicolumn{20}{c}{SSC} \\
		Method
		& SSC Input
		& IoU
		& \rotatebox{90}{\textcolor{road}{$\blacksquare$} road\classfreq{road}} 
		& \rotatebox{90}{\textcolor{sidewalk}{$\blacksquare$} sidewalk\classfreq{sidewalk}}
		& \rotatebox{90}{\textcolor{parking}{$\blacksquare$} parking\classfreq{parking}} 
		& \rotatebox{90}{\textcolor{other-ground}{$\blacksquare$} other-ground\classfreq{otherground}} 
		& \rotatebox{90}{\textcolor{building}{$\blacksquare$} building\classfreq{building}} 
		& \rotatebox{90}{\textcolor{car}{$\blacksquare$} car\classfreq{car}} 
		& \rotatebox{90}{\textcolor{truck}{$\blacksquare$} truck\classfreq{truck}} 
		& \rotatebox{90}{\textcolor{bicycle}{$\blacksquare$} bicycle\classfreq{bicycle}} 
		& \rotatebox{90}{\textcolor{motorcycle}{$\blacksquare$} motorcycle\classfreq{motorcycle}} 
		& \rotatebox{90}{\textcolor{other-vehicle}{$\blacksquare$} other-vehicle\classfreq{othervehicle}} 
		& \rotatebox{90}{\textcolor{vegetation}{$\blacksquare$} vegetation\classfreq{vegetation}} 
		& \rotatebox{90}{\textcolor{trunk}{$\blacksquare$} trunk\classfreq{trunk}} 
		& \rotatebox{90}{\textcolor{terrain}{$\blacksquare$} terrain\classfreq{terrain}} 
		& \rotatebox{90}{\textcolor{person}{$\blacksquare$} person\classfreq{person}} 
		& \rotatebox{90}{\textcolor{bicyclist}{$\blacksquare$} bicyclist\classfreq{bicyclist}} 
		& \rotatebox{90}{\textcolor{motorcyclist}{$\blacksquare$} motorcyclist\classfreq{motorcyclist}} 
		& \rotatebox{90}{\textcolor{fence}{$\blacksquare$} fence\classfreq{fence}} 
		& \rotatebox{90}{\textcolor{pole}{$\blacksquare$} pole\classfreq{pole}} 
		& \rotatebox{90}{\textcolor{traffic-sign}{$\blacksquare$} traffic-sign\classfreq{trafficsign}} 
		& mIoU\\
		\midrule
		LMSCNet$^\text{rgb}$~\cite{roldao2020lmscnet} & $\hat{x}^{\text{occ}}_{\text{3D}}$ & 28.61 & 40.68 & 18.22 & 4.38 & 0.00 & 10.31 & 18.33 & 0.00 & 0.00 & 0.00 & 0.00 & 13.66 & 0.02 & 20.54 & 0.00 & 0.00 & 0.00 & 1.21 & 0.00 & 0.00 & 6.70 \\ %
		3DSketch$^\text{rgb}$~\cite{3DSketch} & $x^{\text{rgb}}$,$\hat{x}^{\text{TSDF}}$ & 33.30 & 41.32 & 21.63 & 0.00 & 0.00 & \underline{14.81} & 18.59 & 0.00 & 0.00 & 0.00 & 0.00 & \textbf{19.09} & 0.00 & 26.40 & 0.00 & 0.00 & 0.00 & 0.73 & 0.00 & 0.00 & 7.50  \\ %
		AICNet$^\text{rgb}$~\cite{Li2020aicnet} & $x^{\text{rgb}}$,$\hat{x}^{\text{depth}}$ & 29.59 & 43.55 & 20.55 & \underline{11.97} & \underline{0.07} & 12.94 & 14.71 & \underline{4.53} & 0.00 & 0.00 & 0.00 & 15.37 & \underline{2.90} & \textbf{28.71} & 0.00 & 0.00 & 0.00 & 2.52 & 0.06 & 0.00 & 8.31  \\ %
		*JS3C-Net$^\text{rgb}$~\cite{yan2021JS3CNet} & $\hat{x}^{\text{pts}}$ & \textbf{38.98} & 50.49 & \underline{23.74} & 11.94 & \underline{0.07} & \textbf{15.03} & \textbf{24.65} & 4.41 & 0.00 & 0.00 & \textbf{6.15} & 18.11 & \textbf{4.33} & 26.86 & \underline{0.67} & 0.27 & 0.00 & \underline{3.94} & \underline{3.77} & \underline{1.45} & \underline{10.31}  \\ %
		
		\midrule
		MonoScene (ours) & $x^{\text{rgb}}$ & \underline{37.12} & \textbf{57.47} & \textbf{27.05} & \textbf{15.72} & \textbf{0.87} & 14.24 & \underline{23.55} & \textbf{7.83} & \textbf{0.20} & \textbf{0.77} & \underline{3.59} & \underline{18.12} & 2.57 & \textbf{30.76} & \textbf{1.79} & \textbf{1.03} & 0.00 & \textbf{6.39} & \textbf{4.11} & \textbf{2.48} & \textbf{11.50} \\
		\bottomrule
	\end{tabular}\\
	* Uses pretrained semantic segmentation network.
	\caption{\textbf{Performance on SemanticKITTI~\cite{behley2019iccv} (validation set).} We report the performance on semantic scene completion (SSC - mIoU) and scene completion (SC - IoU) for RGB-inferred baselines and our method.}
	\label{table:kitti_val_perf}
\end{table*}

\section{Additional results}
\label{sec:supp-addresults}
\subsection{SemanticKITTI}
\label{sec:supp-addresults-kitti}
\paragraph{Quantitative performance.}
We report performance on validation set in \cref{table:kitti_val_perf}. Comparing against the test set performance from the main paper Tab.~1b, we notice MonoScene generalizes better than JS3C-Net$^\text{rgb}$ and AICNet$^\text{rgb}$ since the validation and test set gap is smaller (${-}0.42$ vs ${-}1.34$ and ${-}1.22$).
We also report the complete SemanticKITTI official benchmark (\ie hidden test set) in \cref{tab:perf3d_semkitti_supp} showing that while MonoScene uses only RGB, it still outperforms some of the 3D input SSC baselines. 

\begin{table}[h]
	\scriptsize
	\setlength{\tabcolsep}{0.009\linewidth}
	\centering
	\begin{tabular}{l|c|c c}
		\toprule
		Method & Input & {IoU} & mIoU\\
		\midrule
		\textbf{3D} &&&\\[-0.1em]
		SSCNet~\cite{song2016ssc} & $x^{\text{TSDF}}$  & 29.8 & 9.5
		\\
		TS3D~\cite{TS3D} & $x^{\text{TSDF}}{+}x^{\text{rgb}}$  & 29.8 & 9.5
		\\
		TS3D+DNet~\cite{behley2019iccv} & $x^{\text{TSDF}}{+}x^{\text{rgb}}$ & 25.0 & 10.2
		\\	
		ESSCNet~\cite{zhang2018efficient} & $x^{\text{pts}}$ & 41.8 & 17.5
		\\			
		LMSCNet~\cite{roldao2020lmscnet} & $x^{\text{occ}}$ & \underline{56.7} &  17.6		
		\\
		TS3D+DNet+SATNet~\cite{behley2019iccv} & $x^{\text{occ}}$ & 50.6  & 17.7
		\\
		Local-DIFs~\cite{localDif} & $x^{\text{occ}}$ & \textbf{57.7} &  22.7
		\\
		JS3C-Net~\cite{yan2021JS3CNet} & $x^{\text{pts}}$  & 56.6 & \underline{23.8}
		\\
		S3CNet~\cite{Cheng2020S3CNetAS} & $x^{\text{occ}}$ & 45.6 & \textbf{29.5}
		\\
		\midrule
		\textbf{2D} &&&\\[-0.1em]
		MonoScene & $x^{\text{rgb}}$  & {34.2} & {11.1} \\
		\bottomrule
	\end{tabular}
	\caption{\textbf{Complete SemanticKITTI official benchmark (hidden test set)}. Results are taken from~\cite{sscsurvey}. Despite using only single RGB image as input, MonoScene still surpasses some of the SSC baselines with 3D input.}
	\label{tab:perf3d_semkitti_supp}
\end{table}

\begin{figure*}
	\centering
	\newcolumntype{P}[1]{>{\centering\arraybackslash}m{#1}}
	\setlength{\tabcolsep}{0.001\textwidth}
	\renewcommand{\arraystretch}{0.8}
	\footnotesize
	\begin{tabular}{P{0.157\textwidth} P{0.157\textwidth} P{0.157\textwidth} P{0.157\textwidth} P{0.157\textwidth} P{0.157\textwidth}}		
		Input & AICNet$^\text{rgb}$~\cite{Li2020aicnet} & LMSCNet$^\text{rgb}$~\cite{roldao2020lmscnet}  & JS3CNet$^\text{rgb}$~\cite{yan2021JS3CNet}  & \OurMethod\ \scriptsize{(ours)} & Ground Truth
		\\
		\includegraphics[width=.9\linewidth]{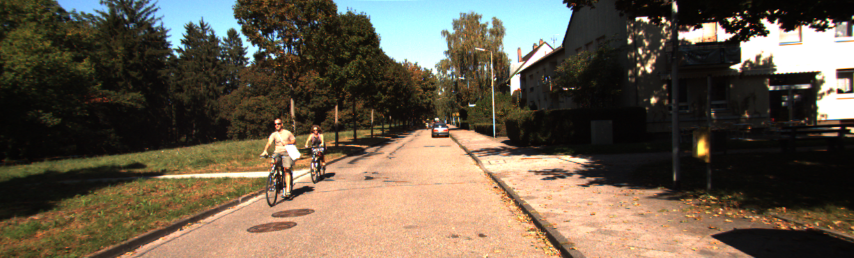} & 
		\includegraphics[width=.9\linewidth]{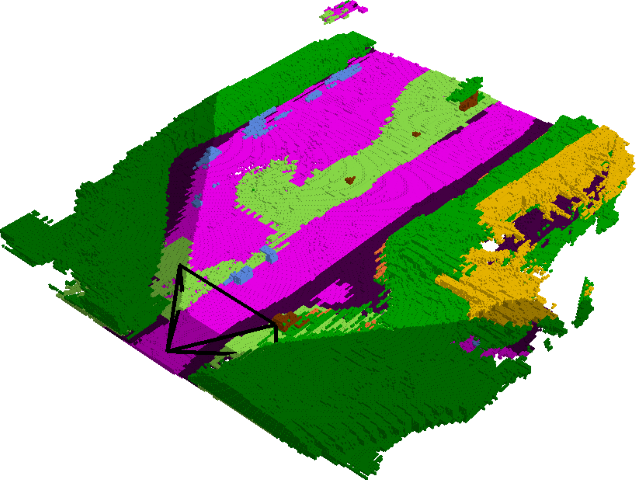} & 
		\includegraphics[width=.9\linewidth]{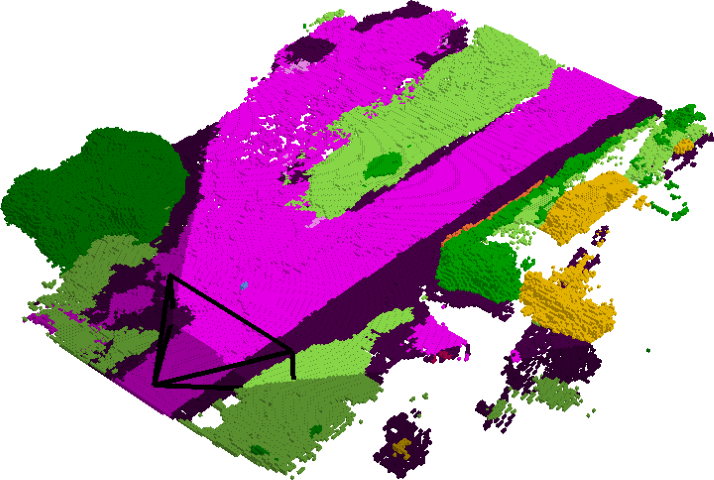} & 
		\includegraphics[width=.9\linewidth]{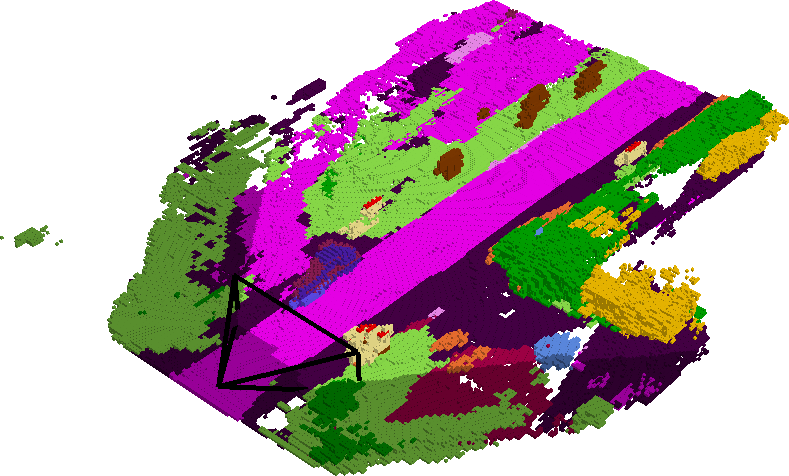} & 
		\includegraphics[width=.9\linewidth]{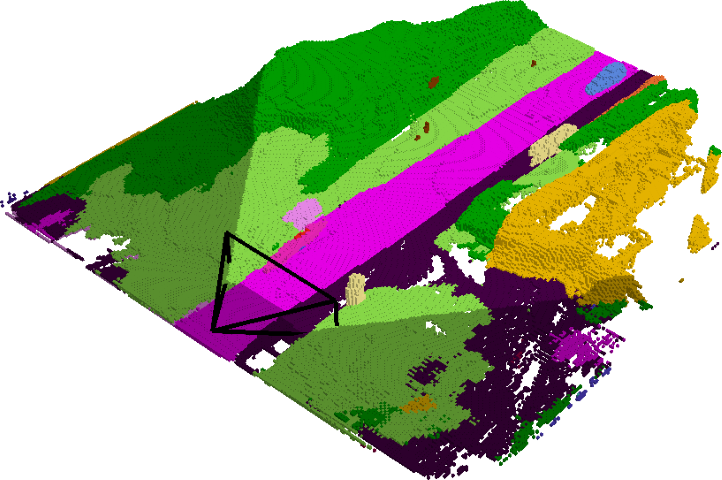} & 
		\includegraphics[width=\linewidth]{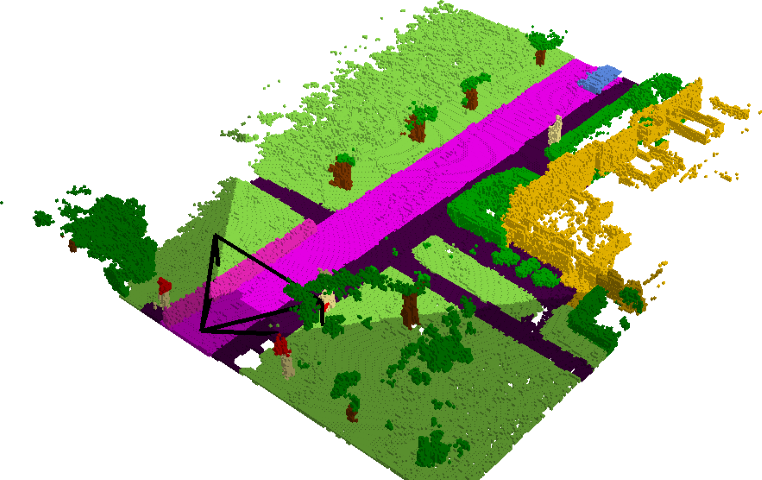} 
		\\		
		\includegraphics[width=.9\linewidth]{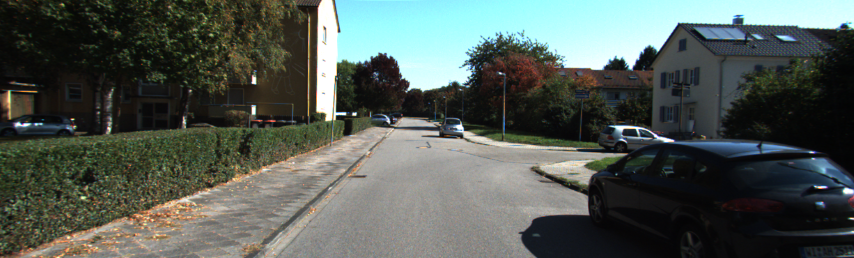} & 
		\includegraphics[width=.9\linewidth]{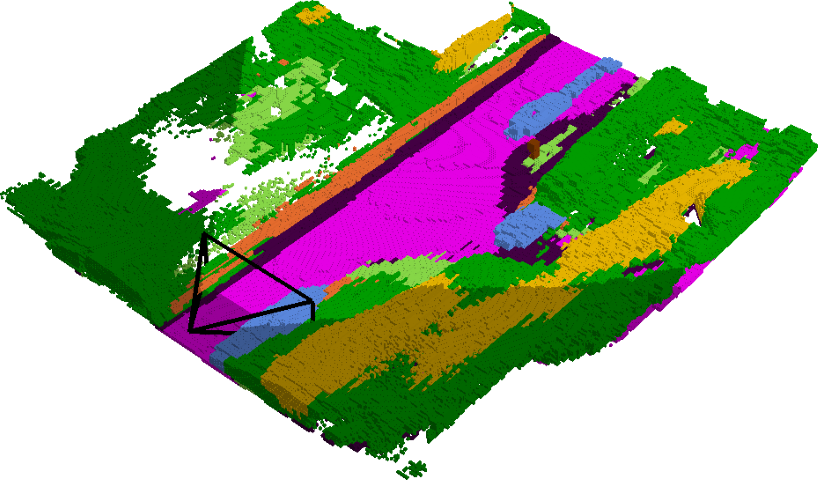} & 
		\includegraphics[width=.9\linewidth]{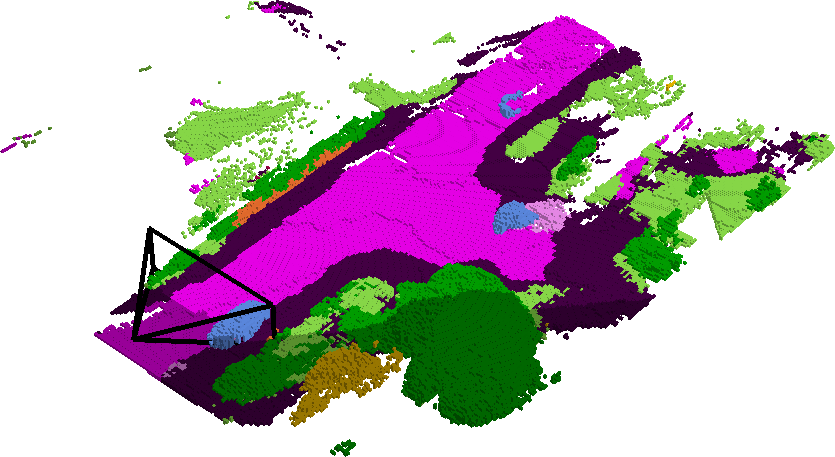} & 
		\includegraphics[width=.9\linewidth]{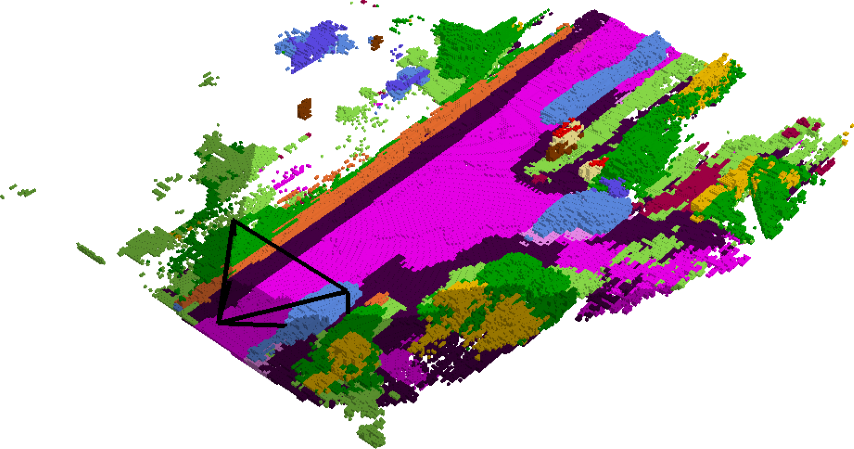} & 
		\includegraphics[width=.9\linewidth]{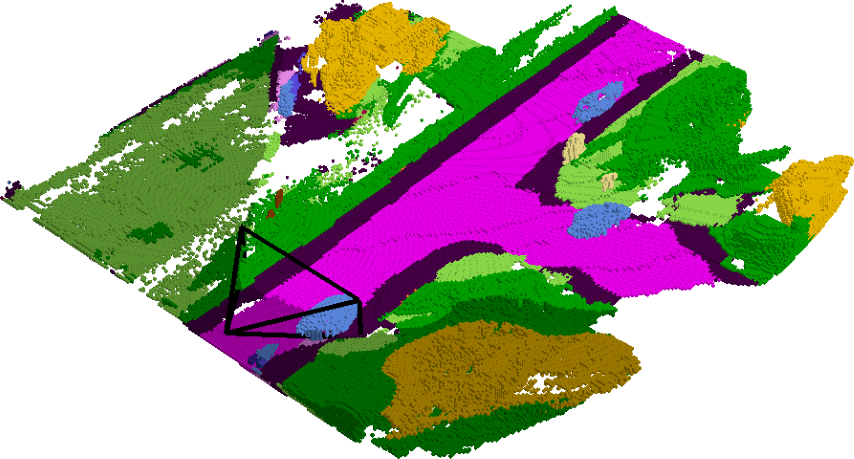} & 
		\includegraphics[width=\linewidth]{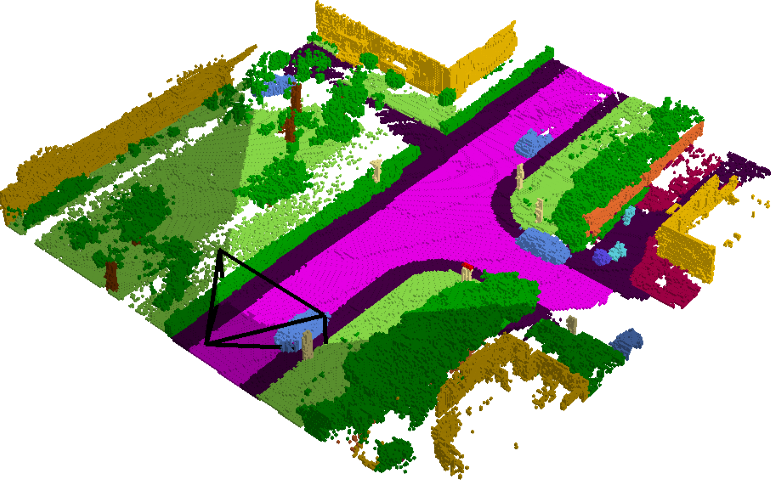} 
		\\
		\includegraphics[width=.9\linewidth]{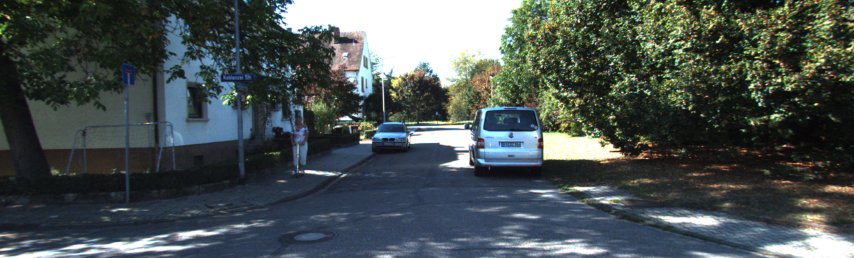} & 
		\includegraphics[width=.9\linewidth]{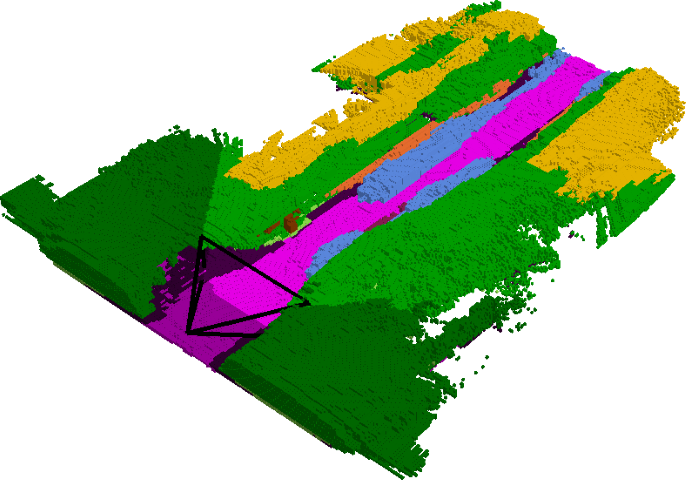} & 
		\includegraphics[width=.9\linewidth]{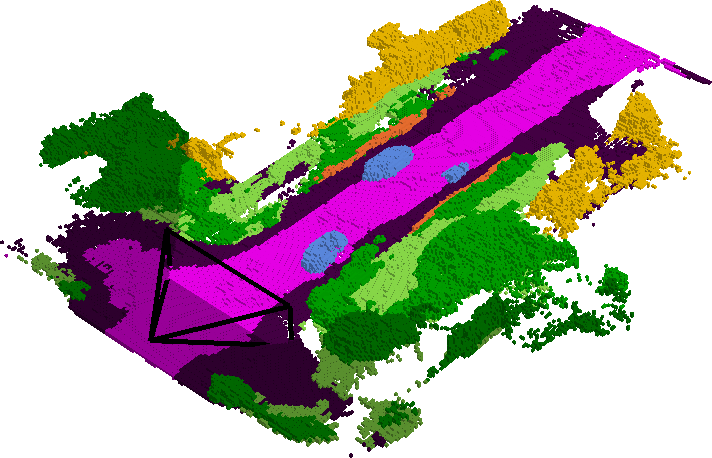} & 
		\includegraphics[width=.9\linewidth]{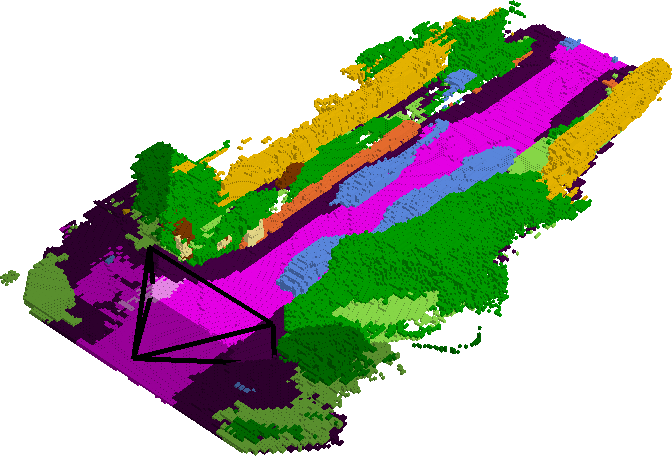} & 
		\includegraphics[width=.9\linewidth]{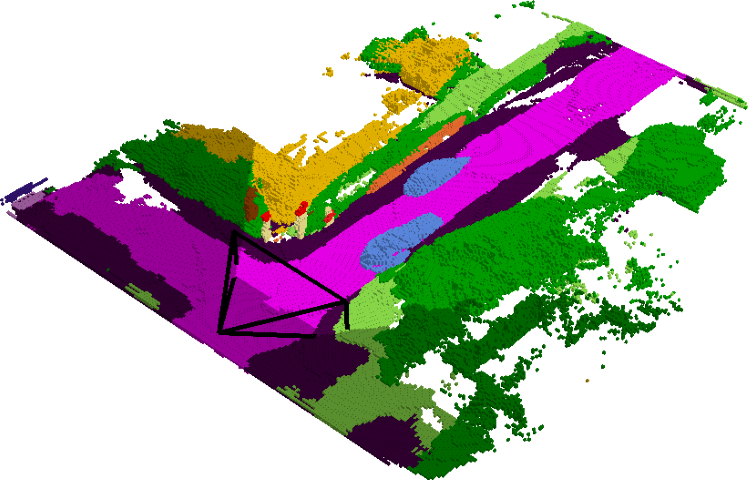} & 
		\includegraphics[width=\linewidth]{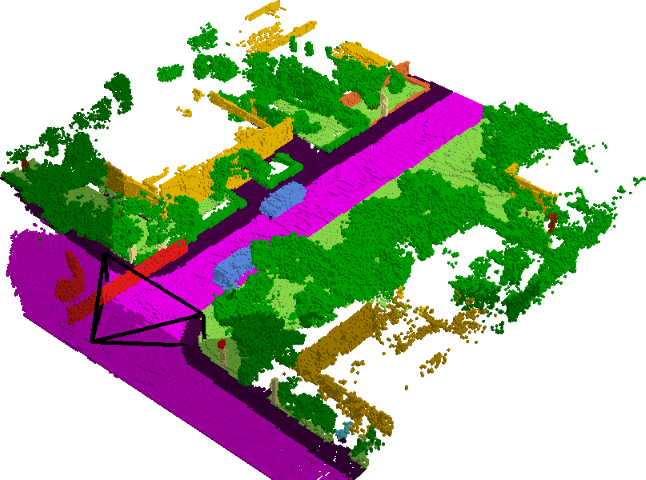} 
		\\
		\includegraphics[width=.9\linewidth]{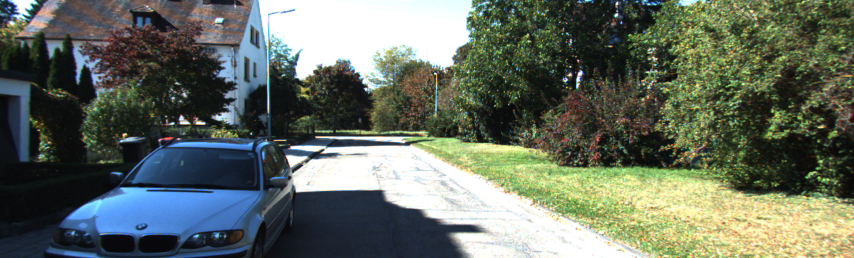} & 
		\includegraphics[width=.9\linewidth]{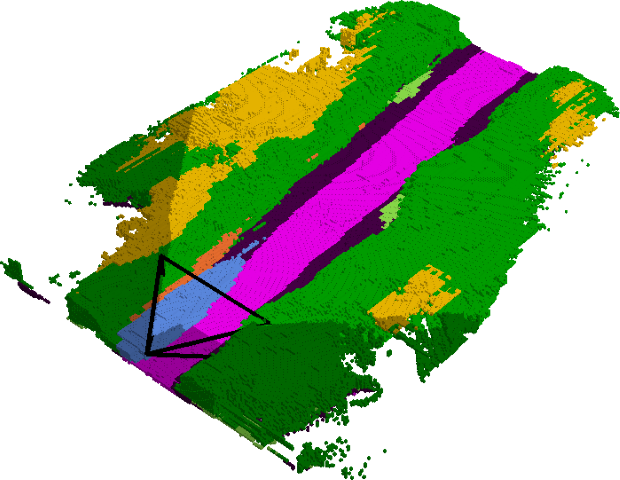} & 
		\includegraphics[width=.9\linewidth]{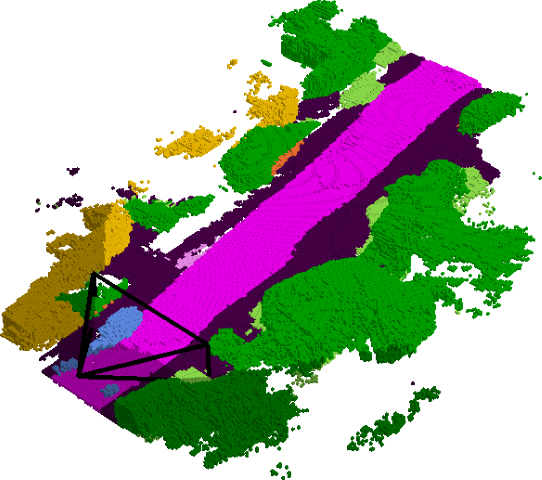} & 
		\includegraphics[width=.9\linewidth]{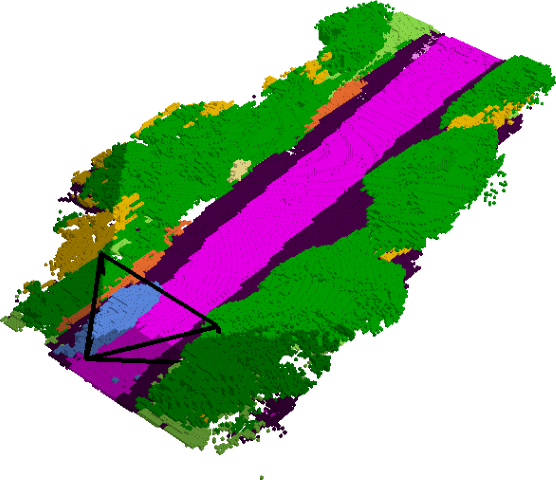} & 
		\includegraphics[width=.9\linewidth]{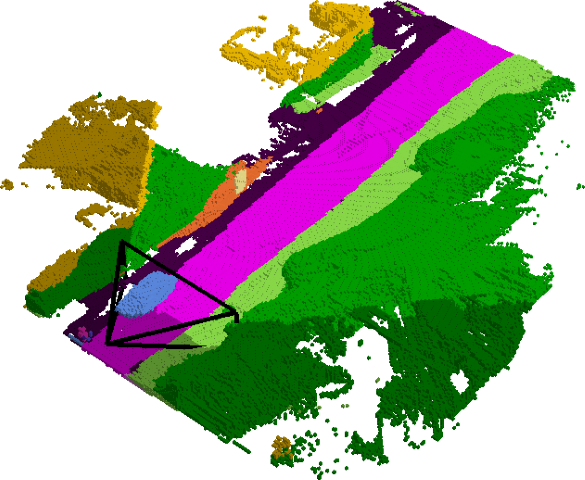} & 
		\includegraphics[width=\linewidth]{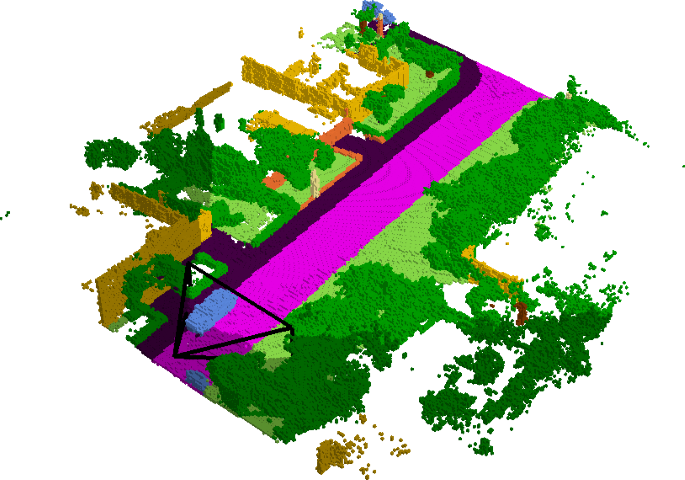} 
		\\
		\includegraphics[width=.9\linewidth]{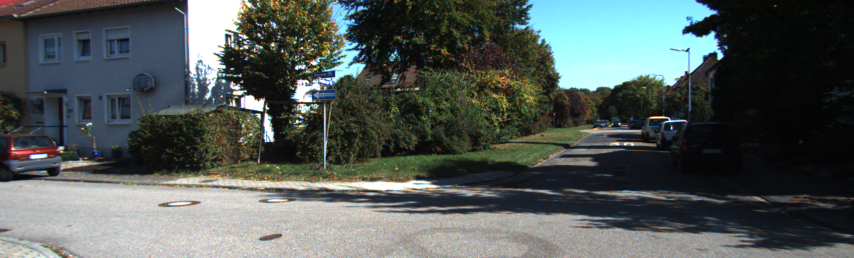} & 
		\includegraphics[width=.9\linewidth]{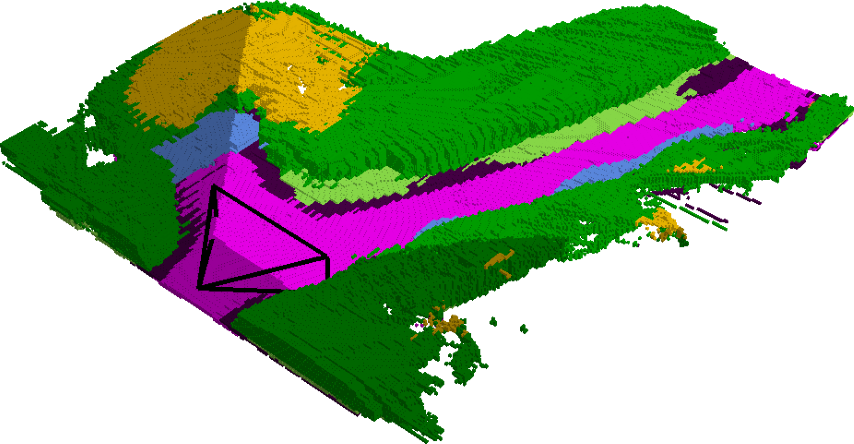} & 
		\includegraphics[width=.9\linewidth]{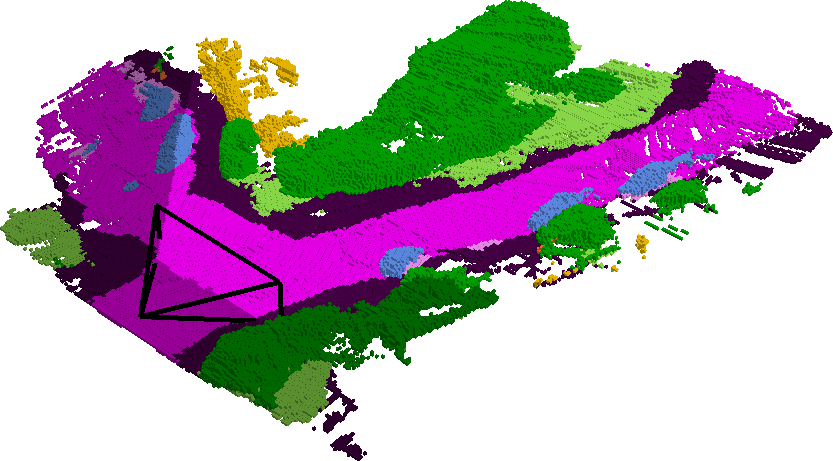} & 
		\includegraphics[width=.9\linewidth]{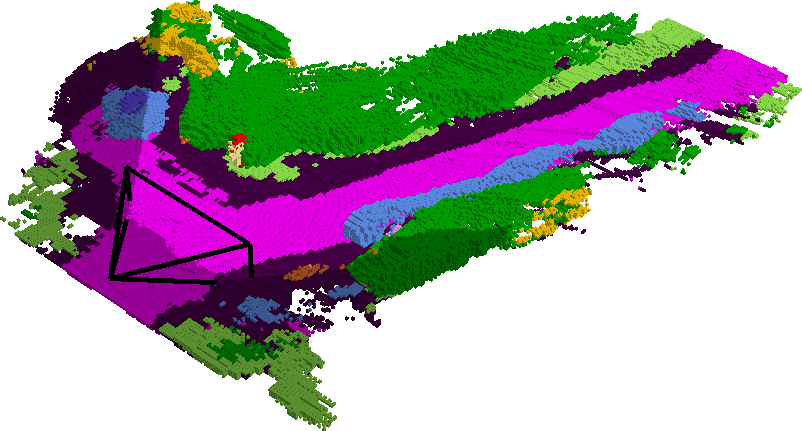} & 
		\includegraphics[width=.9\linewidth]{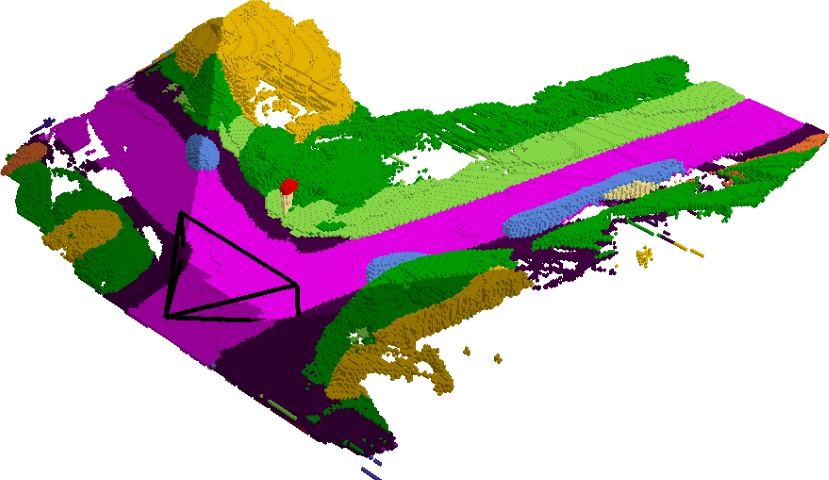} & 
		\includegraphics[width=\linewidth]{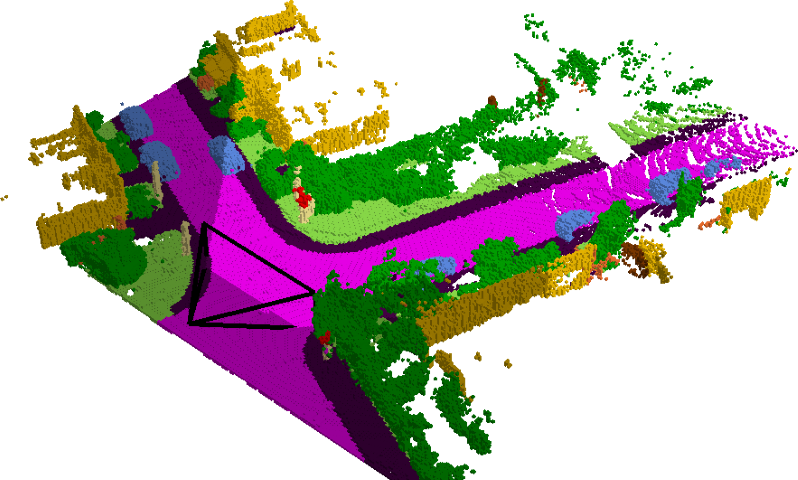} 
		\\
		\includegraphics[width=.9\linewidth]{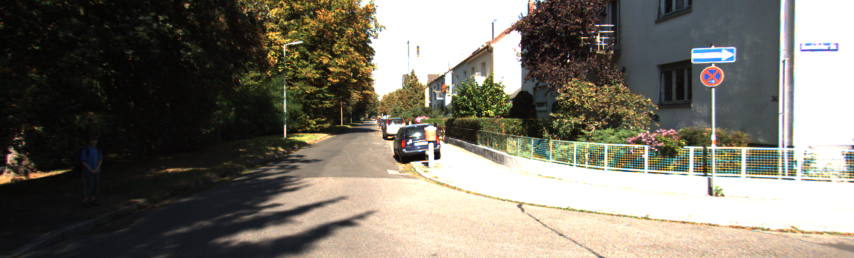} & 
		\includegraphics[width=.9\linewidth]{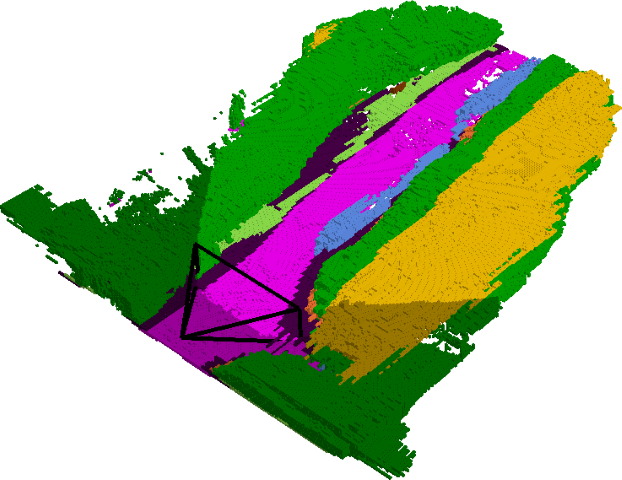} & 
		\includegraphics[width=.9\linewidth]{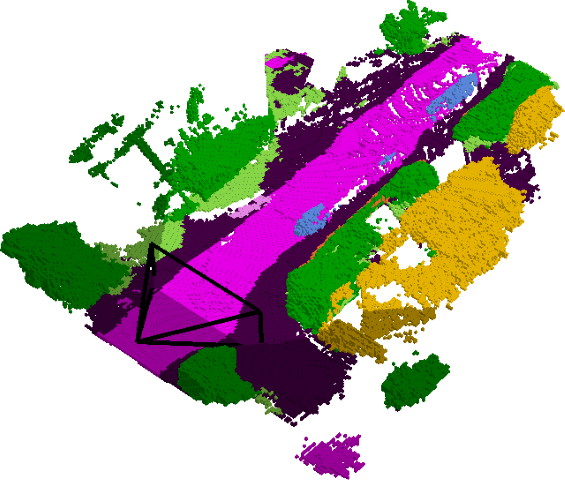} & 
		\includegraphics[width=.9\linewidth]{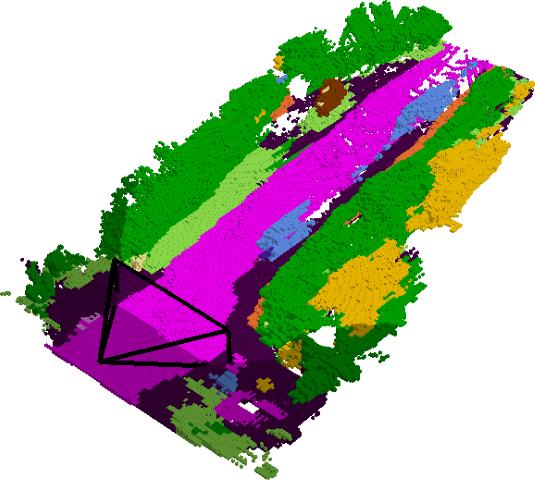} & 
		\includegraphics[width=.9\linewidth]{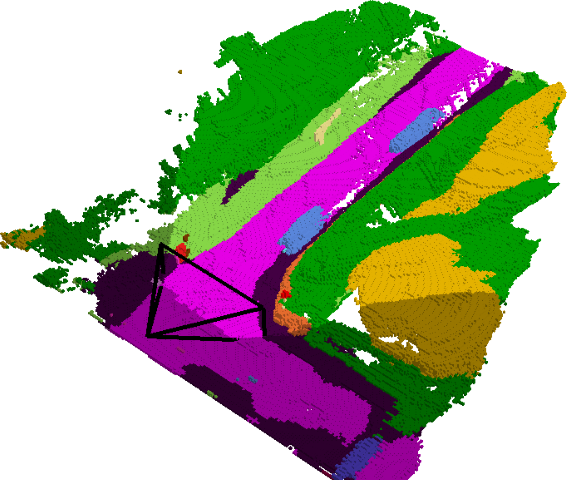} & 
		\includegraphics[width=\linewidth]{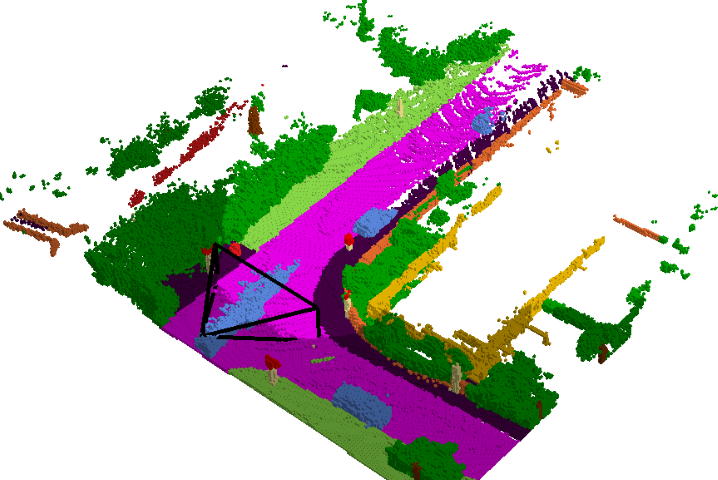} 
		\\
		\includegraphics[width=.9\linewidth]{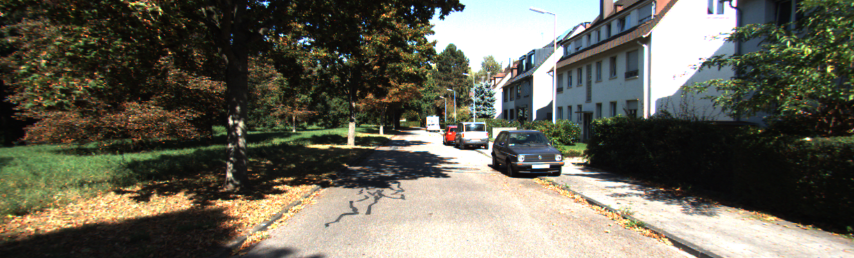} & 
		\includegraphics[width=.9\linewidth]{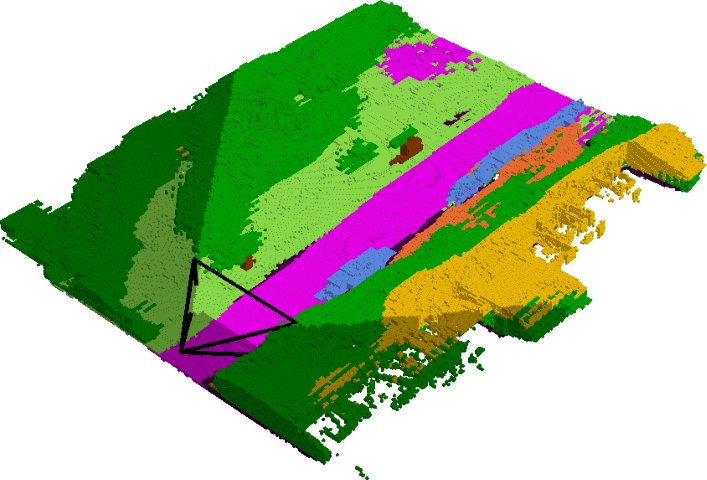} & 
		\includegraphics[width=.9\linewidth]{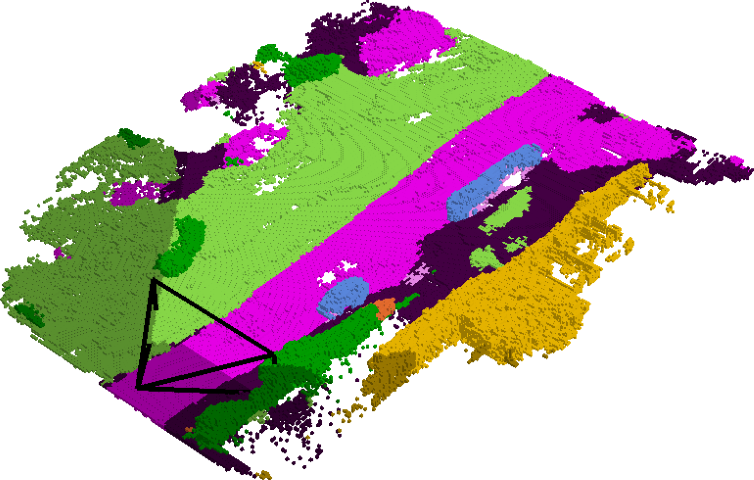} & 
		\includegraphics[width=.9\linewidth]{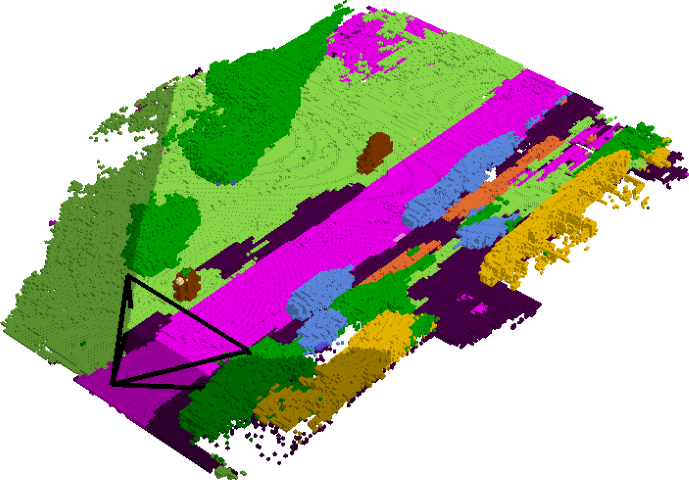} & 
		\includegraphics[width=.9\linewidth]{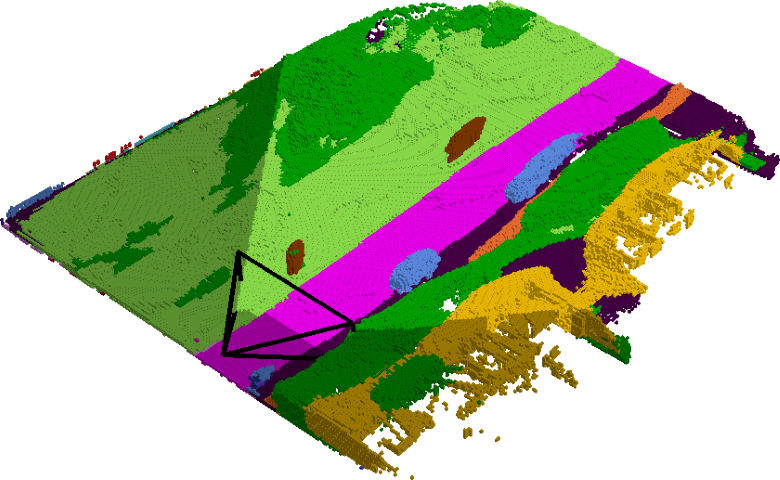} & 
		\includegraphics[width=\linewidth]{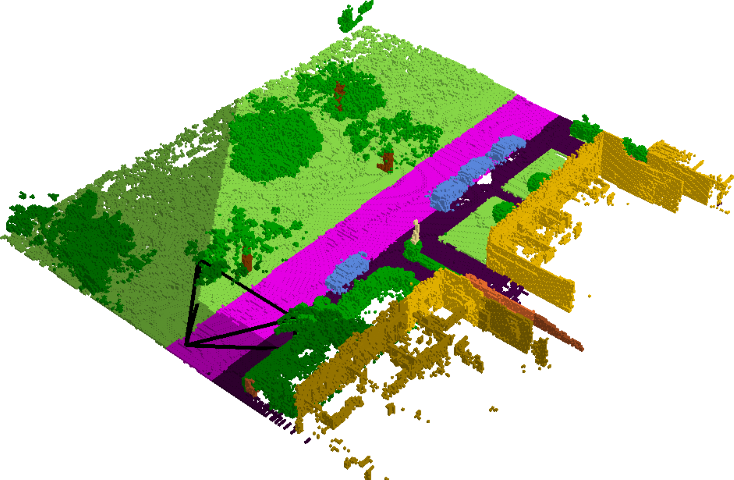} 
		\\
		\includegraphics[width=.9\linewidth]{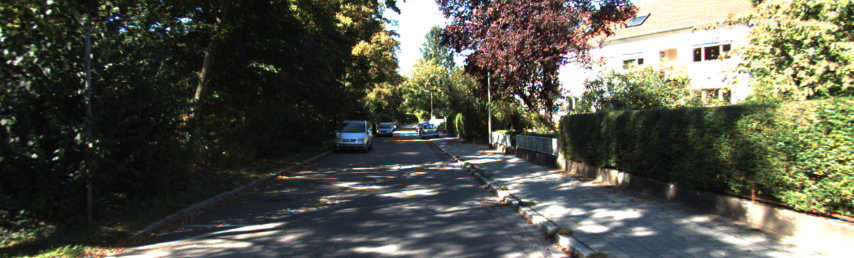} & 
		\includegraphics[width=.9\linewidth]{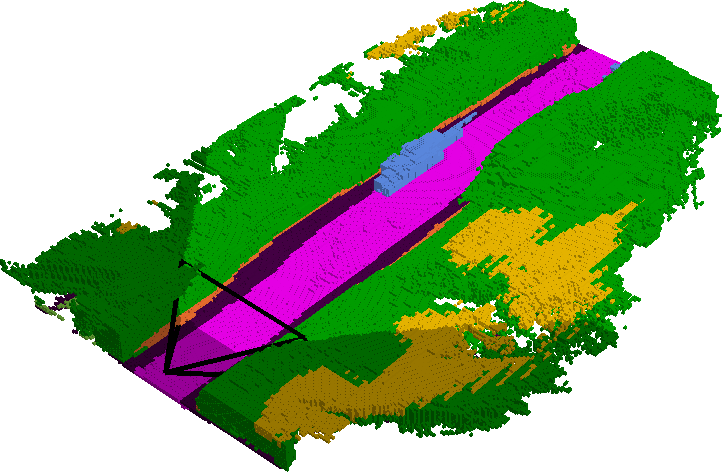} & 
		\includegraphics[width=.9\linewidth]{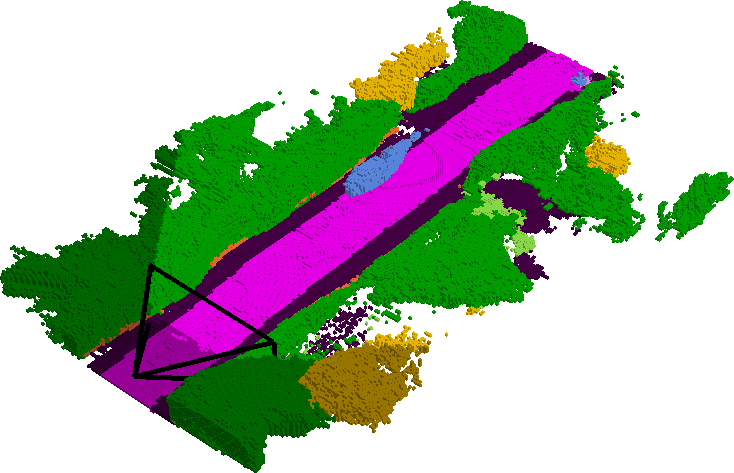} & 
		\includegraphics[width=.9\linewidth]{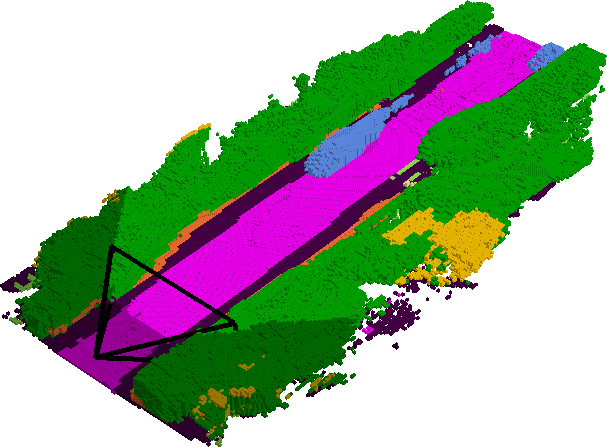} & 
		\includegraphics[width=.9\linewidth]{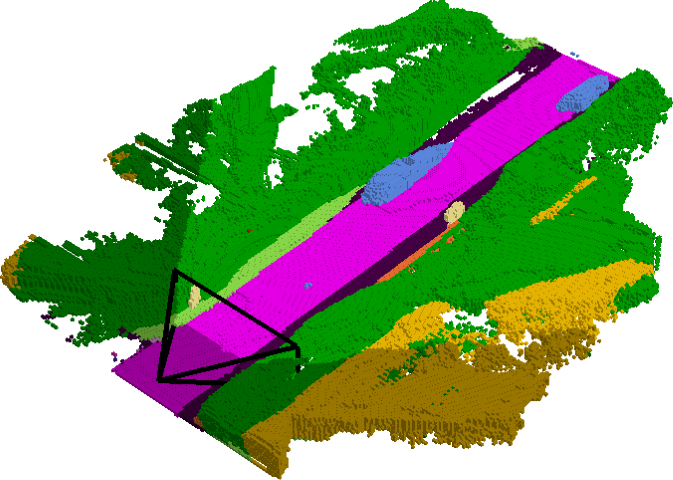} & 
		\includegraphics[width=\linewidth]{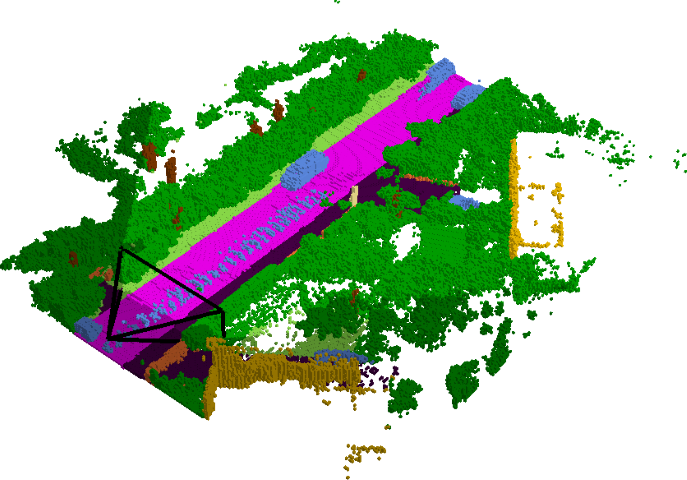} 
		\\
		\includegraphics[width=.9\linewidth]{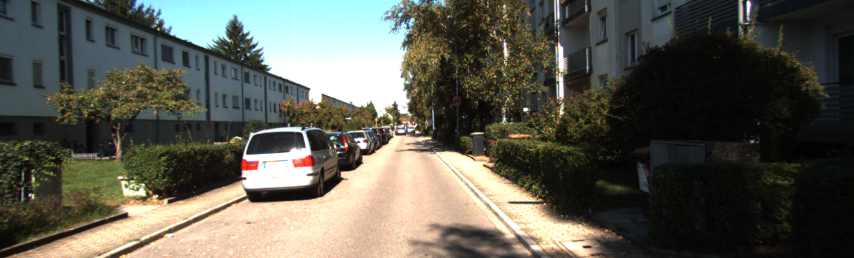} & 
		\includegraphics[width=.9\linewidth]{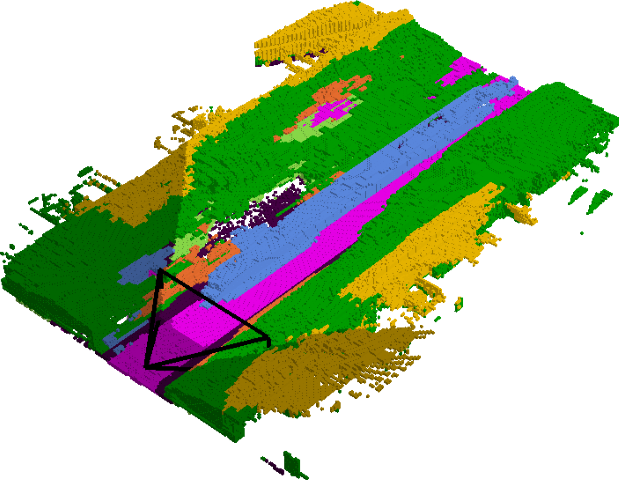} & 
		\includegraphics[width=.9\linewidth]{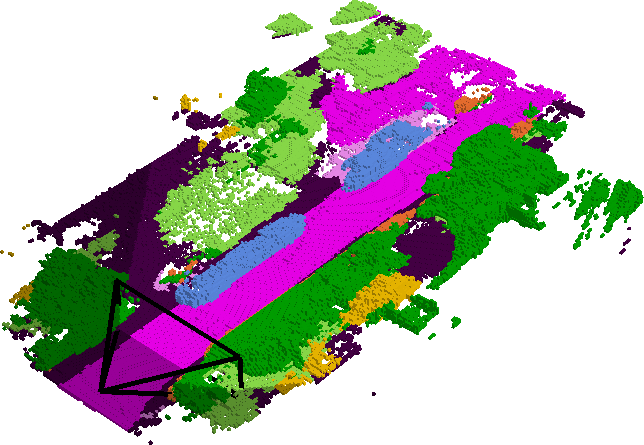} & 
		\includegraphics[width=.9\linewidth]{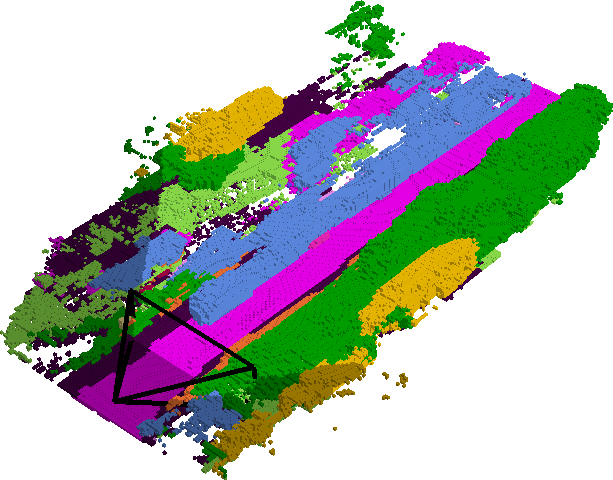} & 
		\includegraphics[width=.9\linewidth]{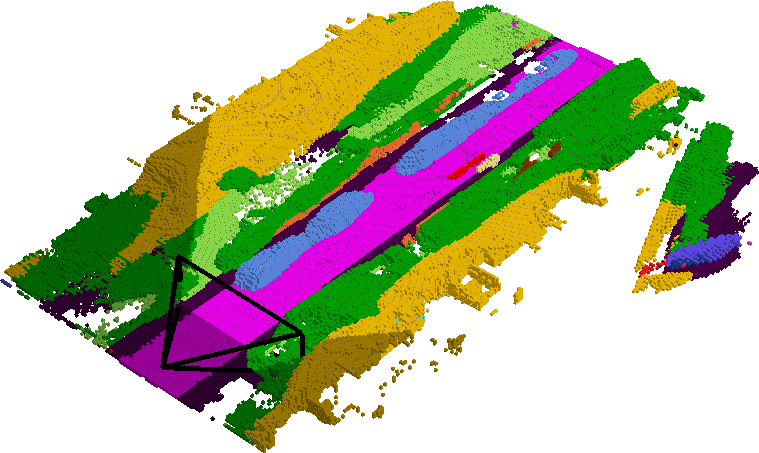} & 
		\includegraphics[width=\linewidth]{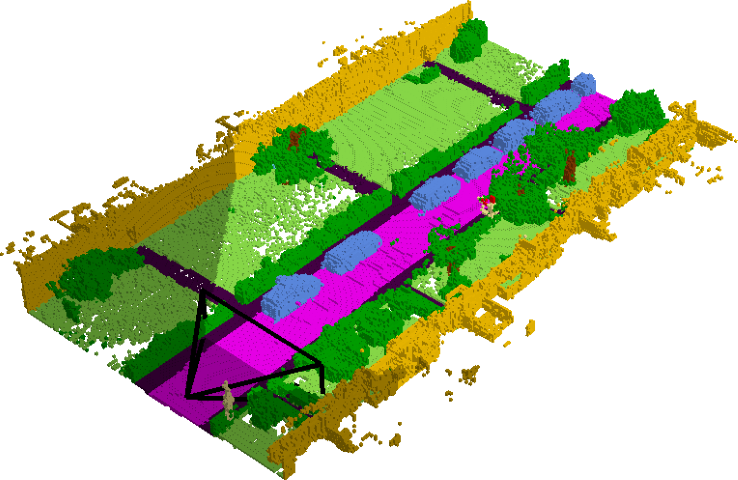} 
		\\
		\includegraphics[width=.9\linewidth]{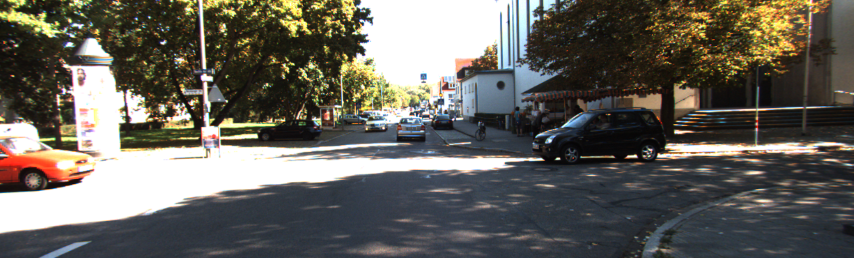} &
		\includegraphics[width=.9\linewidth]{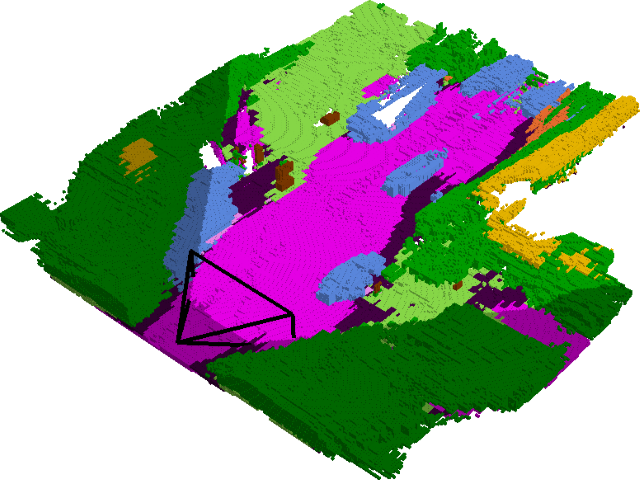} & 
		\includegraphics[width=.9\linewidth]{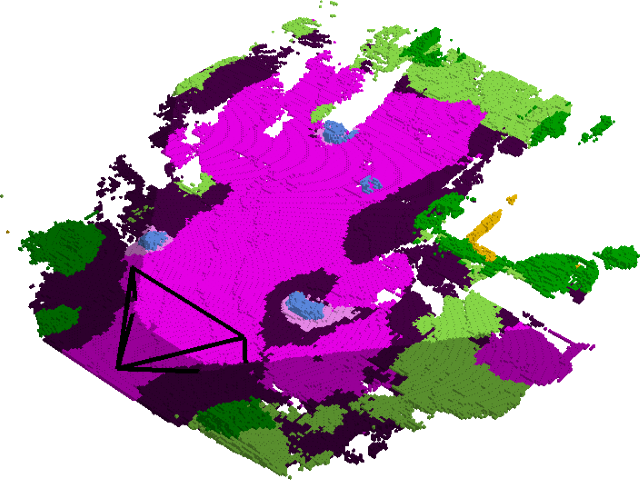} & 
		\includegraphics[width=.9\linewidth]{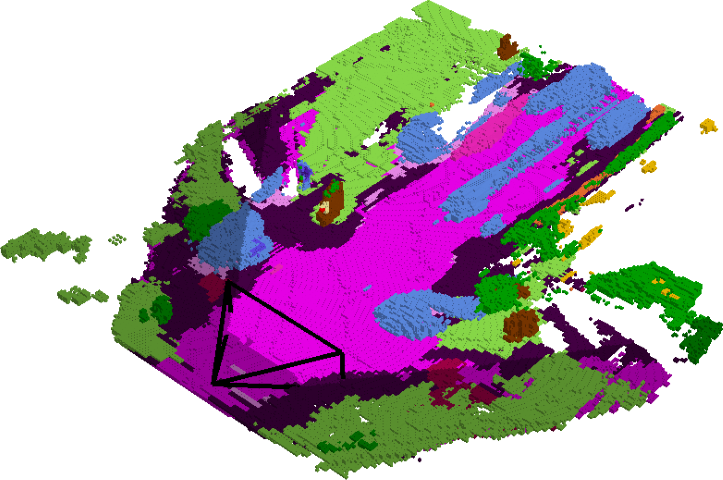} & 
		\includegraphics[width=.9\linewidth]{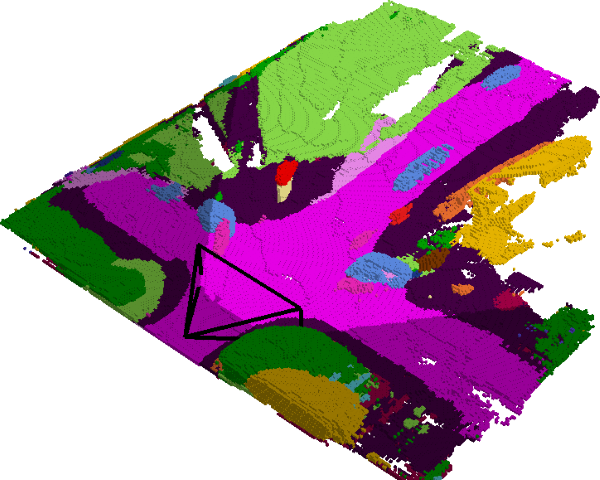} & 
		\includegraphics[width=\linewidth]{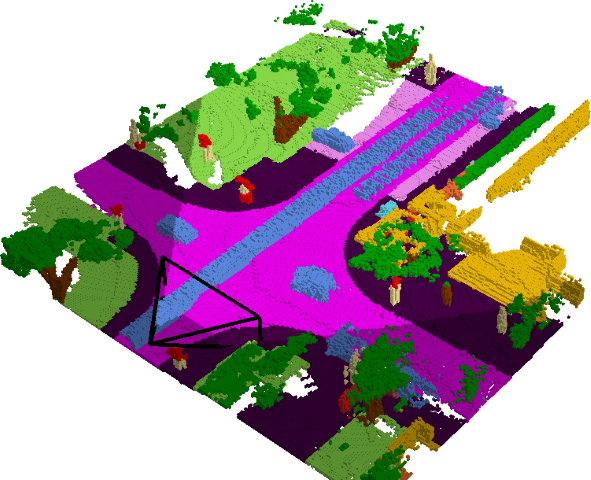} 
		\\
		\multicolumn{6}{c}{
			\scriptsize
			\textcolor{bicycle}{$\blacksquare$}bicycle~
			\textcolor{car}{$\blacksquare$}car~
			\textcolor{motorcycle}{$\blacksquare$}motorcycle~
			\textcolor{truck}{$\blacksquare$}truck~
			\textcolor{other-vehicle}{$\blacksquare$}other vehicle~
			\textcolor{person}{$\blacksquare$}person~
			\textcolor{bicyclist}{$\blacksquare$}bicyclist~
			\textcolor{motorcyclist}{$\blacksquare$}motorcyclist~
			\textcolor{road}{$\blacksquare$}road~
			\textcolor{parking}{$\blacksquare$}parking~}\\
		\multicolumn{6}{c}{
			\scriptsize
			\textcolor{sidewalk}{$\blacksquare$}sidewalk~
			\textcolor{other-ground}{$\blacksquare$}other ground~
			\textcolor{building}{$\blacksquare$}building~
			\textcolor{fence}{$\blacksquare$}fence~
			\textcolor{vegetation}{$\blacksquare$}vegetation~
			\textcolor{trunk}{$\blacksquare$}trunk~
			\textcolor{terrain}{$\blacksquare$}terrain~
			\textcolor{pole}{$\blacksquare$}pole~
			\textcolor{traffic-sign}{$\blacksquare$}traffic sign			
		}
	\end{tabular}		
	\caption{\textbf{Results on SemanticKITTI~\cite{behley2019iccv} (validation set).} The input is shown left. Darker voxels represent the scenery outside the viewing frustum (\ie unseen by the image).}
	\label{fig:qualitative_kitti_supp}
\end{figure*}

\paragraph{Qualitative performance.}
In \cref{fig:qualitative_kitti_supp} we also include additional qualitative results. Compared to all baselines, MonoScene captures better landscape and objects (\eg cars, rows 3-9; pedestrian, rows 6, 10; traffic-sign, rows 3, 5). Still, it struggles to predict thin small objects (\eg trunk, row 1; pedestrian, row 3; traffic-sign, row 2, 6), separate far away consecutive cars (\eg row 5, 7, 8), and infer very complex, highly cluttered scenes (\eg rows 9, 10).

\paragraph{Evaluation scope.} \cref{tab:out_fov_perf} reports the performance when considering either only the voxels inside FOV (in-FOV), outside FOV (out-FOV), or all voxels (Whole Scene) as reported in the main paper. Compared to the Whole Scene, the in-FOV performance is higher since it considers visible surfaces, whereas the out-FOV performance is significantly lower since the image does not observe it.

\begin{table}[b]
	\footnotesize
	\centering
	\setlength{\tabcolsep}{0.01\linewidth}
	\begin{tabular}[b]{l|cc|cc|cc}
		\toprule
		& \multicolumn{2}{c}{in-FOV} & \multicolumn{2}{c}{out-FOV} & \multicolumn{2}{c}{Whole Scene} \\
		& IoU $\uparrow$ & mIoU $\uparrow$ & IoU $\uparrow$ & mIoU $\uparrow$ & IoU $\uparrow$ & mIoU $\uparrow$ \\
		\midrule
		LMSCNet$^\text{rgb}$~\cite{roldao2020lmscnet} & 37.62 & 8.87 & 25.36 & 5.48 & 34.41 & 8.17 \\
		3DSketch$^\text{rgb}$~\cite{3DSketch} & 32.24 & 7.82 & 26.50 & 5.83 & 33.30 & 7.50 \\
		AICNet$^\text{rgb}$~\cite{Li2020aicnet} & 35.69 & 8.75 & 25.79 & 5.61 & 29.59 & 8.31 \\
		*JS3C-Net$^\text{rgb}$~\cite{yan2021JS3CNet} & \textbf{42.22} & \underline{11.29} & \underline{28.27} & \underline{6.31} & \textbf{38.98} & \underline{10.31} \\
		\midrule
		MonoScene(ours) & \underline{39.13} & \textbf{12.78} & \textbf{31.60} & \textbf{7.45} & \underline{37.12} & \textbf{11.50} \\
		\bottomrule
	\end{tabular}\\
	{* Uses pretrained semantic segmentation network.}
	
	\caption{\textbf{SemanticKITTI performance (validation set) on in-/out-FOV and the Whole Scene.} We report the performance on the scenery inside (in-FOV), outside (out-FOV) camera FOV, and considering all voxels (Whole Scene). MonoScene is best in most cases, with in-FOV performance logically higher.}
	\label{tab:out_fov_perf}
\end{table}

\subsection{NYUv2}
\begin{figure*}	
	\centering
	\footnotesize
	\renewcommand{\arraystretch}{0.0}
	\setlength{\tabcolsep}{0.003\textwidth}
	\newcolumntype{P}[1]{>{\centering\arraybackslash}m{#1}}
	\begin{tabular}{P{0.14\textwidth} P{0.15\textwidth} P{0.15\textwidth} P{0.15\textwidth} P{0.15\textwidth} P{0.15\textwidth}}
		Input & LMSCNet$^\text{rgb}$~\cite{roldao2020lmscnet} & AICNet$^\text{rgb}$~\cite{Li2020aicnet} & 3DSketch$^\text{rgb}$~\cite{3DSketch} & MonoScene \scriptsize{(ours)} & Ground Truth
		\\
		\includegraphics[width=\linewidth]{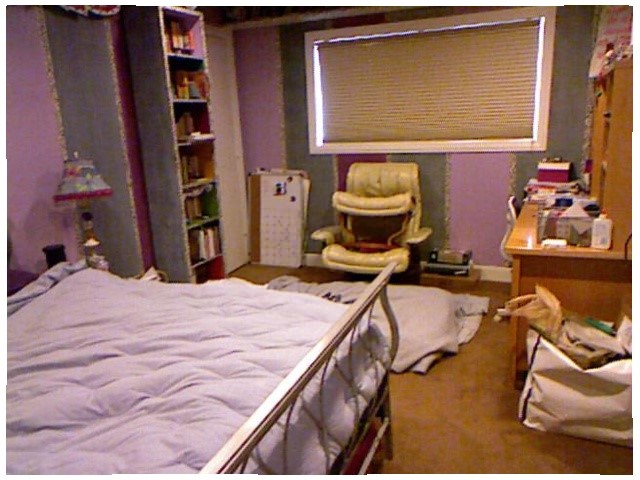} & 
		\includegraphics[width=\linewidth]{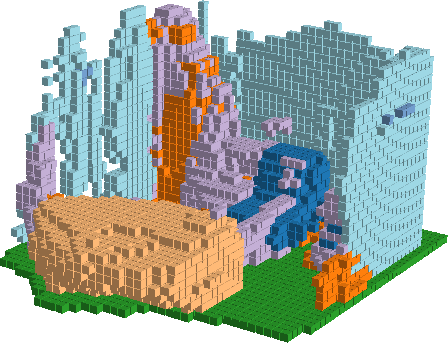} &
		\includegraphics[width=\linewidth]{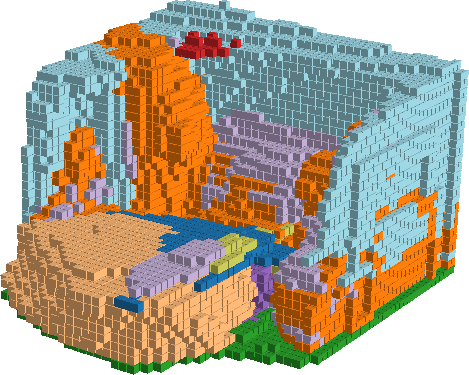} &
		\includegraphics[width=\linewidth]{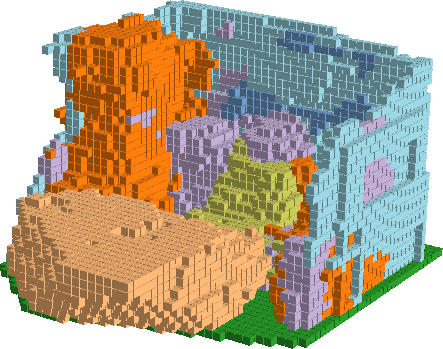} &
		\includegraphics[width=\linewidth]{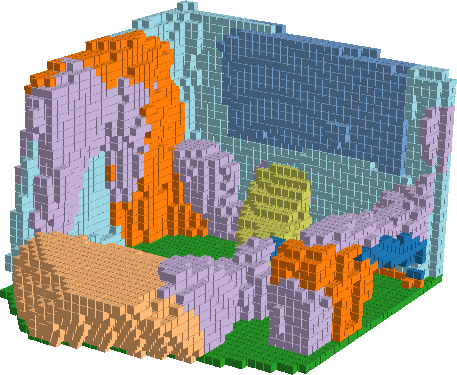} &
		\includegraphics[width=\linewidth]{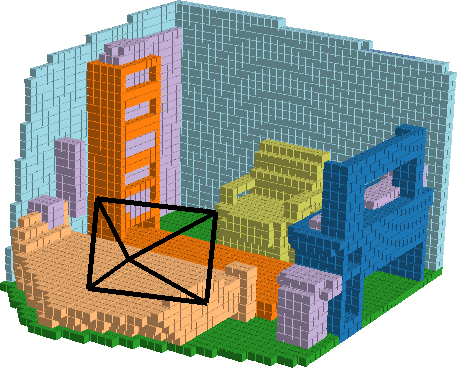}
		\\	
		\includegraphics[width=\linewidth]{imgs/nyu_qualitative/NYU0435_0000_2_color.jpg} & 
		\includegraphics[width=\linewidth]{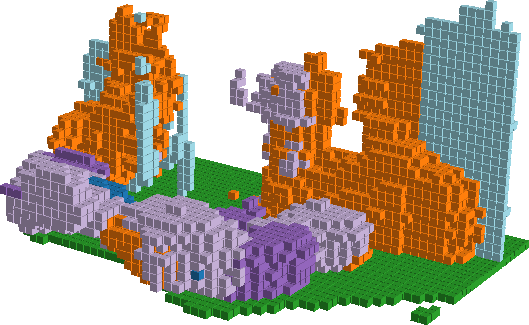} &
		\includegraphics[width=\linewidth]{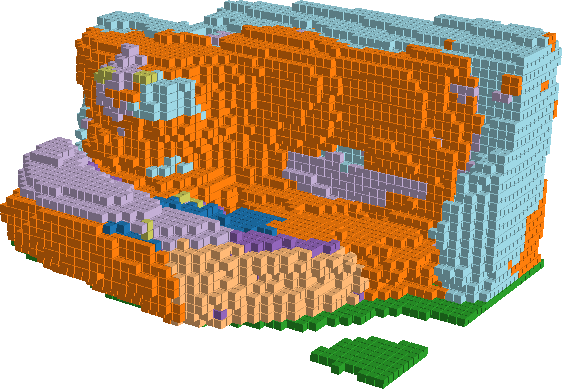} &
		\includegraphics[width=\linewidth]{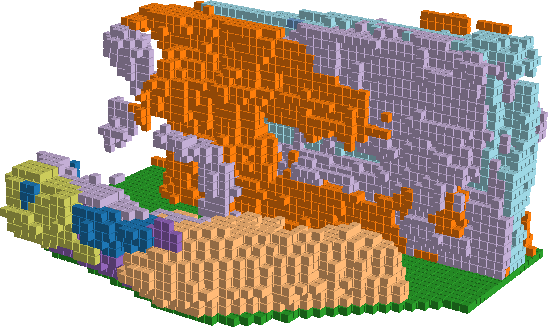} &
		\includegraphics[width=\linewidth]{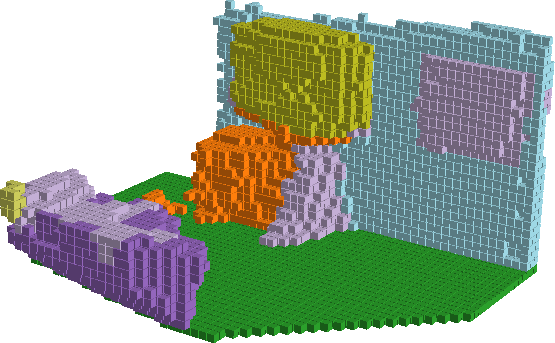} &
		\includegraphics[width=\linewidth]{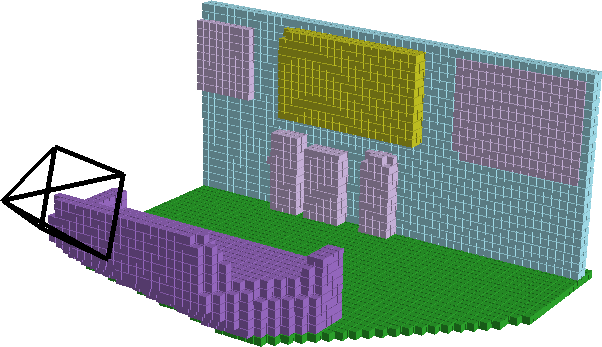}
		\\	
		\includegraphics[width=\linewidth]{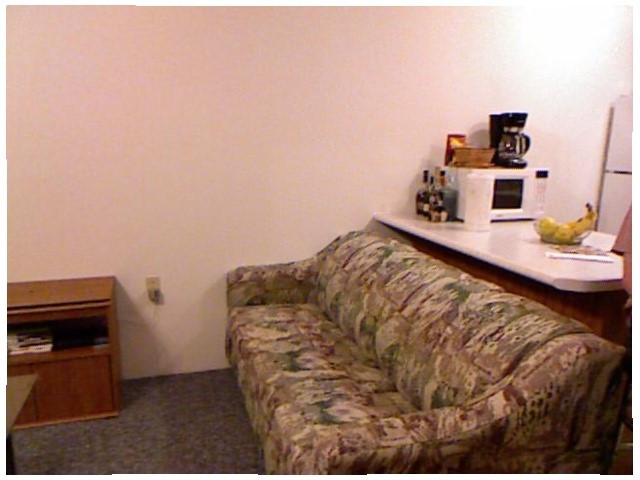} & 
		\includegraphics[width=\linewidth]{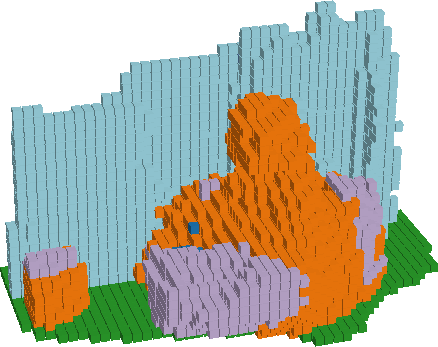} &
		\includegraphics[width=\linewidth]{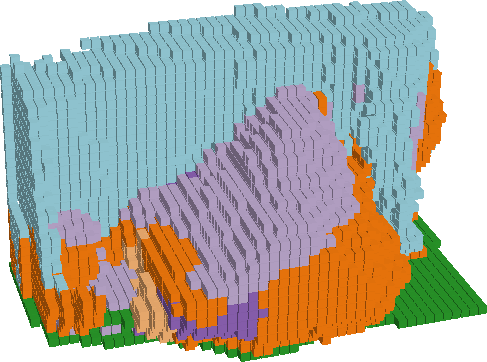} &
		\includegraphics[width=\linewidth]{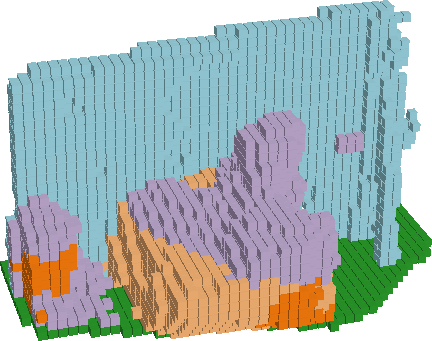} &
		\includegraphics[width=\linewidth]{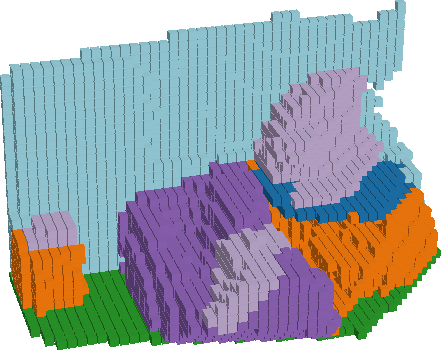} &
		\includegraphics[width=\linewidth]{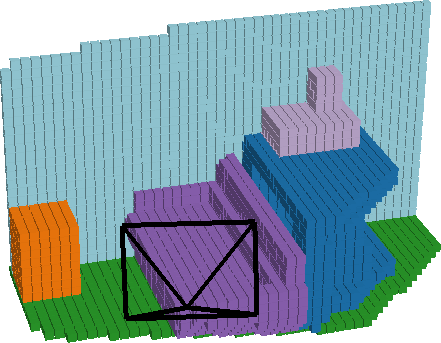}
		\\	
		\includegraphics[width=\linewidth]{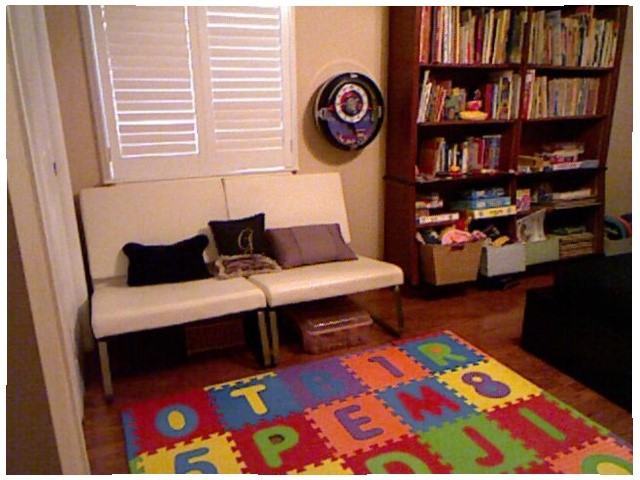} & 
		\includegraphics[width=\linewidth]{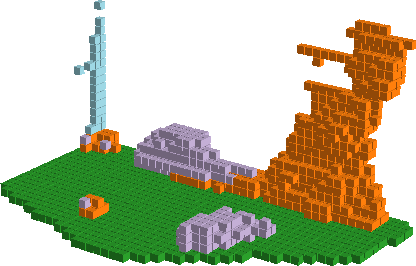} &
		\includegraphics[width=\linewidth]{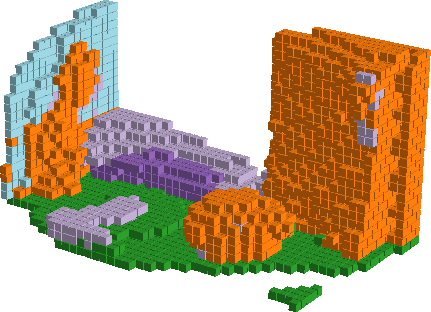} &
		\includegraphics[width=\linewidth]{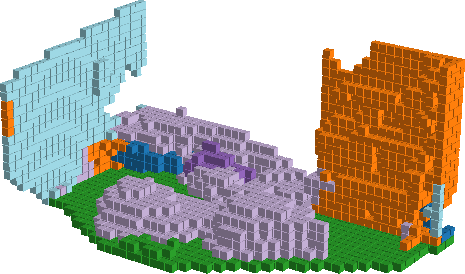} &
		\includegraphics[width=\linewidth]{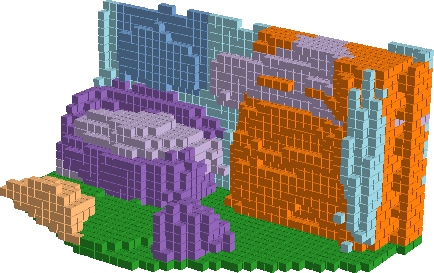} &
		\includegraphics[width=\linewidth]{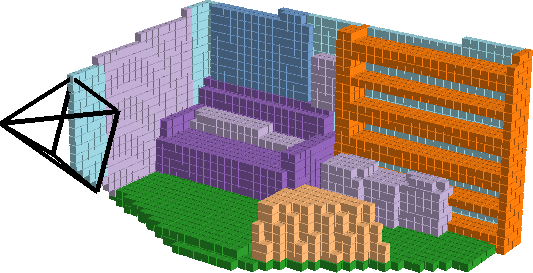}
		\\	
		\includegraphics[width=\linewidth]{imgs/nyu_qualitative/NYU0917_0000_2_color.jpg} & 
		\includegraphics[width=\linewidth]{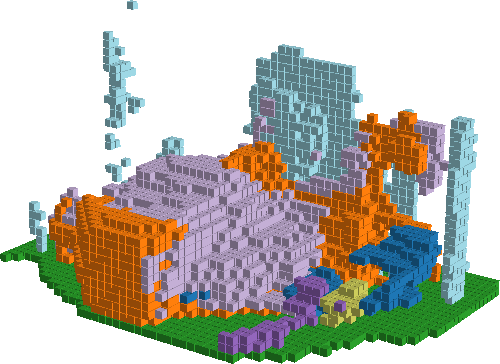} &
		\includegraphics[width=\linewidth]{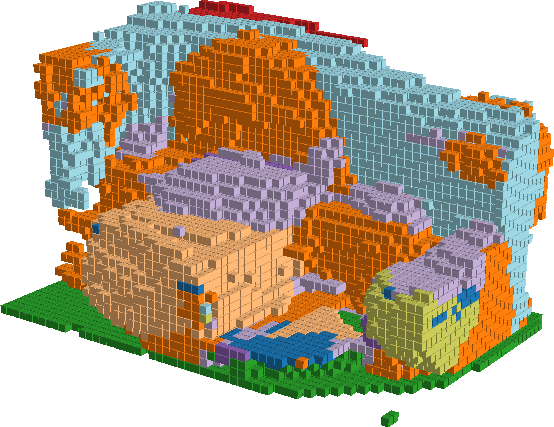} &
		\includegraphics[width=\linewidth]{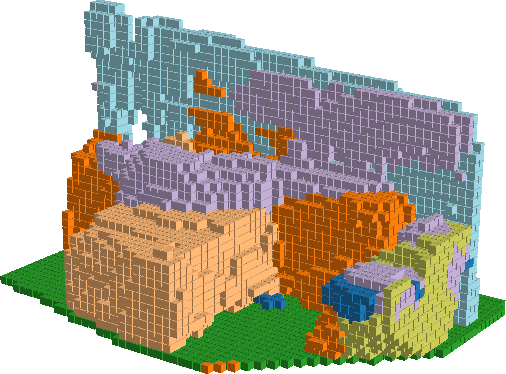} &
		\includegraphics[width=\linewidth]{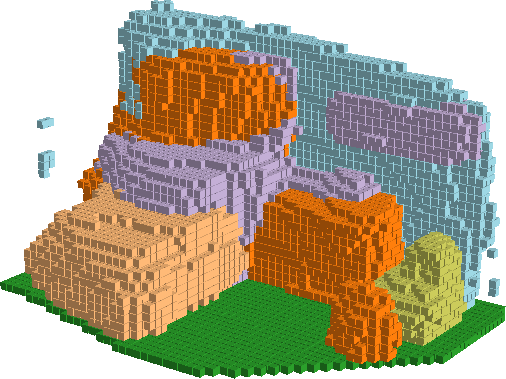} &
		\includegraphics[width=\linewidth]{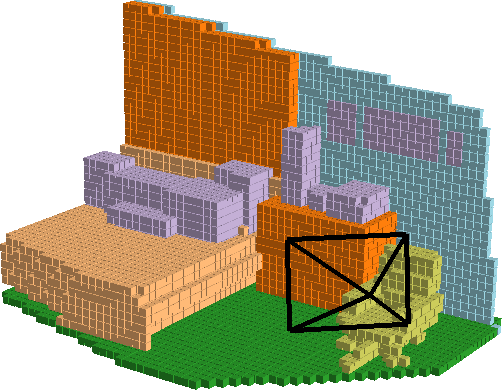}
		\\
		\includegraphics[width=\linewidth]{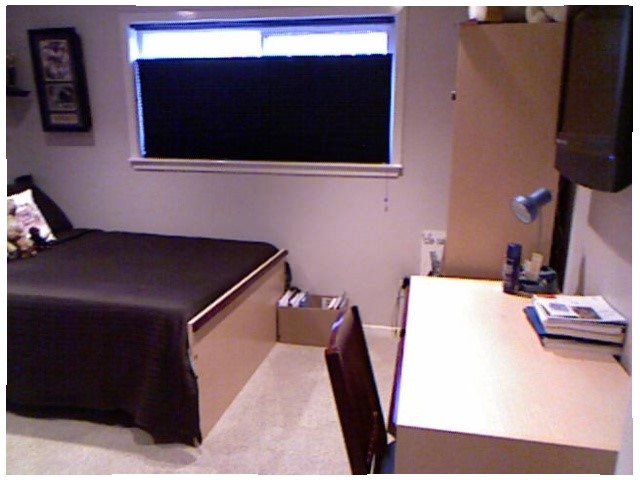} & 
		\includegraphics[width=\linewidth]{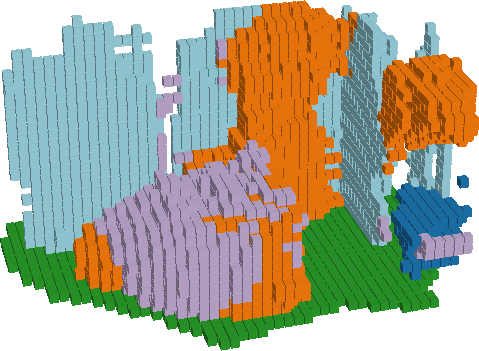} &
		\includegraphics[width=\linewidth]{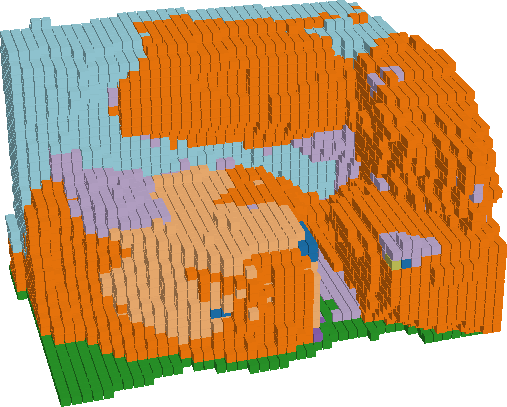} &
		\includegraphics[width=\linewidth]{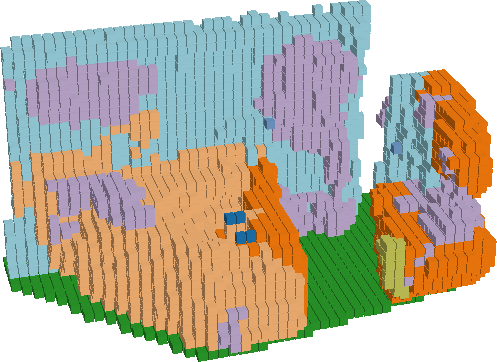} &
		\includegraphics[width=\linewidth]{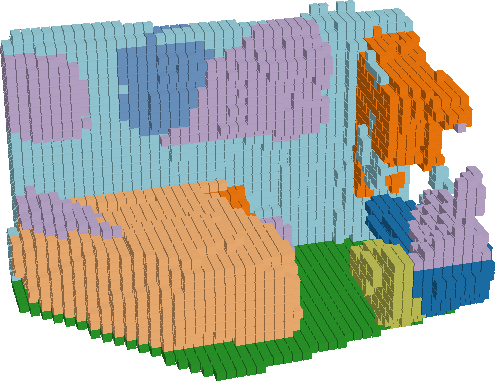} &
		\includegraphics[width=\linewidth]{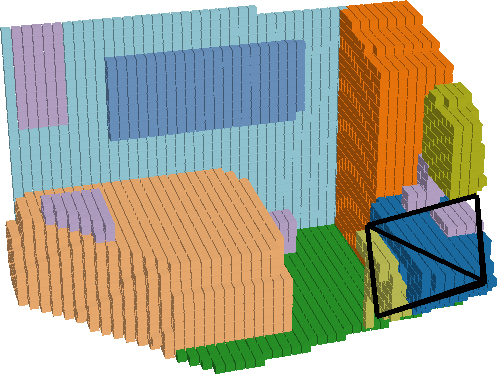}
		\\
		\includegraphics[width=.9\linewidth]{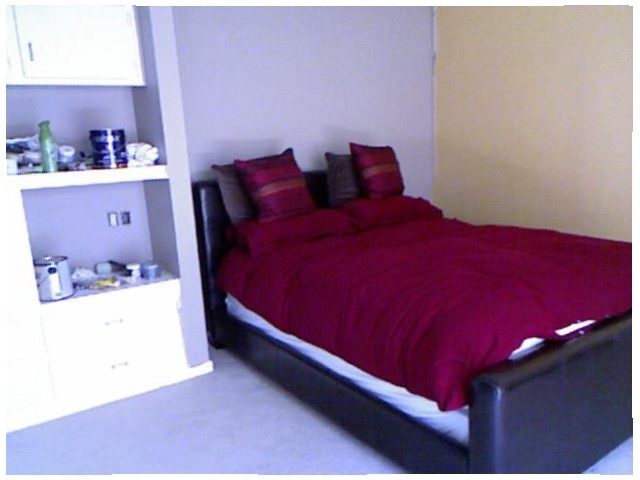} & 
		\includegraphics[width=.9\linewidth]{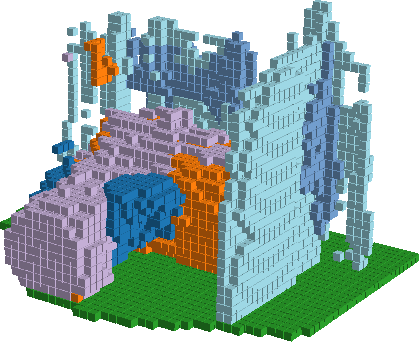} &
		\includegraphics[width=.9\linewidth]{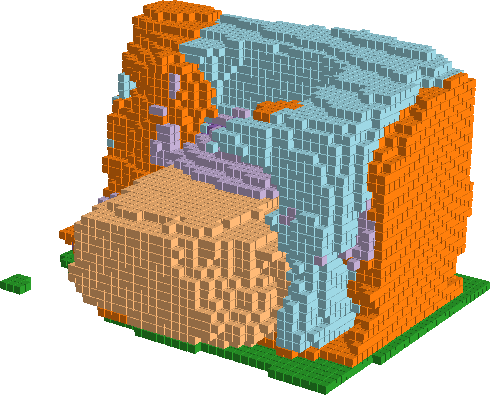} &
		\includegraphics[width=.9\linewidth]{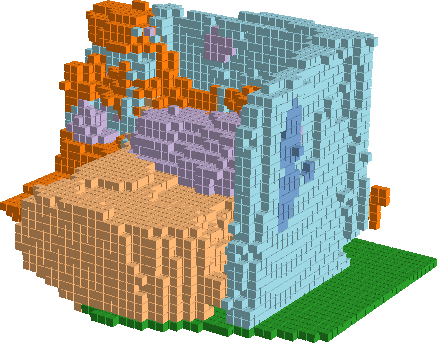} &
		\includegraphics[width=.8\linewidth]{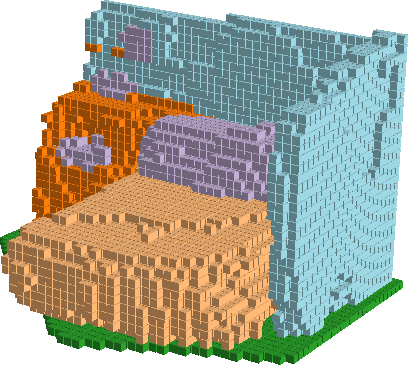} &
		\includegraphics[width=\linewidth]{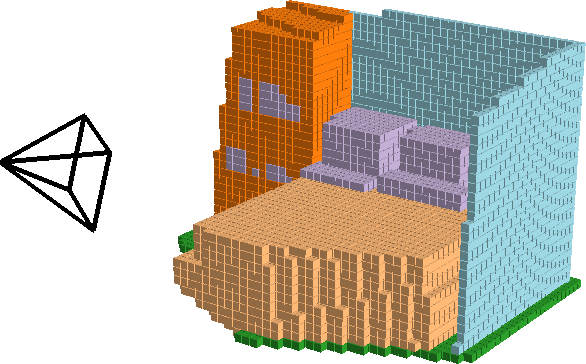}
		\\
		\includegraphics[width=\linewidth]{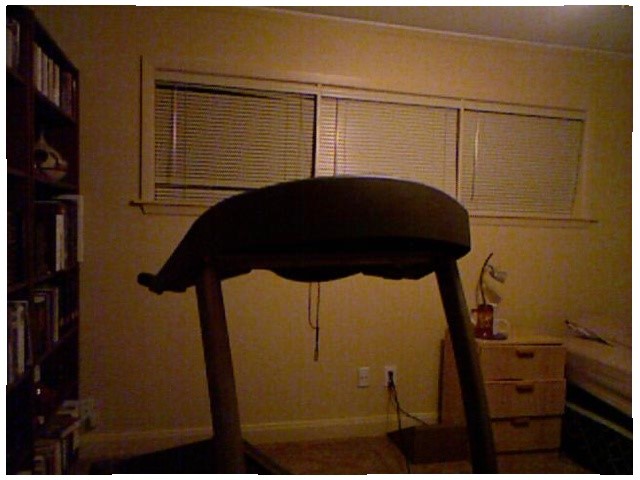} & 
		\includegraphics[width=\linewidth]{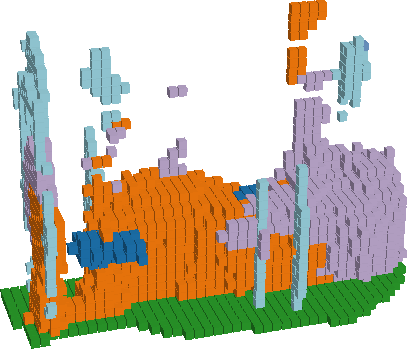} &
		\includegraphics[width=\linewidth]{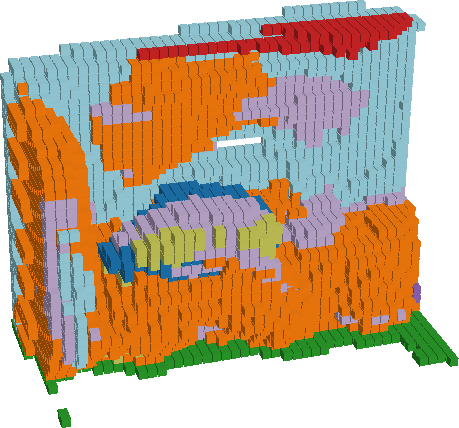} &
		\includegraphics[width=\linewidth]{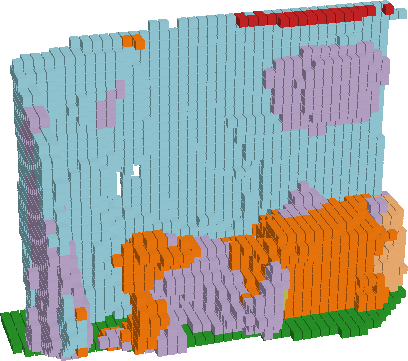} &
		\includegraphics[width=\linewidth]{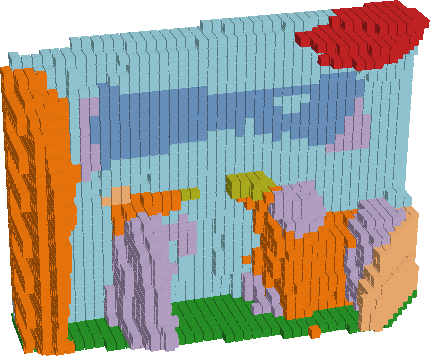} &
		\includegraphics[width=\linewidth]{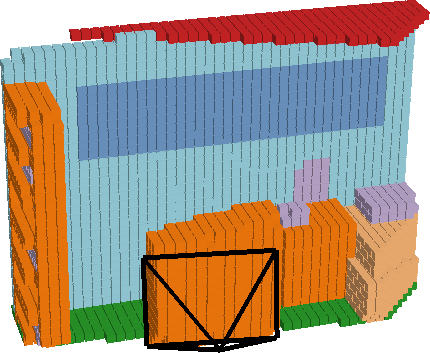}
		\\
		\includegraphics[width=.9\linewidth]{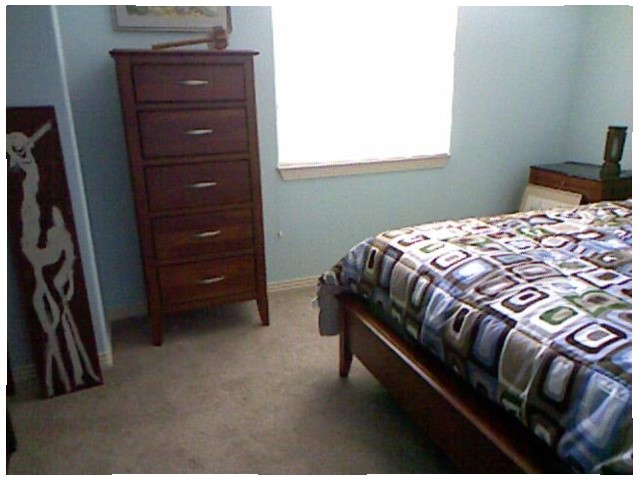} & 
		\includegraphics[width=.9\linewidth]{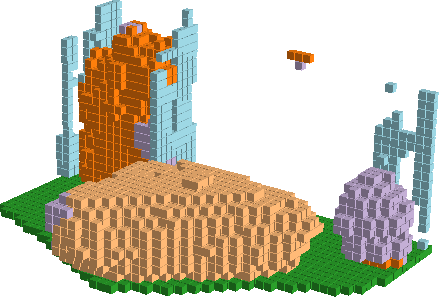} &
		\includegraphics[width=.9\linewidth]{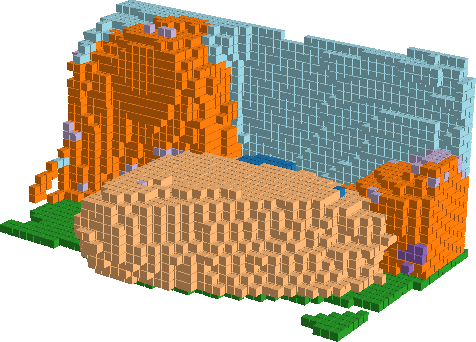} &
		\includegraphics[width=.9\linewidth]{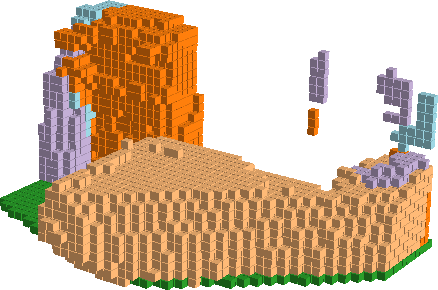} &
		\includegraphics[width=.9\linewidth]{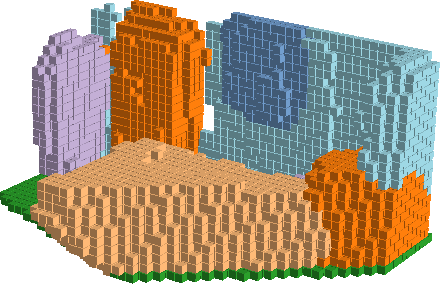} &
		\includegraphics[width=\linewidth]{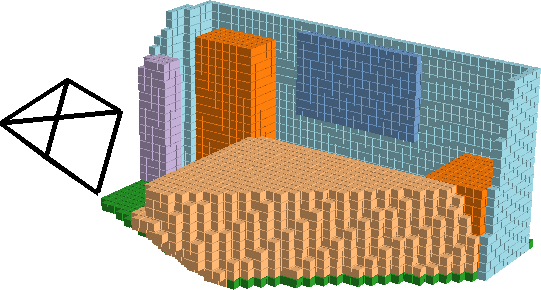}
		\\
		\includegraphics[width=\linewidth]{imgs/nyu_qualitative/NYU1162_0000_2_color.jpg} & 
		\includegraphics[width=\linewidth]{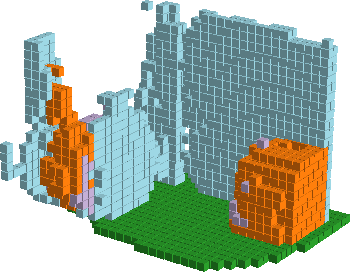} &
		\includegraphics[width=\linewidth]{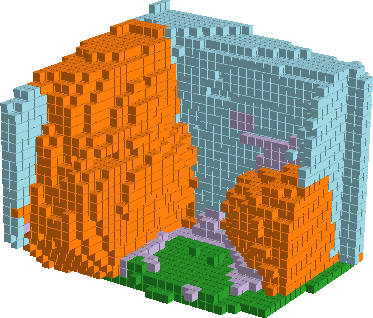} &
		\includegraphics[width=\linewidth]{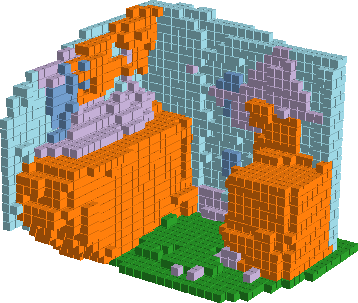} &
		\includegraphics[width=\linewidth]{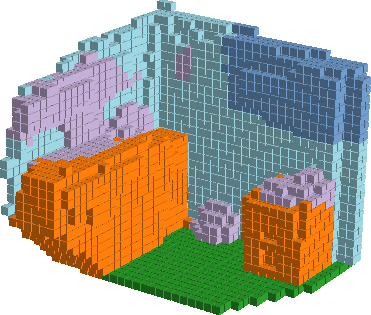} &
		\includegraphics[width=\linewidth]{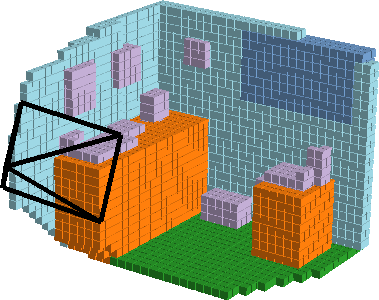}
		\\
		\multicolumn{6}{c}{
			\scriptsize
			\textcolor{ceiling}{$\blacksquare$}ceiling~
			\textcolor{floor}{$\blacksquare$}floor~
			\textcolor{wall}{$\blacksquare$}wall~
			\textcolor{window}{$\blacksquare$}window~
			\textcolor{chair}{$\blacksquare$}chair~
			\textcolor{sofa}{$\blacksquare$}sofa~
			\textcolor{table}{$\blacksquare$}table~
			\textcolor{tvs}{$\blacksquare$}tvs~
			\textcolor{furniture}{$\blacksquare$}furniture~
			\textcolor{objects}{$\blacksquare$}objects
		}
	\end{tabular}
	\caption{\textbf{Results on NYUv2\cite{NYUv2} (test set).} The input is shown leftmost and the camera viewing frustum is shown in the ground truth (rightmost).}
	\label{fig:qualitative_NYU_supp}
\end{figure*}

We show additional qualitative results in \cref{fig:qualitative_NYU_supp}. In overall, MonoScene predicts better scene layouts and better objects geometry, evidently in rows 1-4, 6, 9, 10. Still, MonoScene mispredicts complex (\eg bookshelfs, row 1, 4, 6), or rare objects (running machine, row 8). Sometimes, it confuses semantically-similar classes (\eg window/objects, row 6, 8; beds/objects, row 1, 5; furniture/table, row 1, 2) due to the high variance of indoor scene \ie wide range of camera poses, objects have completely different appearances, poses and positions even in the same category \eg beds (rows 1, 5-7, 9); sofa (row 2-4).  

\subsection{Generalization}
\begin{figure*}
	\centering
	\footnotesize
	\setlength{\fboxsep}{0.005pt}
	\newcommand{\tmpframe}[1]{\fbox{#1}}
	\newcommand{\im}[3]{\stackinset{r}{0cm}{b}{0pt}{\tmpframe{\includegraphics[height=#3]{#1}}}{\includegraphics[width=\linewidth]{#2}}} %
	\newcommand{\imcs}[1]{\im{imgs/generalization/cityscapes/#1.png}{imgs/generalization/cityscapes/#1_our.png}{1.0cm}}
	\newcommand{\imns}[1]{\im{imgs/generalization/nuscenes/#1_nclass.jpg}{imgs/generalization/nuscenes/#1_nclass_0_our.png}{1.0cm}}
	\newcommand{\imkt}[1]{\im{imgs/kitti_qualitative/#1.png}{imgs/kitti_qualitative/#1_0_our.png}{0.72cm}}
	\newcommand{\imktnew}[1]{\im{imgs/generalization/kitti360/#1.png}{imgs/generalization/kitti360/#1_0_our.png}{0.7cm}}
	\newcolumntype{P}[1]{>{\centering\arraybackslash}m{#1}}
	\setlength{\tabcolsep}{0.01\linewidth}
	\renewcommand{\arraystretch}{0.0}
	\begin{tabular}{P{0.235\linewidth} P{0.235\linewidth} P{0.235\linewidth} P{0.235\linewidth}}		
		Cityscapes~\cite{cityscapes} & 
		nuScenes~\cite{nuscenes2019} & 
		\textbf{SemanticKITTI~\cite{behley2019iccv}}
		& KITTI-360~\cite{KITTI360}\\
		\imcs{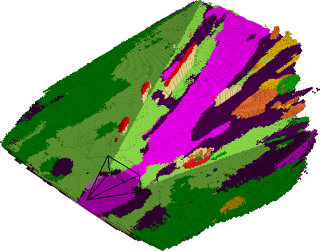} & \imns{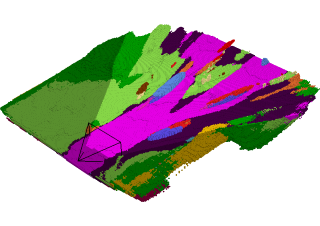} & \imkt{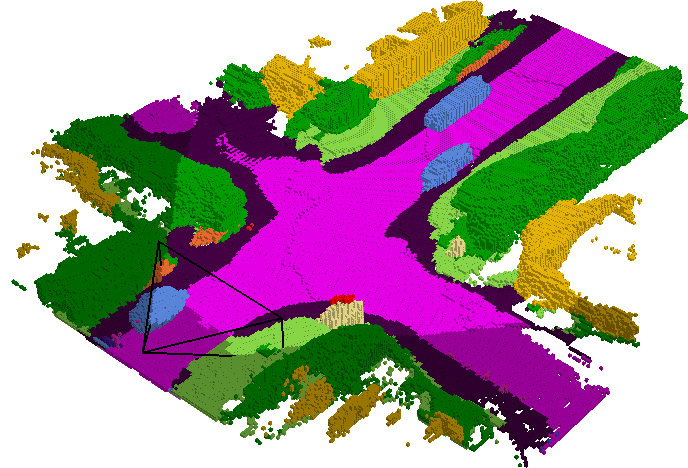} & \imktnew{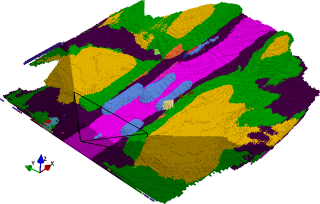}
		\\
		\imcs{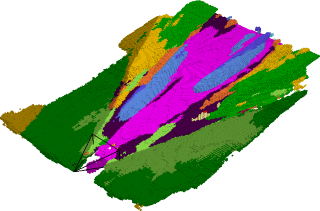} & \imns{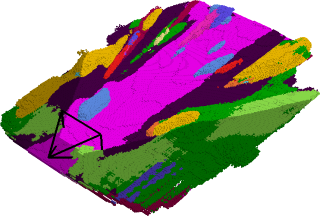} & \imkt{000290} & \imktnew{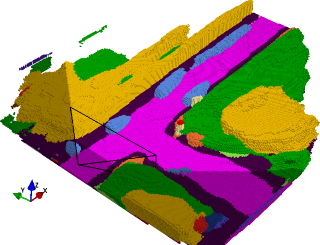}
		\\
		\imcs{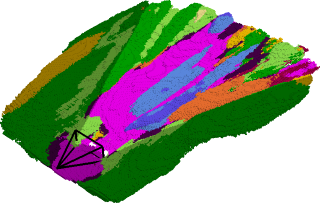} & \imns{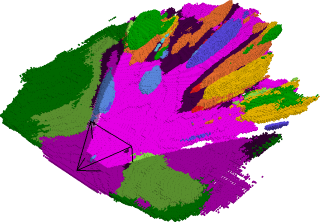} & \imkt{001500} & \imktnew{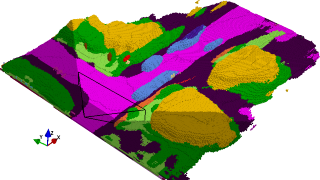}
		\\
		\imcs{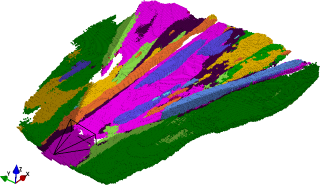} & \imns{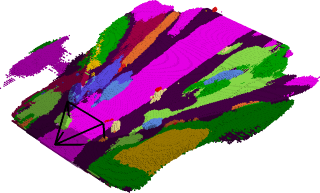} & \imkt{002505} & \imktnew{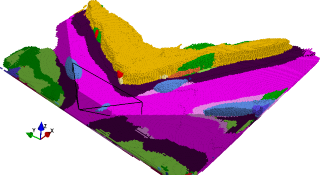}
		\\
		\imcs{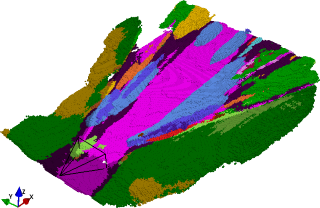} & \imns{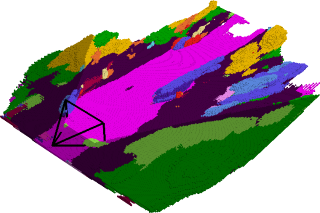} & \imkt{000010} & \imktnew{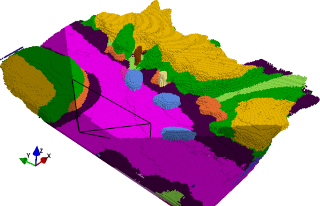}
		\\
		(49$^{\circ}$, 25$^{\circ}$) & (65$^{\circ}$, 39$^{\circ}$) & (H=82$^{\circ}$, V=29$^{\circ}$) & (104$^{\circ}$, 38$^{\circ}$)
		\\
	\end{tabular}
	\caption{\textbf{Domain gap and Camera effects.} Outputs of MonoScene when trained on SemanticKITTI having horizontal FOV of 82$^\circ$, and tested on datasets with decreasing (left) or increasing (right) FOV. SemanticKITTI and KITTI-360 are recored in mid-size German city of Karlsruhe while nuScenes and Cityscapes are from large metropolitan areas (\eg Stuttgart - Cityscapes; Singapore, Boston - nuScenes) whose streets are much wider, denser and have different landscapes.}
	\label{fig:fov_generalization_supp}
\end{figure*}

\cref{fig:fov_generalization_supp} illustrates the predictions of MonoScene, trained on SemanticKITTI, on datasets with different camera setups. 
We can see the increase in distortion as the camera setups depart from the ones used during training. Furthermore, the domain gap (\ie city, country, etc.) also plays an important role. 
As MonoScene is trained on the mid-size German city of Karlsruhe, with residential scenes and narrow roads, the gap is smaller with KITTI-360 having similar scenes.
The results on nuScenes and Cityscapes suffer both from the camera setup changes and the large metropolitan scenes (\ie Stuttgart - Cityscapes; Singapore, Boston - nuScenes) having wider streets.

\newpage

{\small
\bibliographystyle{ieee_fullname}
\bibliography{egbib}
}

\end{document}